\documentclass{article}

\usepackage{microtype}
\usepackage{graphicx}
\usepackage{subcaption}
\usepackage{booktabs} % for professional tables
\usepackage{wrapfig}
\usepackage{enumitem}

\usepackage{hyperref}

\usepackage[accepted]{icml2026}

\usepackage{amsmath}
\usepackage{amssymb}
\usepackage{mathtools}
\usepackage{amsthm}

\usepackage[capitalize,noabbrev]{cleveref}

\theoremstyle{plain}
\newtheorem{theorem}{Theorem}[section]
\newtheorem{proposition}[theorem]{Proposition}
\newtheorem{lemma}[theorem]{Lemma}
\newtheorem{corollary}[theorem]{Corollary}
\theoremstyle{definition}
\newtheorem{definition}[theorem]{Definition}

\theoremstyle{remark}
\newtheorem{remark}[theorem]{Remark}

\usepackage[textsize=tiny]{todonotes}

\newcommand{\step}[2]{\par\noindent\textbf{#1.~#2}}

\let\ge\geqslant
\let\geq\geqslant
\let\le\leqslant
\let\leq\leqslant

\begin{document}

\twocolumn[
  \icmltitle{Token Sample Complexity of Attention}

  \icmlsetsymbol{equal}{*}

  \begin{icmlauthorlist}
  \icmlauthor{Léa Bohbot}{ens}
  \icmlauthor{Cyril Letrouit}{lmo}
  \icmlauthor{Gabriel Peyré}{cnrsens}
  \icmlauthor{François-Xavier Vialard}{ligm}
\end{icmlauthorlist}

\icmlaffiliation{ens}{ENS, PSL Univ.}
\icmlaffiliation{lmo}{LMO, Univ. Paris-Saclay, CNRS}
\icmlaffiliation{cnrsens}{CNRS, ENS, PSL Univ.}
\icmlaffiliation{ligm}{LIGM, Univ. Gustave Eiffel, CNRS}

 \icmlcorrespondingauthor{Léa Bohbot}{lea.bohbot@ens.psl.eu}

  \icmlkeywords{Machine Learning, ICML}

  \vskip 0.3in
]



\printAffiliationsAndNotice{}  % no special notice (required even if empty)

\begin{abstract}
  As context windows in large language models continue to expand, it is essential to characterize how attention behaves at extreme sequence lengths. We introduce token sample complexity: the rate at which attention computed on $n$ tokens converges to its infinite-token limit. We estimate finite-$n$ convergence bounds at two levels: pointwise uniform convergence of the attention map, and convergence of moments for the transformed token distribution. 
    For compactly supported (and more generally sub-Gaussian) distributions, our first result shows that the attention map converges uniformly on a ball of radius $R$ at rate $C(R)/\sqrt{n}$, where $C(R)$ grows exponentially with $R$. For large $R$, this estimate loses practical value, and our second result addresses this issue by establishing convergence rates for the moments of the transformed distribution (the token output of the attention layer). In this case, the rate is $C'(R)/n^{\beta}$ with $\beta<\tfrac{1}{2}$, and $C'(R)$ depends polynomially on the size of the support of the distribution. The exponent $\beta$ depends on the attention geometry and the spectral properties of the token distribution. We also examine the regime in which the attention parameter tends to infinity and the softmax approaches a hardmax, and in this setting, we establish a logarithmic rate of convergence. Experiments on synthetic and real data support our predictions and show that the predicted slowdown is reflected in downstream accuracy.
\end{abstract}

\section{Introduction} 
The remarkable success of Transformer models has been driven by attention mechanisms that flexibly handle very long input sequences \cite{Vaswani17,Devlin19}. Recent large-context Transformers extend context windows from $100$K to $10$M tokens \cite{Reid24,Hooper24}, raising fundamental questions about how attention behaves in the regime of extremely large sequences, and how much benefit is gained by providing ever more tokens as input. In this work, we introduce the notion of \emph{token sample complexity} to characterize the precision gained by feeding additional tokens into an already-trained Transformer. Unlike classical sample complexity, which studies performance as training data increases, we fix the Transformer parameters and examine a different asymptotic regime: for a fixed input distribution, we let the number $n$ of context tokens grow. 
Our goal is to understand how attention outputs behave in this limit, and how many tokens are required to approach the limiting behavior.

\begin{figure}[ht]
\vskip  -0.08in
  \begin{center}
  \includegraphics[width=0.8\columnwidth]{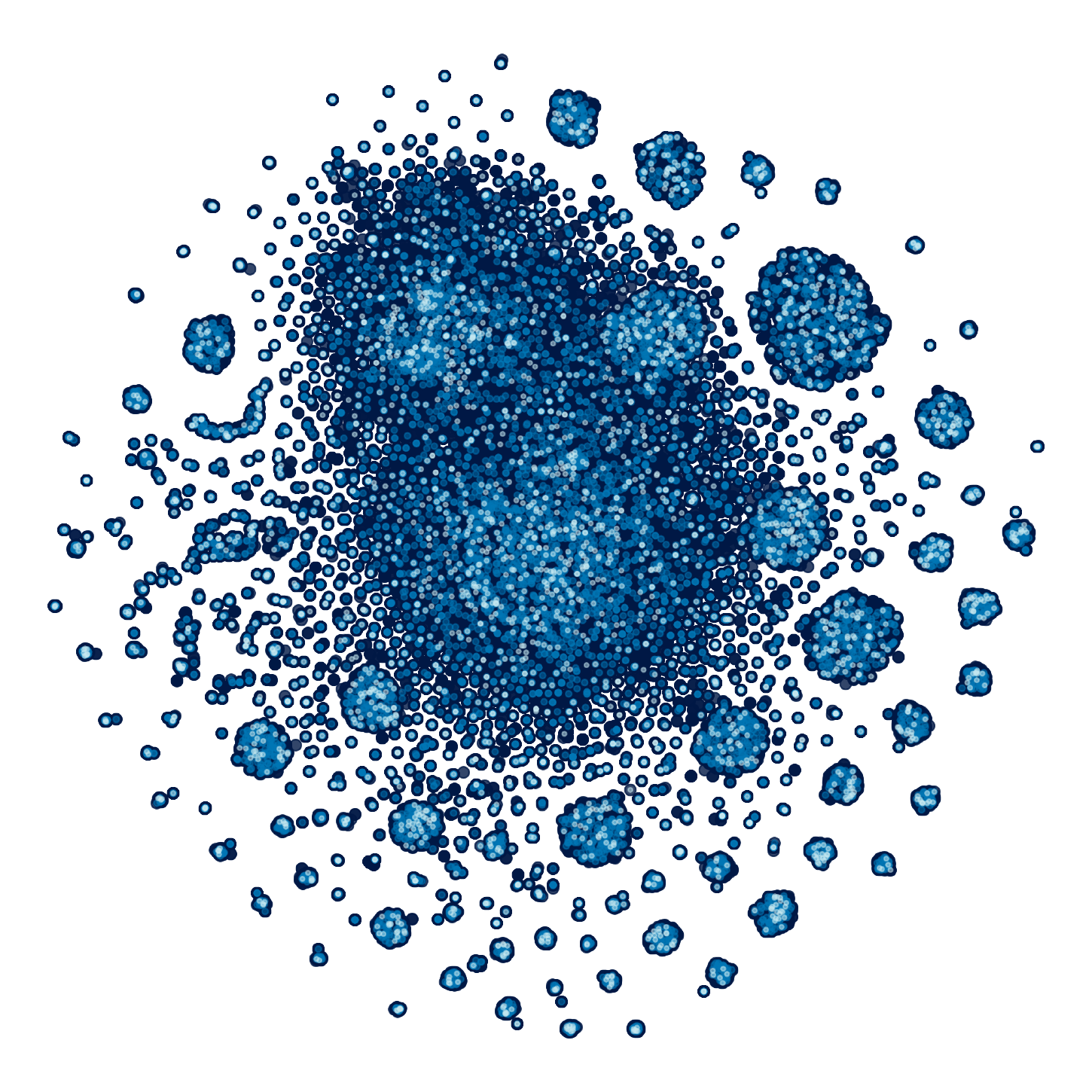}
  \caption{\textbf{Illustration of token distribution convergence toward a limit distribution} using t-SNE visualization of the first $50000$ token embeddings from WikiText-103 (BigBird tokenizer). Colors (light/medium/dark blue) represent $n=3000, 10000,$ and $50000$ subsampled tokens: as $n$ grows, the token distribution $\nu_n$ covers the support more uniformly.}
  \label{fig:tsne_limit}
    \end{center}
   \vskip -0.2in
\end{figure}
\begin{figure*}[t!]
    \centering
    \begin{subfigure}[t!]{0.32\textwidth}
        \centering
        \includegraphics[width=\linewidth]{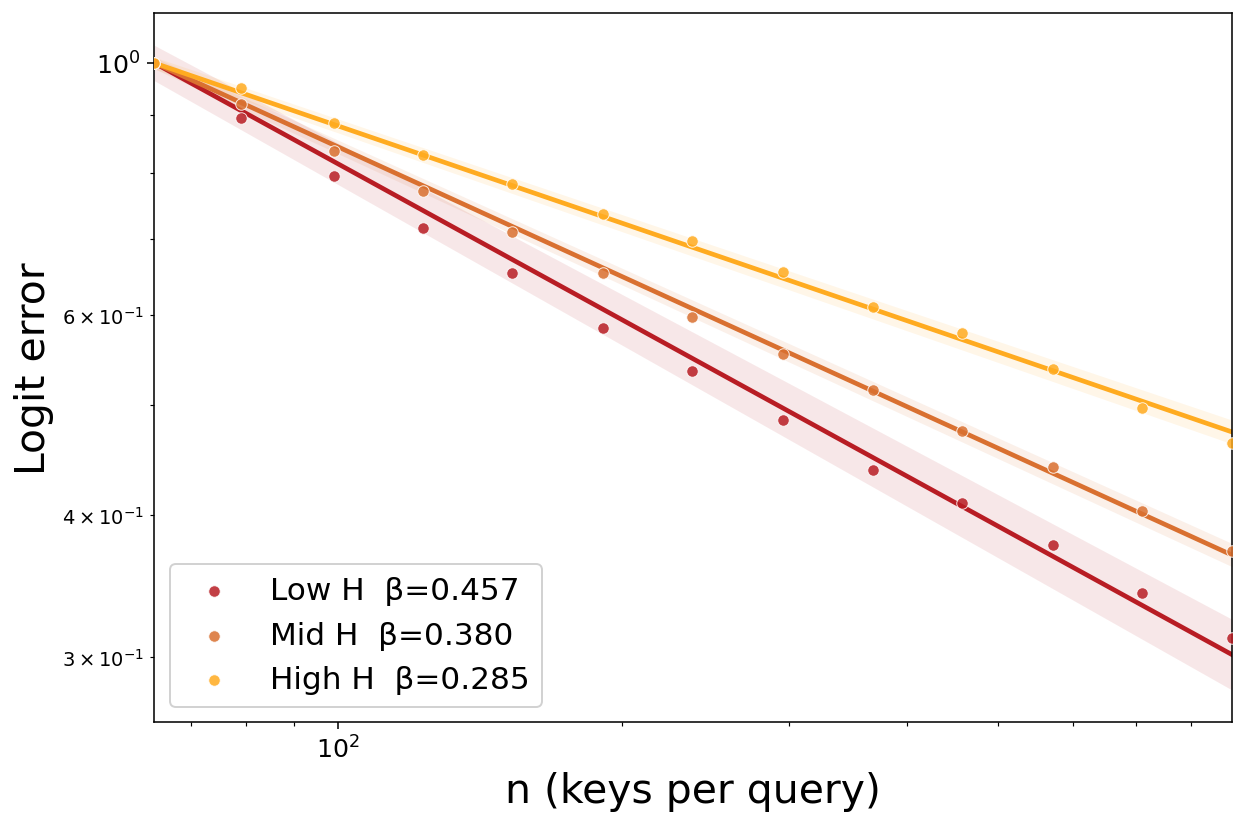}
        \caption{Logit error}
        \label{fig:downstream-logit}
    \end{subfigure}
    \hfill
    \begin{subfigure}[t!]{0.32\textwidth}
        \centering
        \includegraphics[width=\linewidth]{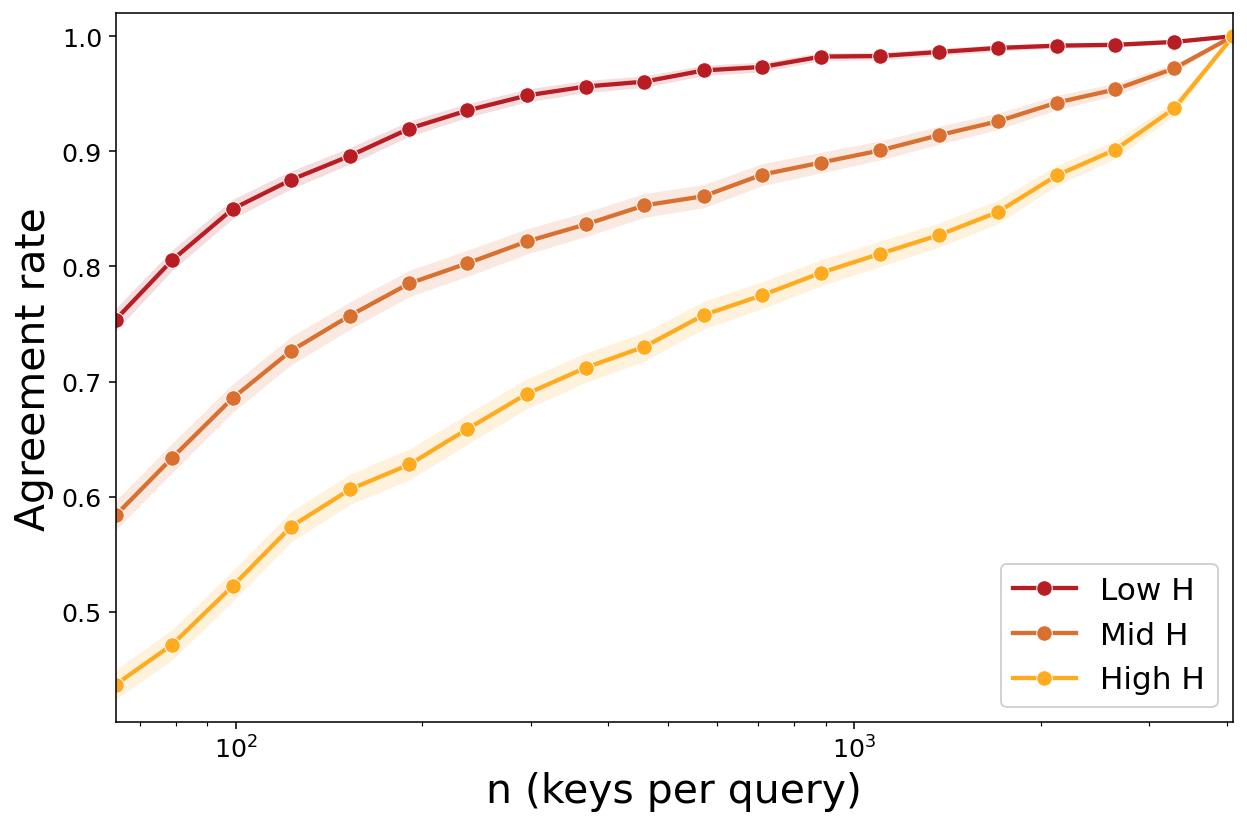}
        \caption{Agreement rate}
        \label{fig:downstream-agreement}
    \end{subfigure}
    \hfill
    \begin{subfigure}[t!]{0.32\textwidth}
        \centering
        \includegraphics[width=\linewidth]{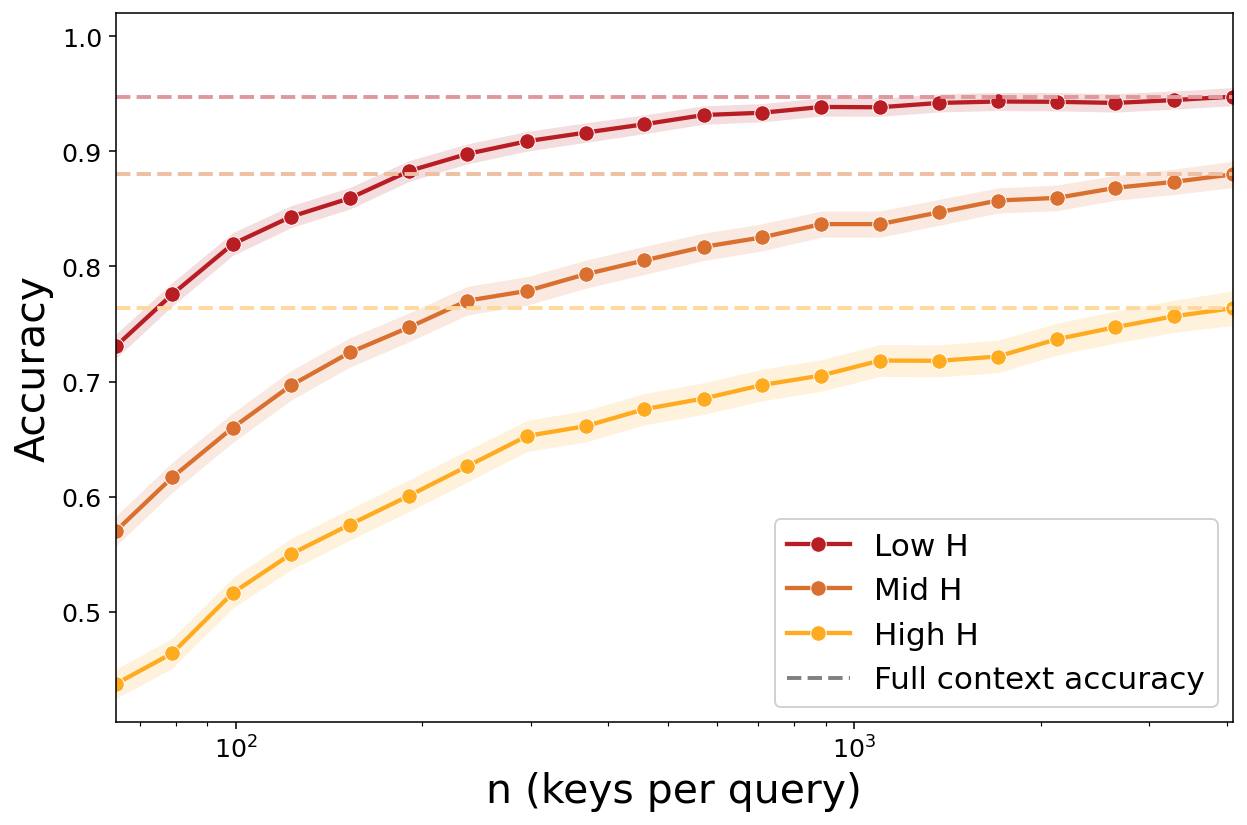}
        \caption{Accuracy}
        \label{fig:downstream-accuracy}
    \end{subfigure}

    \caption{
    \textbf{Downstream token sample complexity.}
    BigBird--RoBERTa--base is fine-tuned on arxiv-classification and evaluated by replacing dense attention with $n$ i.i.d.\ sampled keys per query.
    Curves are stratified by attention horizon $H$.
    Logit error, agreement with the dense model, and classification accuracy all show horizon-dependent convergence and diminishing returns as $n$ increases.
    Shaded bands indicate $\pm 3 \sigma$ over Monte Carlo repetitions.}
    \label{fig:downstream-practical}
\end{figure*}

We formalize the infinite-token limit of a Transformer's self-attention as a continuous operator on probability measures. Following \cite{DeBie19,Pevny19}, we view a sequence of $n$ tokens as an empirical distribution $\nu_n$ that approximates an underlying token distribution $\nu$. \cref{fig:tsne_limit} illustrates this convergence, showing how $\nu_n$ approaches $\nu$ as $n$ grows. We define the attention mechanism in the limit of infinitely many tokens as an operator acting on $\nu$. This measure-theoretic perspective was pioneered by \cite{Sander22} and extended to study Transformers as interacting particle systems \cite{Lu19,Geshkovski24,Geshkovski25,Castin25,Burger25,Bruno25,Zimin25,Chen25}. 

We use this infinite-token limit to formalize \emph{token sample complexity}: the rate at which finite-$n$ attention approaches its infinite-limit counterpart. With compactly supported token distributions, one naturally expects parametric convergence at rate $1/\sqrt{n}$. However, on both synthetic and real data, we observe slower sub-parametric rates $n^{-\beta}$ with $\beta < 1/2$. Our analysis identifies an effective \emph{attention horizon} $H := \|\Sigma^{1/2}A\Sigma^{1/2}\|_2$ as
a key quantity governing this slowdown, where $\Sigma$ is the covariance matrix of the token distribution, and $A := K^\top Q$ is the attention matrix, itself a product of the key $K$ and query $Q$ matrices. It can be interpreted as the largest query-key score attainable by two tokens in the unit ball of the covariance geometry (see Definition \ref{def:spectral_reach_arc}). When it grows large, convergence slows down, and when $H \to +\infty$, softmax attention converges to hardmax, in which case convergence is only logarithmic (see Section~\ref{sec:hardmax}).

Figure~\ref{fig:downstream-practical} empirically connects our token sample complexity
bounds to downstream errors on an end-to-end classification task.
When dense attention is replaced by attention over $n$ sampled keys per query, the sparse
model progressively recovers the dense model's logits and predictions as $n$
increases. Consistent with our theoretical analysis, this recovery is
sub-parametric and visibly structured by the attention horizon: high-horizon
regimes require more keys to approach dense-model
behavior and reach lower accuracy. Thus, the slowdown identified by our analysis at the attention-output level
propagates through the full network to downstream prediction quality. 
% A mettre ici ou plus tard dans la partie expérience
This yields a quantitative inference-time insight: in our experiment,
using only $25\%$ of keys already recovers most of the dense-model accuracy,
while doubling the number of sampled keys beyond this point brings only a small
additional gain (see \ref{sec:downstream_sec}).

More precisely, our key contributions are the following\begin{itemize}[leftmargin=*, nosep]
  \item In Section~\ref{sec:cv_map}, we quantify almost-sure convergence of the \emph{attention map} itself at a given query $x$. Theorem~\ref{thm:uniform_convergence} provides uniform convergence on balls $B_R$ for sub-Gaussian tokens, with explicit quantitative estimation of the rate.
% \item   In Section~\ref{sec:cv_moment}, we analyze moment convergence, relevant for downstream tasks relying on aggregate statistics. For compact support, a straightforward approach yields rates for moments of any Lipschitz function in $C(R)/\sqrt{n}$ with $C(R)$ exponential in radius $R$, becoming uninformative for large $R$. Theorem~\ref{thm:convergence_rate_lipschitz} refines this result for sub-Gaussian tokens by accounting for data anisotropy through the sub-Gaussianity parameter, achieving polynomial (rather than exponential) dependence on scale, with rate $\mathcal{O}(n^{-\beta})$, where $\beta < \tfrac{1}{2}$. This rate depends on attention horizon, which accounts for the spectral properties of the token distribution, and better captures real-data behavior. We complete our analysis by deriving exact asymptotics of convergence rates for infinite horizon (hardmax) attention in the Gaussian case. Proposition~\ref{prop:hardmax} shows that convergence becomes only logarithmic in $n$, of order $1/\sqrt{\ln n}$.
\item In Section~\ref{sec:cv_moment}, we study moment convergence, which matters for downstream tasks based on aggregate token statistics. Compact-support arguments yield $C(R)/\sqrt n$ rates, but with constants exponential in the radius $R$ and thus uninformative for large supports. Theorem~\ref{thm:convergence_rate_lipschitz} gives sharper sub-Gaussian rates $\mathcal{O}(n^{-\beta})$, $\beta<\tfrac12$, with polynomial scale dependence governed by the attention horizon and token anisotropy, matching real-data behavior more closely. We also analyze the Gaussian hardmax limit, where Proposition~\ref{prop:hardmax} gives the exact logarithmic rate $1/\sqrt{\ln n}$.
\item In Section~\ref{sec:experiments}, we empirically assess the tightness of bounds for Gaussian data and demonstrate their practical relevance on real text: using BigBird and BERT models on various long text inputs, we observe convergence rates strictly slower than $1/\sqrt{n}$, confirming that the sub-parametric regime governs attention behavior in practice. Finally, we show on an end-to-end classification task that the slowdown identified by our analysis is reflected in downstream behavior: sparse attention approaches dense inference at horizon-dependent sub-parametric rates.
\end{itemize}

\section{Background and related work}

\textbf{Classical PAC sample complexity.} In statistical learning theory, \emph{sample complexity} refers to the number of training samples $N$ required for a learning algorithm to achieve a probably approximately correct (PAC) solution \cite{ShalevShwartz14}. The classical question is: given i.i.d.\ examples of fixed dimension, how many samples are needed so that, with high probability, the learned model's error is within $\varepsilon$ of the optimum? Classical results show $N$ scales with hypothesis class complexity (VC dimension, Rademacher complexity), but crucially assume each sample has fixed, finite dimension. This is the regime of standard PAC theory: we collect more fixed-size data points (images, sentences) to improve generalization.

\textbf{Token sample complexity (this work).} In our setting, we consider an entirely different kind of complexity question: the model is \emph{already trained}, and we are not collecting new independent samples; instead, we are increasing the number of tokens in a single input sequence. Modern Transformer encoders like BERT process an input as a single sequence of tokens (e.g.\ words or subword tokens in a sentence) of length $n$. Importantly, $n$ is not fixed—the architecture allows variable-length input, and indeed self-attention inherently permits an unbounded number of tokens by design \cite{Vaswani17}. Thus, we can think of $n$ as a parameter of the problem instance, akin to an input dimension that can grow. 

This inference-time perspective aligns with the growing use of extended contexts at inference. In-context learning and chain-of-thought prompting rely on providing Transformers with sequences that are often much longer than those seen during training \cite{Brown20,Zhang23,Sander24,Huang24,Zhang25,Wei22}. As these approaches push the sequence length $n$ beyond training-time regimes, understanding how attention accuracy scales with $n$ at inference becomes increasingly important.

\textbf{Transformers and self-attention.}  
A Transformer relies on attention mechanisms to model interactions between tokens. Each encoder layer applies self-attention, where output vectors $z_i$ are computed as weighted combinations of value vectors based on token affinities (see Definition \ref{def:sh_self_attn}), followed by a feed-forward network. Unlike CNNs or RNNs, self-attention imposes no intrinsic limitation on sequence length $n$: arbitrarily many tokens can be processed. This ability to capture long-range interactions comes at quadratic cost $\mathcal{O}(n^2)$, motivating efficient attention variants.

\textbf{Sparse attention and efficient variants} Our results also give insight into the error created when the input tokens are randomly sub-sampled, which can be viewed as a simple model of attention sparsification. In practice, more advanced sparsification strategies are used, such as restricting tokens to local neighborhoods as in Longformer \cite{Beltagy20} and BigBird \cite{Zaheer20}, or explicitly projecting keys and values to lower-dimensional subspaces as in Linformer \cite{Wang20}. Although our findings do not cover those methods, they offer an initial step toward a theoretical understanding of such approaches.

\textbf{Transformers with arbitrarily many tokens.} The mean field model which corresponds to defining attention as operating over a probability distribution of tokens was initially presented in \cite{DeBie19} and \cite{Pevny19}, and subsequently adapted to Transformers in \cite{Vuckovic20,Sander22,Geshkovski24,Geshkovski25}.
Building on this framework, \cite{Geshkovski24}, \cite{Geshkovski25} and \cite{Castin25} develop a \emph{Transformer PDE} for standard softmax attention and prove it is the mean-field limit of an $n$-particle system as $n\to\infty$.  \cite{Geshkovski25} models tokens as points evolving on the \emph{sphere} $\mathbb{S}^{d-1}$ to capture normalization effects and connects the resulting dynamics to models from mathematical physics. Related developments include \cite{Karagodin24} (clustering with causal masking) and \cite{Bruno25} (meta-stable clustering). Notably, \cite{Castin25} prove that the Transformer PDE preserves Gaussianity, reducing the dynamics to mean and covariance evolution. In contrast, we focus on a single attention layer in the infinite-token limit, using the Gaussian case as a tractable baseline to derive explicit convergence rates and motivate extensions to sub-Gaussian distributions.

\textbf{Relation to \cite{Boursier25}}
In an independent work, \cite{Boursier25} also study token sample complexity. A first key difference with our work is that we consider the setting in which queries, keys, and values are drawn from the same token set, as is typically the case when modeling self-attention. This dependence between tokens makes the analysis substantially more involved, and requires the use of the chaining method in the proof of \cref{thm:uniform_convergence}; see \cref{sec:proof_uniform} for details. We also establish $O(1/n^\beta)$ error bounds with smaller values of $\beta$; in particular, when $A \to 0$ we recover the exponent $\beta = 1/2$, and we derive exact asymptotics in the hardmax limit $A \to \infty$. On the other hand, \cite{Boursier25} establish concentration bounds for the Jacobian of the attention map, with applications to in-context linear regression under Gaussian inputs.

\section{Mathematical preliminaries}

\subsection{Self-attention}
Let $d\in\mathbb{N}$ be the hidden dimension of tokens embeddings that live in $\mathbb{R}^d$, with inner product $\langle\cdot,\cdot\rangle$ and $\|\cdot\|_2$ the Euclidean norm for vectors and the associated operator norm. 
\begin{definition}[Single-head self-attention]
\label{def:sh_self_attn}
Let $k,d\in\mathbb{N}$. Let $Q,K,V\in\mathbb{R}^{k\times d}$ the query, key and value matrices and set $\tilde A:=K^\top Q/\sqrt{k}\in\mathbb{R}^{d\times d}$ and $A:=K^\top Q\in\mathbb{R}^{d\times d}$ (where scaling is absorbed in either $K$ or $Q$). 
Given an $n$-sequence $X=(x_i)_{1 \le i \le n}$, embedded in dimension $d$, the single-head self-attention function $f_n:\,\mathbb{R}^{d }\to \mathbb{R}^{k }$ maps each token $x_i$ of the sequence to
\begin{equation}
\begin{aligned}
f_n(x_i)
= \frac{\sum_{j=1}^n e^{\langle A x_i, x_j\rangle}V x_j}{\sum_{j'=1}^n e^{\langle A x_i, x_{j'}\rangle}},
\end{aligned}
\end{equation}
which corresponds to a softmax operation defined for a vector $w \in \mathbb{R}^d$ by $\textnormal{SoftMax}(w) := \bigl(\exp(w_i)/\sum_{j=1}^n \exp(w_j)\bigr)_{1 \le i \le n},$ and applied here for $w = \bigl(\langle A x_i, x_j\rangle\bigr)_{1 \le i \le n}$.
Note that $f_n$ depends on the entire context $X=(x_i)_{1\le i\le n}$. 
For notational simplicity, we encode this dependence only through the index $n$, 
writing $f_n$ instead of the more precise $f_{(x_i)_{1\le i\le n}}$ or $f_{\nu_n}$
when $X\sim \nu_n$ (see \cref{sec:cv_mmt_subsection} for details on these notations). 
\end{definition}
All our results naturally extend to multihead self-attention (see Definition \ref{def:mh_attn}) and all experiments are conducted in the multihead self-attention setting.
\paragraph{Permutation equivariance and measure modeling.}
Self-attention without positional information is equivariant to reindexing: for any permutation $\sigma$ of $\{1,\dots,n\}$, if $\big(x_{1},\dots,x_{n}\big) \sim \nu_n$, then $\big(x_{\sigma(1)},\dots,x_{\sigma(n)}\big) \sim \nu_n$ and for every token $x$,
\[
f_{(x_i)_{1 \le i \le n}}(x) = f_{(x_{\sigma(i)})_{1 \le i \le n}}(x)
\]
Hence, one may represent the sequence via its empirical distribution $\nu_n \;:=\; \frac{1}{n}\sum_{i=1}^n \delta_{x_i},$
and analyze attention at the level of probability measures. This naturally leads to formalizing the infinite-token limit by viewing attention as an operator on such measures.

\begin{definition}[Mean-field self-attention]
Let $\mathcal P (\mathbb{R}^d)$ be the set of Borel probability measures. For $\nu\in \mathcal P(\mathbb{R}^d)$ and $x\in\mathbb{R}^d$, define 
$$f(x) \;:=\;  \frac{  \int_{\mathbb{R}^d} e^{\langle Ax, y\rangle}\, V \, y \;d\nu(y)}{ \int_{\mathbb{R}^d} e^{\langle Ax, y\rangle}\; d\nu(y)} \;\in\mathbb{R}^k$$
When $\nu=\nu_n$, $f$ is equal to the empirical attention map $f_n$. As for $f_n$, $f$ depends on the context distribution $\nu$. As we fix $\nu$ in our work, we omit this dependence to alleviate notations. 
\end{definition}
\paragraph{Reduction to the centered case.}
All our results are stated for centered sub-Gaussian distributions, but they extend straightforwardly to non-centered sub-Gaussian distributions. The only modification in the main results is that the expectation of the distribution shows up in the bounds (in an explicit and controlled way).

\section{Convergence of the attention map}\label{sec:cv_map}
In this section, we examine how fast $\bigl\| f_n(x) -f(x) \bigr\|_2$ converges to zero. We provide a quantitative uniform convergence rate over all tokens in a ball of radius $R$, highlighting the dependence of the estimated rate on the attention parameters and the token distribution properties. 

A uniform—rather than simply pointwise (see Appendix \ref{app:ptw_cv})—convergence result is required to translate convergence of the attention map into results on moments of the transformed distribution. 

In the self-attention setting, the attention map itself is computed from the same set of tokens on which it is subsequently applied: the queries, keys, and values all originate from a common context. This induces a nontrivial autocorrelation structure, as the randomness governing the attention weights and that of the aggregated outputs are no longer independent. When the context size grows, moment estimates therefore involve aggregating attention outputs across a continuum of queries while the underlying attention operator simultaneously depends on the entire token set. As a consequence, the operator must be controlled uniformly over all queries $x$ in the relevant region of space: a single poorly controlled query can dominate the aggregate and ruin the global behavior of the attention operator across the distribution. In this sense, uniform control becomes indispensable to propagate convergence results to moment-level statements (see Appendix \ref{app:proof_cv_lip} for details).

\subsection{Sub-Gaussian Model}
\label{sec:sub-gaussian}
% In transformer architectures, tokens are normalized before each attention map
% (e.g. via LayerNorm or RMSNorm), and therefore remain bounded. Since
% bounded tokens are sub-Gaussian, the sub-Gaussian framework is a natural
% way to move beyond compact-support assumptions that only control the radius of
% the token cloud, while still including mathematically important cases such as
% Gaussian distributions \cite{Castin25}. More importantly, it introduces the parameter $\Sigma$ which allows
% our convergence bounds to account for the anisotropy of the token
% distribution, leading to more informative bounds in the regimes observed in our
% experiments. To formalize this framework, we rely on the standard characterization of sub-Gaussian random vectors in terms of their log-moment generating function (log-MGF) \cite{Boucheron13}. 
In transformer architectures, normalization layers (e.g. LayerNorm or RMSNorm) keep tokens bounded, hence sub-Gaussian. The sub-Gaussian framework therefore goes beyond compact-support assumptions, while also including important unbounded cases such as Gaussian distributions \cite{Castin25}. More importantly, its parameter $\Sigma$ captures token anisotropy, yielding more informative convergence bounds in the regimes observed experimentally. We use the standard log-MGF characterization of sub-Gaussian random vectors \cite{Boucheron13}.

\begin{definition}[Sub-Gaussian vector]\label{def:subGaussian}
A centered random vector $X\in\mathbb{R}^d$ with distribution $\nu$ is \emph{sub-Gaussian with parameter matrix} $\Sigma\succ0$ if
$\forall t\in\mathbb{R}^d,
\mathbb{E}\!\left[\exp\{\langle t,X\rangle\}\right] \;\le\; \exp\!\left(\tfrac{1}{2}\, t^\top \Sigma\, t\right)
$.
\end{definition}
If no prior information is available aside from a compact support $\operatorname{supp}(\nu) \subset B(0,R)$, one can take $\Sigma=R^2 \mathrm{Id}$.

\subsection{Uniform Convergence Result}
Establishing an explicit quantitative estimation of the uniform convergence (Theorem \ref{thm:uniform_convergence}) is  challenging. First, the attention map is a ratio of empirical averages whose kernel
$e^{\langle Ax, y\rangle}$ is not uniformly Lipschitz, leading to a function
class $\mathcal{F} = \{y \mapsto y e^{\langle Ax, y\rangle}, x \in B_R\}$ with
unbounded envelope $F(y) = \|y\|_2 e^{R\|A^\top y\|_2}$, outside the scope of
standard empirical process theory. 

Furthermore, to obtain explicit parameter-dependent bounds, we use Dudley's entropy integral and covering
estimates, with a duality argument to handle the vector-valued empirical
process. 
% A further difficulty is that the empirical denominator can approach zero, precluding
% any uniform delta-method argument: numerator and denominator must be controlled
% separately and then recombined uniformly in $x$.

Theorem~4.2 quantifies how the interaction between the token distribution geometry $\Sigma$ and the model parameter $A$ influences the convergence of the attention map. 
The key novelty and technical challenge of this result lies in capturing the behavior of SoftMax mappings, which are not uniformly Lipschitz. This prevents the use of standard uniform central limit theorems and requires a dedicated analysis.

\begin{theorem}[Uniform convergence of the attention map for sub-Gaussian tokens]\label{thm:uniform_convergence}
Let $X \sim \nu$ be centered and sub-Gaussian with parameter matrix $\Sigma \succ0$. For any $R > 0$, $\delta > 0$, there exists a constant $C>0$, such that for $n \ge n_{min}(\delta, \Sigma, A, R) := 4e^{2 R^2 \|\Sigma^{1/2}A\|_2^2}\, \Bigl(\frac{1}{\delta} - 1\Bigr)$, with probability at least $1 - \delta$,
\begin{equation}\label{eq:uniform_convergence}
  \sup_{x \in B_R}\bigl\| f_n(x) -f(x) \bigr\|_2
\;\le\;  q_{(\Sigma, A, V, R, d, \delta)} \cdot \frac{e^{8 \,R^2\|\Sigma^{1/2}A\|_2^2}}{\sqrt{n}} ,
\end{equation}
where $q_{(\Sigma, A, V, R, d, \delta)}:= \frac{C \, \sqrt{d}}{\delta} \|V\|_2  \cdot \|\Sigma\|_2^{1/2} \cdot(R \|A\|_2 \|\Sigma\|_2^{1/2}  d + \operatorname{tr}(\Sigma)^{1/2} \|A\|_2 + 5^{d/2}\sqrt{d}).$ 
\end{theorem}
See Appendix \ref{app:proof_thm_unif_cv} for full proof, and Appendix \ref{app:ptwise_gaussian} for a specific estimate for Gaussians. Note that these rates are not sharp in full generality. For instance, in low dimension, faster asymptotic rates are expected; Appendix \ref{app:diracs} shows $1/\sqrt{n}$ rates in the one-dimensional case. 

\subsection{Sketch of proof of \cref{thm:uniform_convergence}}\label{sec:proof_uniform}
% The proof of Theorem~\ref{thm:uniform_convergence} proceeds by controlling the numerator and denominator of the attention ratio separately, and then combining these bounds to form the ratio. We first establish a result in expectation and subsequently derive high-probability bounds.
The proof of Theorem~\ref{thm:uniform_convergence} controls the numerator and denominator separately before recombining them uniformly in $x$, since the empirical denominator may approach zero and precludes a direct uniform delta-method argument. We first prove an expectation bound, then derive high-probability bounds.

Since the numerator and denominator follow the same proof structure, we detail only the numerator. The only additional difficulty is its vector-valued nature ($\mathbb{R}^d$), requiring a technical adaptation of Dudley's theorem to the multidimensional setting (see Step~2).

We reformulate the problem as the uniform convergence of a supremum over a parametric class of
\emph{unbounded Lipschitz functions}, where the parameter lies in a compact set. Define the function class $\mathcal{F} := \{ f_x : x \in B_R \},$ where $
f_x : y \mapsto y \, e^{\langle Ax, y \rangle}
\in L^2(\nu; \mathbb{R}^d)$ (see \cref{sec:empirical_proc_tools} for empirical process theory's details). 
Our goal is to bound the uniform deviation $\|\mathbb{P}_n - \mathbb{P}\|_{\mathcal{F}}
:= \sup_{x \in B_R}
\Big\|
\frac{1}{n}\sum_{i=1}^n f_x(Y_i)
- \mathbb{E}[f_x(Y)]
\Big\|_2.$
The proof of this intermediate result proceeds in two steps.

\paragraph{Step 1: Symmetrization.}
We first relate $\mathbb{E}[\|\mathbb{P}_n - \mathbb{P}\|_{\mathcal{F}}]$ to the Rademacher complexity
of $\mathcal{F}$. Introducing an i.i.d.\ ghost sample $(X_1,\ldots,X_n)$ and using Jensen’s inequality,
one shows that $\mathbb{E}[\|\mathbb{P}_n - \mathbb{P}\|_{\mathcal{F}}]
\leq 2\,\mathfrak{R}_n(\mathcal{F}),$ where $$\mathfrak{R}_n(\mathcal{F})
:= \mathbb{E}\!\Big[
\sup_{x \in B_R}
\big\|
\frac{1}{n}\sum_{i=1}^n \varepsilon_i f_x(Y_i)
\big\|_2
\Big].$$
\paragraph{Step 2: Bounding the Rademacher complexity.}
To bound $\mathfrak{R}_n(\mathcal{F})$, we apply Dudley’s entropy integral result  (\cref{thm:dudley}). Since the process
$
Z_x := \frac{1}{n}\sum_{i=1}^n \varepsilon_i f_x(Y_i)
$
is vector-valued, we use the dual norm representation
$
\|Z_x\|_2 = \sup_{z \in B_1} \langle z, Z_x \rangle,
$
and consider the scalar process
$
Z_{(z,x)} := \langle z, Z_x \rangle,
$ for $ (z,x) \in \mathcal{G} := B_1 \times B_R$.

One verifies that $Z_{(z,x)}$ is sub-Gaussian with respect to an appropriate empirical metric
$d_{\mathcal{G}}$ on $\mathcal{G}$. Dudley’s theorem then yields
\[
\mathbb{E}_{\varepsilon}
\left[
\sup_{(z,x) \in \mathcal{G}} Z_{(z,x)}
\right]
\;\lesssim\;
\frac{1}{\sqrt{n}}
\int_0^{\mathrm{diam}(\mathcal{G})}
\sqrt{\ln \mathcal{N}(\delta, \mathcal{G}, d_{\mathcal{G}})} \, d\delta.
\]
To evaluate this integral, we bound three key quantities: the moments of the envelope function~$F$ (Lemma \ref{lem:bound_envelope}),
the Lipschitz constant of functions in~$\mathcal{F}$, (Lemmas \ref{lem:bound_lipschitz1} and \ref{lem:bound_lipschitz2}), and
the covering number $\mathcal N(\varepsilon,  \mathcal G, d_{\mathcal G}((z_1,x_1),(z_2,x_2)))$ of~$\mathcal{G}$ (Lemmas \ref{lem:cov_number1} and \ref{lem:cov_number2}).

Combining these bounds and integrating yields a control of
$\mathbb{E}[\|\mathbb{P}_n - \mathbb{P}\|_{\mathcal{F}}]$.
A direct application of Markov’s inequality then converts this expectation bound into the
high-probability result of Proposition~\ref{prop:numerator}.

Applying Steps~1--2 to the scalar function class
$
\mathcal{F}' := \{ h_x : y \mapsto e^{\langle Ax, y \rangle} \}
$
yields Proposition~\ref{prop:denominator}, with a simpler analysis in dimension~$1$.

\paragraph{Step 3: Combining the bounds.}
Write $N_n$ and $D_n$ for the empirical numerator and denominator, and $N$ and $D$ for their
expectations. The ratio error decomposes as
\[
\left\|
\frac{N_n}{D_n} - \frac{N}{D}
\right\|_2
\leq
\frac{\|N_n - N\|_2}{D_n}
+
\frac{\|N\|_2 \, |D_n - D|}{D\,D_n}.
\]
Applying Jensen’s inequality, $D = \mathbb{E}[e^{\langle Ax, Y \rangle}] \geq 1$.
Then, a control the deviation of $D_n$ around $D$ along with the application of Propositions~\ref{prop:numerator} and~\ref{prop:denominator} concludes the proof.

\section{Convergence of the transformed distribution}\label{sec:cv_moment}

\subsection{Moments convergence}\label{sec:cv_mmt_subsection}
So far, we have analyzed the convergence of the attention map $\|f_n(x)-f(x)\|_2$ at a single query $x$ in a fixed compact set. However, it is also important to study the convergence of the output distribution itself, which corresponds to the self-attention setting (where the attention map is applied to the tokens themselves).
Let $X$ (resp.\ $\hat X$) be a random vector with distribution $\nu$ (resp.\ $\nu_n$). We therefore consider the convergence of the distribution $(f_n)_\sharp \nu_n$ of $f_n(\hat X)$ toward the distribution $f_\sharp \nu$ of $f(X)$, where $f_\sharp$ denotes the push-forward operator.
The convergence of the output distribution is particularly relevant for downstream tasks that rely on aggregate token statistics, such as classification or sequence-level predictions. Moreover, in deep transformer architectures, attention layers are stacked, so that the inputs to subsequent layers depend on the entire reweighted distribution produced by earlier attention mechanisms. Understanding distributional convergence is therefore crucial for analyzing error propagation across network depth.

% A natural way to quantify this convergence is through moments such as the mean, second-order statistics and the mean squared error (MSE). 
% %
% Moments capture aggregate properties of attention outputs and correspond to operations used in downstream tasks such as average pooling, token aggregation, or sequence-level representations. 
A natural way to quantify this convergence is through moments, such as the mean, second-order statistics, and MSE, which capture aggregate properties of attention outputs used in downstream operations like average pooling, token aggregation, and sequence-level representations.
We consider moments defined as expectations of functions $h:\mathbb{R}^d\to\mathbb{R}^{d'}$ applied to the attention outputs, and denote by $\mathbb{E}_n[h\circ f_n(\hat X)]$ (resp.\ $\mathbb{E}[h\circ f(X)]$) the empirical (resp.\ exact) moments.
Taking $h(x)=x$ recovers the mean; more generally, we focus on Lipschitz functions $h$, useful for bounding Fourier moments which is equivalent to controlling translation-invariant maximum mean discrepancies (MMD) \cite{gretton2012kernel}. We also extend this to quadratic functions to capture covariance. 

% In addition to moment convergence, Corollary \ref{coro:mse_convergence_rate_lipschitz} broadens the applicability of our results by controlling the mean squared error
% $\mathbb{E}\Bigl[
% \bigl\|f_n(X)-f(X)\bigr\|_2^2
% \Bigr]$ which can directly be used to assess the quality of downstream tasks relying on each individual token embedding. Finally, our theoretical findings persist in deep architectures and is extended to compositions of layers in Corollary \ref{coro:multi_layer} in Appendix. 
Corollary~\ref{coro:mse_convergence_rate_lipschitz} broadens our results by controlling the MSE
$\mathbb{E}\|f_n(X)-f(X)\|_2^2$, which measures errors at the level of individual token embeddings. Corollary~\ref{coro:multi_layer} further extends the theory to deep compositions of layers.
% Provided that the subsampling remains i.i.d.\ at each layer, local sketching errors accumulate through the network, with amplification governed by the Lipschitz constants of downstream attention layers.

\subsection{Sub-Gaussian Moment Convergence}

For compactly supported token distributions ($\mathrm{supp}(\nu) \subset B(0,R)$) our uniform pointwise convergence result (Theorem~\ref{thm:uniform_convergence}) directly implies analogous rates of order $O(C(R)/\sqrt{n})$ for the convergence of distribution moments. However, the resulting constant $C(R)$ grows exponentially with $R$, which makes these bounds essentially useless in practice. Moreover, this analysis does not apply to the refined sub-Gaussian model introduced in Section~\ref{sec:sub-gaussian}, and therefore cannot account for the anisotropy of the token distribution encoded by $\Sigma$.
We therefore pursue a more refined approach, aiming to obtain slower rates of the form $O(C'/n^\beta)$ with $\beta<1/2$, but where the constant $C'$ depends only mildly on $\Sigma$. As illustrated by our numerical experiments, these slower rates are consistent with empirical observations.
The rate $\beta$ depends on both the attention parameter and the token anisotropy $\Sigma$ through the attention horizon $H$ (see Definition \ref{def:spectral_reach_arc} below). In the weaker setting where one only assumes compact support of the token
distribution, we derive analogous rates in Appendix~\ref{app:compact_lip_cv}.

\begin{definition}[Token-parameters coupling coefficient]\label{def:spectral_reach_arc}
Let $X$ be centered and sub-Gaussian with parameter matrix $\Sigma\succ0$. Recall $A = K^{\top}Q$, 
The \textit{Attention Horizon} $H$ is defined as 
$$
    H \;:=\; \bigl\|\Sigma^{1/2}A\Sigma^{1/2}\bigr\|_2,
$$
 where $\|\cdot\|_2$ denotes the operator norm.
 \end{definition}
\begin{remark}[Intuition for attention horizon $H$]
Since the tokens have covariance \(\Sigma\), their natural geometry is the Mahalanobis geometry
$\|x\|_{\Sigma^{-1}}^2=x^\top\Sigma^{-1}x$.
This motivates the definition of the horizon, which can be interpreted as the largest possible query-key score between two tokens in the unit ball of the covariance geometry:
\[
H=
\max_{\|x\|_{\Sigma^{-1}}\le 1,\ \|y\|_{\Sigma^{-1}}\le 1}
\langle Qx,Ky\rangle.
\]
Intuitively, the horizon $H$ quantifies the effective range of attention: how distant token keys can influence a given query, as determined by token geometry ($\Sigma$) and attention weights ($A$). 
\end{remark}

Theorem~5.3 establishes that the non-asymptotic convergence rate of attention moments is sub-parametric, with an exponent $\beta < 1/2$ that depends on $H$. Its main contribution is to provide, for the first time,  a precise and quantitative characterization of how sparse attention mechanisms become harder to approximate with a limited number of tokens.
The resulting bounds are not only tight on synthetic data, but are also supported by experiments on real data, where we show that $H$ governs the approximation rate even for end-to-end downstream tasks. This yields concrete and quantitative insight into the number of tokens required per layer to achieve a desired level of accuracy.

\begin{theorem}[Lipschitz functional convergence rate for sub-Gaussian tokens]\label{thm:convergence_rate_lipschitz}
% Let $X \sim \nu$ and $\hat X \sim \nu_n$ be centered and sub-Gaussian with parameter matrix $\Sigma\succ0$ and define $H$ as above. 
Let $X\sim\nu$ be centered and sub-Gaussian with parameter matrix $\Sigma\succ0$, and let $\nu_n$ be the empirical measure associated with $n$ i.i.d. samples from $\nu$. Denote $\hat X\sim\nu_n$.
Let $h$ be a $L_0$-Lipschitz function, squared integrable with respect to $\nu$. Let us denote by $\mathbb{E}$ the expectation w.r.t. $\nu$, and $\mathbb{E}_n$ the expectation w.r.t. $\nu_n$.  For $n$ i.i.d.\ tokens, with $ n\geq 4\Bigl(\frac1\delta -1\Bigr)n^{1/8} $, with probability at least $1 - \delta$,
\[
\bigl\|\mathbb{E}_n[h \circ f_n(\hat X)] - \mathbb{E} [h \circ f(X)]\bigr\|_2
\;\le\;
\frac{P(\sqrt{\ln n}, L_0) }{ n^{\beta}},
\]
where $P$ is a polynomial function, and $\beta := \frac{1}{2(1+32 \cdot H^2)}$
\end{theorem}

\begin{remark}[Sharpness of the slow-rate result]
% A natural question is to understand when this slow-rate upper bound is sharp. 
Experiments confirm the predicted horizon-dependent slowdown; in the hardmax regime, Section~\ref{sec:hardmax} proves tight asymptotics with an even stronger slowdown.
% The experiments in Section~\ref{sec:experiments} show that slow rates occur across a wide range of attention maps and token distributions, and match the predicted dependence on the attention horizon. In the high-horizon regime, where attention approaches a hardmax map, Section~\ref{sec:hardmax} further establishes tight asymptotics with an extreme slowdown.
%
\end{remark}
See Appendix~\ref{app:proof_cv_lip} for proof, and Appendix \cref{app:iid_assump} for a discussion on the i.i.d. hypothesis. 

\subsection{Sketch of proof for \cref{thm:convergence_rate_lipschitz}}

\subparagraph{Step 1: Error decomposition.}
By the triangle inequality, we decompose the error into two terms in order to bound each separately
\[
\bigl\|\mathbb{E}_n[h\circ f_n(\hat X)]-\mathbb{E}[h\circ f(X)]\bigr\|_2
\le \mathcal{I}+\mathcal{J},
\]
\[
\text{where} \quad
\mathcal{I}
:=\bigl\|\mathbb{E}_n\!\big[h\circ f_n(\hat X)-h\circ f(X)\big]\bigr\|_2,
\] \[
\mathcal{J}:=\bigl\|\mathbb{E}_n[h\circ f(X)]-\mathbb{E}[h\circ f(X)]\bigr\|_2 .
\]

The term $\mathcal{J}$ captures the sampling error of the population attention, while $\mathcal{I}$ captures the approximation error from using finite context. The term $\mathcal{J}$ is straightforward to bound, so we will focus on establishing the bound for $\mathcal{I}$.

\subparagraph{Step 2: Bounding $\mathcal{I}$ (approximation error).}
% The key idea is that sub-Gaussian concentration implies that in practice the norms of the tokens concentrate below an
% \emph{effective radius} that grows slowly with the sample size $n$ and with the scale of the attention parameters. In particular, large-norm tokens are exponentially rare.
The key idea is that sub-Gaussian concentration confines most tokens to an \emph{effective radius}, slowly growing with $n$ and the attention scale, while large-norm tokens are exponentially rare.

% We split the sum according to whether $\|Bx_i\|_2$ exceeds a threshold $R$, where $B$ is an auxiliary invertible matrix which we will optimize later on :
% $
% \mathcal{I} \leq \mathcal{I}_1 + \mathcal{I}_2 + \mathcal{I}_3,
% $
% where $\mathcal{I}_1$ sums over $\|Bx_i\| < R$, and $\mathcal{I}_2$, $\mathcal{I}_3$ handle the tails for $f_n$ and $f$ respectively.
% %
% For $\|Bx_i\| < R$, the Lipschitz property of $h$ and the uniform convergence of the Theorem~\ref{thm:uniform_convergence} generalization Theorem~\ref{thm:uniform_cv_B} over the compact set $K_B := \text{Im }B  \cap B_R$ yield
We split the error between tokens inside the ellipsoid $\|Bx_i\|_2\le R$ and those outside it, where $B$ is an auxiliary invertible matrix optimized later and get: $\mathcal{I}\le \mathcal{I}_1+\mathcal{I}_2+\mathcal{I}_3.$

Here $\mathcal{I}_1$ is the inside contribution, while $\mathcal{I}_2$ and $\mathcal{I}_3$ control the tail contributions for $f_n$ and $f$, respectively. On $K_B(R):=\{x:\|Bx\|_2\le R\}$, the Lipschitz property of $h$ and Lemma \ref{thm:uniform_cv_B} yield
     \[
    \mathcal{I}_1 \leq L_0 \cdot \sup_{x \in K_B} \|f_n(x) - f(x)\|_2 = O\bigl(\frac{e^{8R^2 \|\Sigma^{1/2}AB^{-1}\|_2^2}}{\sqrt{n}} \bigr).
    \]
% For $\|Bx_i\| > R$, the empirical attention map satisfies $\|f_n(x)\|_2 \leq M_n := \max_i \|x_i\|_2$ since $f_n$ is a convex combination of the tokens embeddings. A similar bound holds for $f$ and is derived in  \cref{lem:growth_gamma_sub} from the importance-sampling form of the attention operator. 
% We then control the proportion $N_R/n$ of tokens with norm exceeding $R$ using Hoeffding's inequality (\cref{eq:hoeffding}), and 
% combine this with a union bound on $M_n$ based on
%  sub-Gaussian tail estimates (see \cref{prop:tail_control} for details), which gives
%     $
% \mathcal{I}_2 + \mathcal{I}_3 = O\bigl(e^{-cR^2/\|\Sigma^{1/2}B^\top\|_2}\bigr).
%     $
For the tail terms $\|Bx_i\|_2>R$, the attention map satisfies $\|f_n(x)\|_2\le M_n:=\max_i\|x_i\|_2$, since $f_n$ is a convex combination of token embeddings; the analogous bound for $f$ follows from the importance-sampling form of attention in Lemma \ref{lem:growth_gamma_sub}. Hoeffding's inequality controls the tail proportion $N_R/n$, while a union bound with sub-Gaussian tails controls $M_n$ (see Proposition \ref{prop:tail_control}), yielding $\mathcal{I}_2+\mathcal{I}_3
=O\!\left(e^{-cR^2/\|\Sigma^{1/2}B^\top\|_2}\right).$

\subparagraph{Step 3: Optimizing over $R$.}
% The previous step yields a decomposition of the approximation error $\mathcal{I}$ into two competing
% contributions:  $\mathcal{I}_1$, controlled by uniform convergence on $B_R$, which grows
% exponentially in $R^2$, and the tail terms $\mathcal{I}_2$ and  $\mathcal{I}_3$, arising from large-norm tokens, which decays
% exponentially fast in $R^2$ due to sub-Gaussian concentration.
% Balancing these contributions by choosing
% $
% R^\star
% = \Theta\!\Big(
% \sqrt{\frac{\ln n}{\|\Sigma^{1/2}A B^{-1}\|_2^2+\|\Sigma^{1/2}B^\top\|_2^{-1}}}
% \Big)
% $, and optimizing over all invertible matrices $B$ yields the final rate. See Appendix \ref{app:proof_cv_lip} for full proof.
Thus, $\mathcal{I}$ is bounded by two competing terms: the compact-region term $\mathcal{I}_1$, which grows like $\exp(cR^2)$ through uniform convergence, and the tail terms $\mathcal{I}_2+\mathcal{I}_3$, which decay like $\exp(-cR^2)$ by sub-Gaussian concentration. Balancing them with $R^\star
= \Theta\!\left(
\sqrt{\frac{\ln n}{\|\Sigma^{1/2}A B^{-1}\|_2^2+\|\Sigma^{1/2}B^\top\|_2^{-1}}}
\right)$
and optimizing over invertible $B$ gives the final rate.
% \begin{remark}[The i.i.d. assumption] Having tokens drawn from a distribution with Markovian dependencies does not contradict our theory: as long as tokens are sampled i.i.d. from that distribution — which is the case in our real-data experiments — the theorem applies directly. Nonetheless, in practice, subsampling is often structured, e.g. via windowing, and in that case the i.i.d. assumption no longer holds. Obtaining theoretical guarantees in this setting is significantly harder, which motivates our simplifying assumption. We addressed this concern empirically by conducting an additional experiment using windowing-based subsampling on BigBird (see \cref{sec:experiments}), and observe the same slowdown in convergence, suggesting that our theoretical predictions remain informative beyond the i.i.d. setting. \end{remark}

\subsection{Hardmax regime}\label{sec:hardmax}

% To complete our analysis, we study the case of an infinite horizon regime (the ``hardmax'' attention), and provide an exact asymptotic bound in that case. 
% This limit case is practically relevant because, in real Transformer architectures, matrix norms tend to be significant, creating conditions similar to this limiting regime. 
% %
% In this section, we consider a simplified setting of Gaussian distribution in dimension $d=1$. The key insight is that in the wide horizon regime (where $\|A \cdot \sigma\|_2 \gg 1$), the softmax attention mechanism behaves essentially like a hard-max selector ($A=\infty$), concentrating attention weights on extreme token values.  As a consequence, the empirical attention output reduces to a combination of the sample \emph{maximum} and \emph{minimum}. This phenomenon leads to a remarkable cancellation effect, and the convergence rate is governed by classical Gaussian extreme value theory.
To complete the analysis, we study the infinite-horizon, or hardmax, regime and derive exact asymptotics. This limit is relevant to real Transformers, where large matrix norms can make softmax behave like a hard-max selector. In the simplified one-dimensional Gaussian case, when $\|A\sigma\|_2\gg1$, attention concentrates on extreme tokens, so the empirical output reduces to a combination of the sample maximum and minimum. This phenomenon leads to a remarkable cancellation effect that makes the convergence rate governed by Gaussian extreme value theory.

Let $\bar{f}_n:=\frac{1}{n}\sum_{i=1}^{n}f_n(X_i) $, the empirical mean of the attention outputs.

\begin{proposition}[Hardmax convergence rate]\label{prop:hardmax}
As $n\to\infty$, taking the expectation w.r.t. $X_1,..., X_n:$ 
$$\mathbb{E}[|\bar{f}_n|] = \frac{\ln(4)\,\sigma}{2 \sqrt{2 \ln n}} [1+o(1)], 
\mathbb{E}[\bar{f}_n^2] = \frac{\pi^2 \,\sigma^2}{24 \ln n} [1+o(1)]$$
Therefore, the expectation decreases exactly at the rate $\sigma/\sqrt{\ln n}$.
\end{proposition}

\paragraph{Cancellation phenomenon}
 In the wide-horizon regime, softmax behaves like a hard maximum that focuses on extremes. For Gaussian inputs, the maximum and minimum are asymptotically $\pm \sqrt{2\ln n}$, so the mechanism selects either one with roughly equal probability as $n$ grows, and their contributions cancel, yielding a mean near zero. This symmetry-driven cancellation is specifically Gaussian and implies a slow decay of order $\sigma/\sqrt{\ln n}$, much slower than any fixed polynomial rate. Observe that real token embeddings (e.g., in BigBird) do not display the above saturation. Our interpretation is that their distributions can easily deviate from Gaussian symmetry through skewness, kurtosis, anisotropy, or dependence for example. These departures disrupt the tail balance, removing the cancellation and the characteristic $1/\sqrt{\ln n}$ behavior.

\begin{remark}[Heavy-tailed distributions]
The sub-Gaussian assumption is necessary for hardmax attention to converge;
Appendix~\ref{app:heavy_tail} gives a heavy-tailed counterexample where
convergence fails.
\end{remark}
% \begin{remark}[Heavy-tailed distributions]
% The sub-Gaussian assumption is necessary for the hardmax attention to convergence. Indeed, for heavy-tailed distributions, we can exhibit counter-examples where it does not converge. The key idea follows from the same decomposition as in the proof of \cref{prop:hardmax}: the empirical attention mean splits into a fluctuation term $\frac{M_n^+ - M_n^-}{\sqrt{n}}$
% and a mean-shift term $\frac{M_n^+ + M_n^-}{2}.$
% In the Gaussian case, symmetry forces the latter to zero. For heavy-tailed distributions, this term fails to converge, as we exhibit in Appendix \ref{app:heavy_tail}.
% \end{remark}

\section{Experiments}\label{sec:experiments}
% In this section, we first test the tightness of our rates on Gaussian data, then study the slowdown on real tokens with BigBird--RoBERTa--base \cite{Zaheer20} and BERT. Finally, we go beyond the attention-output level with an end-to-end classification task, which reveals that the slowdown propagates to downstream quality. Our code is available at: https://github.com/leabbt/token-sample-complexity.git.
We validate our rates on Gaussian data, real tokens from BigBird--RoBERTa--base \cite{Zaheer20} and BERT, and an end-to-end classification task showing that the slowdown propagates to downstream quality. Code is available at \url{https://github.com/leabbt/token-sample-complexity}. We use the following experimental protocol common to all experiments. See Appendix \cref{app:xp_details} for additional details.

\paragraph{Token distribution $\nu$}
% We study the convergence of the empirical mean and covariance of attention outputs, computed over a subsample of tokens, toward a limiting attention distribution. For synthetic datasets, this limiting distribution $\nu$ is known. For token distributions arising from real datasets, we model $\nu$ as a discrete distribution of the form
% $\nu = \sum_{i=1}^{N_{\max}} \delta_{x_i}$, with a large number $N_{\max}$ of tokens. The empirical distribution $\nu_n$ is then obtained by drawing $n \ll N_{\max}$ i.i.d.\ tokens from the dataset.
We study how the empirical mean and covariance of attention outputs, computed from $n$ sampled tokens, converge to their limiting values. For synthetic data, the limiting distribution $\nu$ is known; for real datasets, we approximate it by the discrete distribution $\nu=\sum_{i=1}^{N_{\max}}\delta_{x_i}$ over a large token set, and form $\nu_n$ by sampling $n\ll N_{\max}$ i.i.d.\ tokens.

\paragraph{Convergence rate computation}
% To study convergence toward the limiting attention distribution, we uniformly subsample $n$ tokens from $\nu$ and compute the corresponding attention outputs. We then quantify how the empirical mean and covariance of these $n$ outputs converge to their limiting values as $n$ increases. This procedure is implemented as a Monte Carlo experiment, where $n$ ranges over $[n_{\min}, n_{\max}]$. To ensure that $\nu$ provides an accurate approximation of a continuous distribution, we require $N_{\max} \gg n_{\max}$, while $n_{\max}$ must be sufficiently large for convergence to be observable. The convergence rate is estimated by fitting a linear regression of the error against $n$ on a log--log scale.
Our Monte Carlo protocol estimates how fast finite-token attention approaches its limiting distribution. For each subsample size $n\in[n_{\min},n_{\max}]$, we uniformly draw $n$ tokens from $\nu$, compute the corresponding attention outputs, and measure the errors of their empirical mean and covariance relative to the limiting values. We take $N_{\max}\gg n_{\max}$ so that $\nu$ accurately approximates the limit, while $n_{\max}$ is large enough for convergence to be visible. Rates are obtained by fitting the error against $n$ on a log--log scale.

\subsection{Tight bounds for Gaussian data}
% We conduct synthetic experiments in which the limiting token distribution is taken to be $\nu= \mathcal{N}(0,\Sigma)$. For a fixed covariance matrix $\Sigma$, we estimate the empirical convergence rate $\hat{\beta}$ from Monte Carlo simulations and compare it to the theoretical prediction
% for Gaussian data (Proposition \ref{prop:Gaussian_convergence_rate}), $\beta_{\mathrm{th}} = \frac{1}{2(1+H^2)},$ with $H := \|\Sigma^{1/2} A \Sigma^{1/2}\|_2$
% across varying horizon lengths $H$.
For synthetic experiments, we take the limiting token distribution to be $\nu=\mathcal{N}(0,\Sigma)$. For fixed $\Sigma$, we estimate the empirical rate $\hat\beta$ by Monte Carlo simulations and compare it, across varying horizons $H:=\|\Sigma^{1/2}A\Sigma^{1/2}\|_2$, to the Gaussian prediction
$\beta_{\mathrm{th}}=\frac{1}{2(1+H^2)}$ from Proposition~\ref{prop:Gaussian_convergence_rate}.

\begin{figure}[ht]
  \vskip -0.08in
  \begin{center}
    \centerline{\includegraphics[width=\columnwidth]{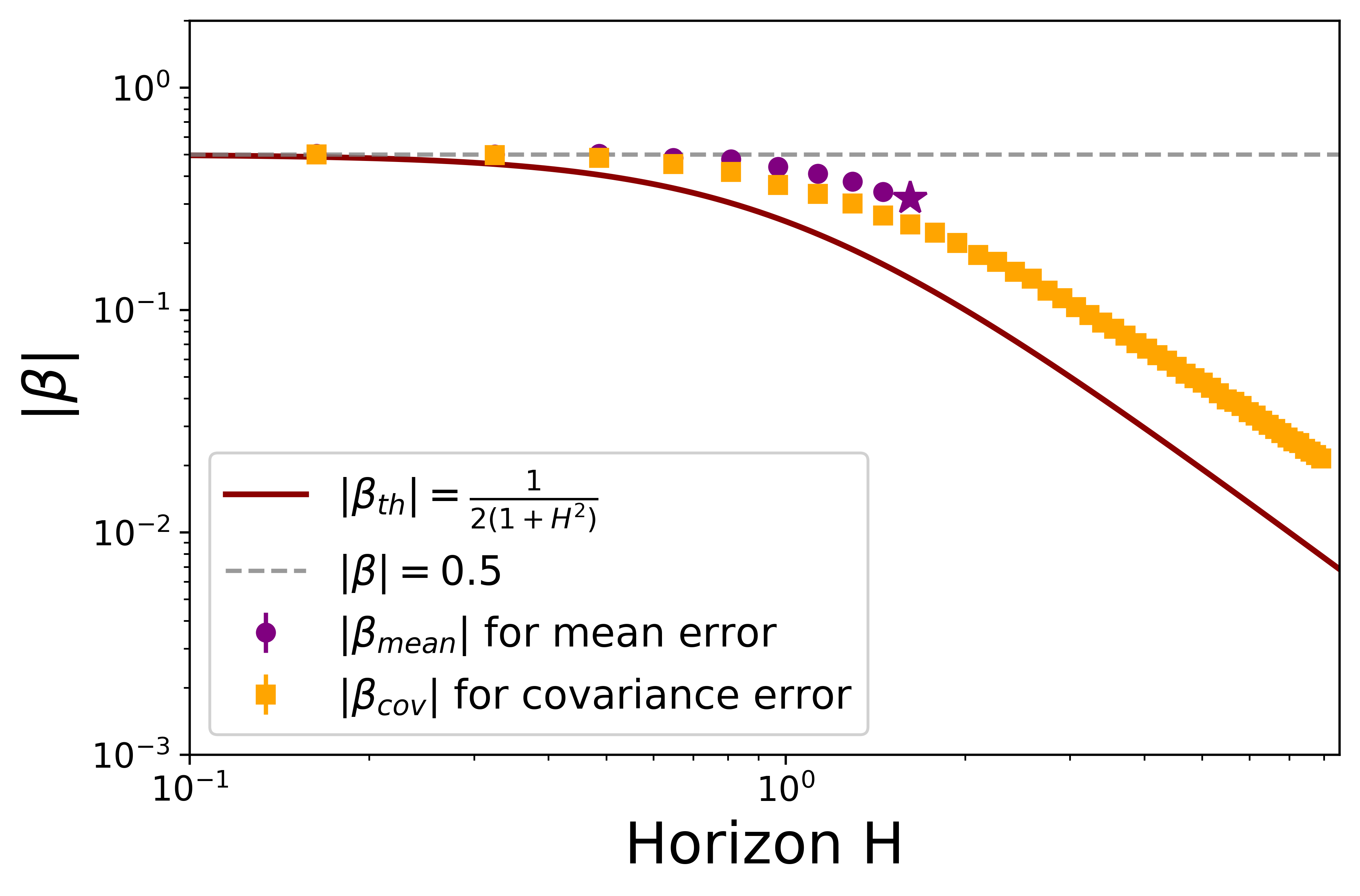}}
    \caption{
     \textbf{ Empirical and theoretical convergence rates $|\hat{\beta}|$ and $|\beta_{th}|$ vs horizon $H$ for Gaussian token} embeddings $\mathcal{N}(0, \Sigma)$ in dimension $4$ with $\Sigma = 0.1 \cdot I_4$. Gray dotted line indicates the classical rate exponent $|\beta_0|=0.5$ corresponding to the parametric rate in $1/\sqrt{n}$.
    }
    \label{gaussian_dim4}
  \end{center}
  \vskip -0.4in
\end{figure}

\cref{gaussian_dim4} shows strong agreement between the empirical rates and the prediction of Corollary \ref{coro:mean_convergence_rate} over a wide range of horizons, suggesting that the bound captures the correct rate in this Gaussian setting. For larger $H$, the mean error becomes too slow to be reliably fitted by a polynomial rate, marking the transition toward the large-horizon regime analyzed in \cref{sec:hardmax}. Additional high-dimensional experiments, for both Gaussian tokens and a non-Gaussian sub-Gaussian distribution, are reported in Appendix~\ref{app:add_results} (Figures~\ref{fig:Gaussian_plots} and~\ref{fig:uniform}).

\subsection{Slow rates on real data}
\paragraph{Model.}
% We conduct experiments using BigBird--RoBERTa--base \cite{Zaheer20}, a sparse-attention Transformer with $12$ layers, $768$ hidden dimensions, and $12$ attention heads. In its standard configuration, BigBird supports a maximum context length of $4096$ tokens. To operate in the large-context regime, we employ positional interpolation of the positional embeddings \cite{Chen23}, allowing us to process sequences of up to approximately $N_{\max} = 500{,}000$ tokens. This long sequence is used to construct the distribution $\nu$. To confirm that the slowdown of convergence we observe is not merely due to the sparsity of the BigBird model, we conduct the same experiment on BERT (see Appendix \ref{app:add_results}).
We use BigBird--RoBERTa--base \cite{Zaheer20}, a sparse-attention Transformer with $12$ layers, $768$ hidden dimensions, and $12$ heads. Positional interpolation \cite{Chen23} extends its $4096$-token context to sequences of up to $N_{\max}\approx 500{,}000$, from which we construct $\nu$. We repeat the experiment on BERT to verify that the slowdown is not specific to the sparsity of the BigBird model (Appendix~\ref{app:add_results}).

\paragraph{Text distributions.}
To study the impact of textual structure on attention convergence, we evaluate seven datasets drawn from Wikipedia English, German, and Chinese (\cite{Lhoest21}), as well as CC-News (\cite{Nagel16}). These datasets are chosen to probe the effects of domain shift, language, and syntactic structure on convergence behavior. 

\begin{figure}[ht]
  \vskip -0.08in
  \begin{center}
    \centerline{\includegraphics[width=\columnwidth]{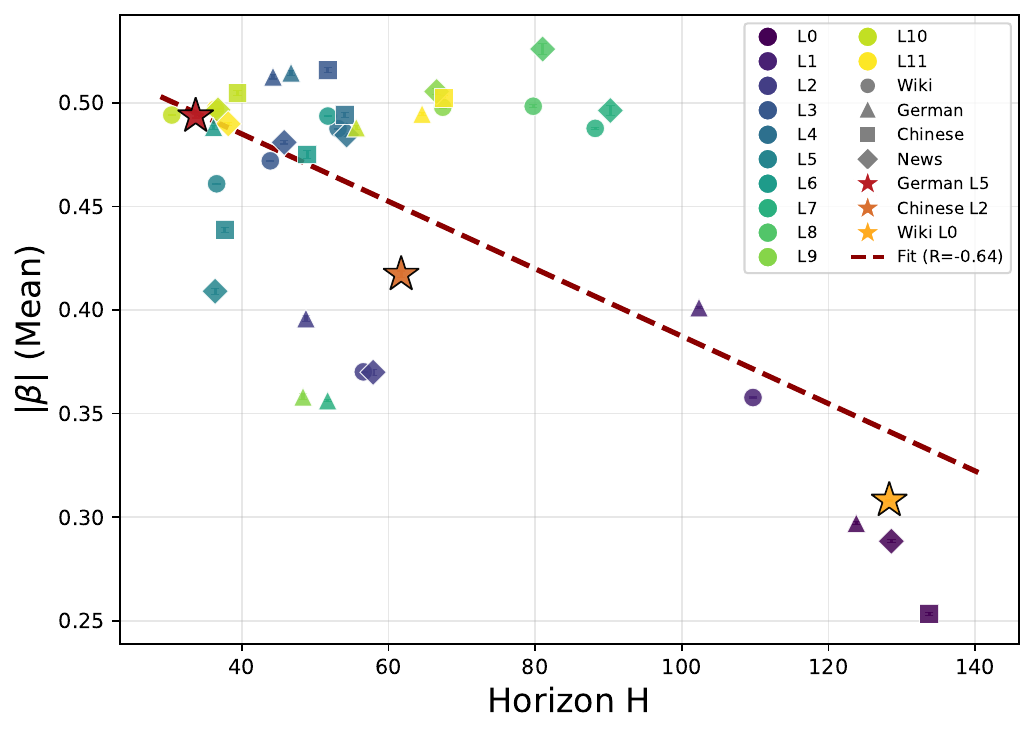}}
    \caption{\textbf{Mean convergence rate $|\beta|$ versus horizon $H$} across BigBird's layers $0$--$11$ and all text distributions. 
Markers represent $(\text{layer}, \text{text configuration})$ pairs; shapes denote text sources 
(Wikipedia EN/DE/ZH and CC-News) and colors indicate layers. 
Larger $|\beta|$ implies faster convergence. Star markers denote the configurations whose detailed 
convergence curves are displayed in \cref{fig:convergence_mean}. }
    \label{fig:slope_vs_H_stars}
  \end{center}
  \vskip -0.34in
\end{figure}
\begin{figure}[t!]
  % \vskip -0.08in
  \begin{center}
  \includegraphics[width=\columnwidth]{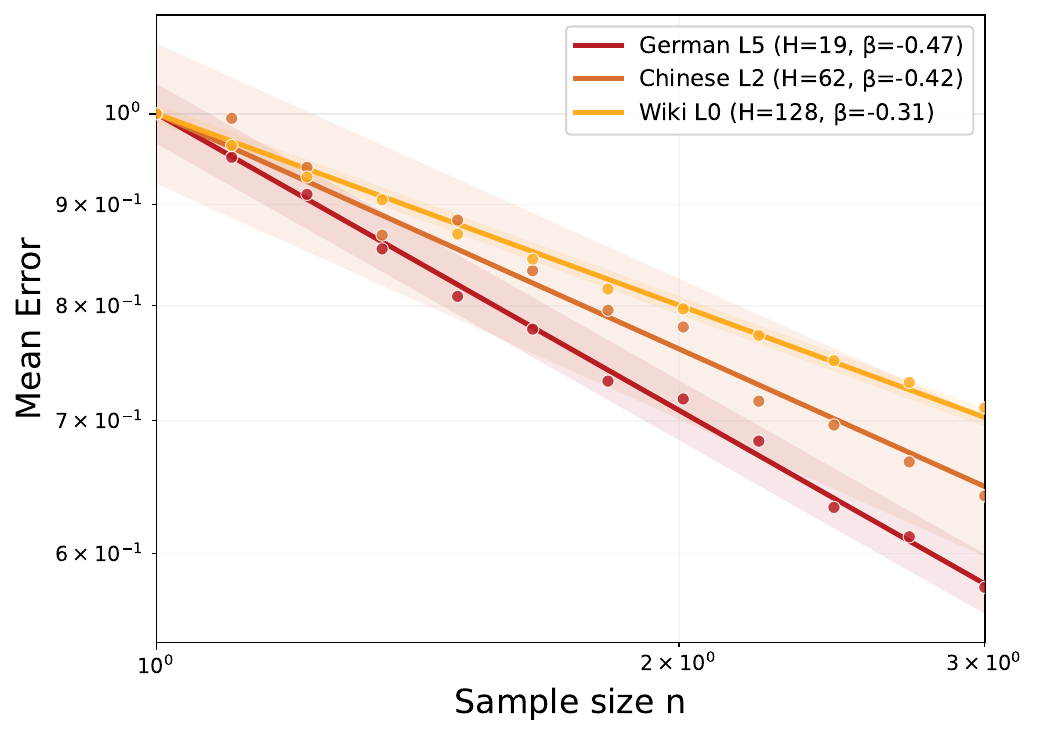}
  \vskip -0.1in
    \caption{\textbf{Mean error convergence versus subsample size $n$.}
For the starred BigBird-RoBERTa configurations in \cref{fig:slope_vs_H_stars},
we plot $\lVert\mathbb E_n[f_n(x)]-\mathbb E[f(x)]\rVert_2$ over $k=500$
subsamples on German Wikipedia. Log--log fits give exponents $|\beta|$, with
error scaling as $n^{-|\beta|}$; larger horizons yield slower convergence.}
    \label{fig:convergence_mean}
  \end{center}
  \vskip -0.28in
\end{figure}
\begin{figure}[ht]
\centering

\begin{subfigure}[t]{0.49\linewidth}
    \centering
    \includegraphics[width=\linewidth]{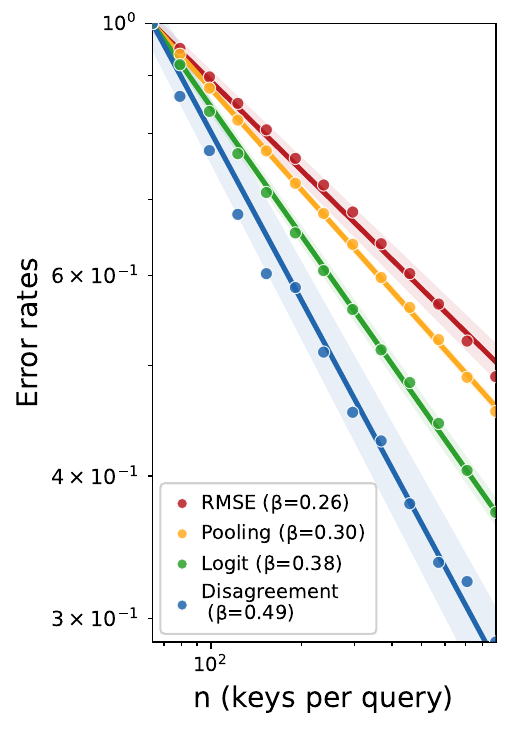}
    \caption{Downstream errors.}
    \label{fig:downstream_errors_panel}
\end{subfigure}
\hfill
\begin{subfigure}[t]{0.49\linewidth}
    \centering
    \includegraphics[width=\linewidth]{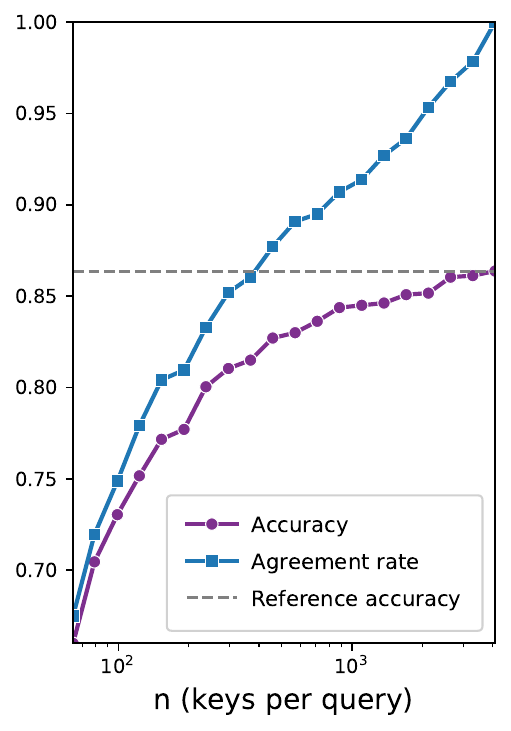}
    \caption{Accuracy/agreement.}
    \label{fig:downstream_accuracy_panel}
\end{subfigure}

\vspace{-1mm}
\caption{
\textbf{Downstream token sample complexity.}
BigBird--RoBERTa--base is evaluated on arxiv-classification by replacing dense
attention with attention over $n$ i.i.d.\ sampled keys per query.
Downstream errors decrease sub-parametrically 
% (RMSE ($\beta=0.27$), pooling error ($\beta=0.30$), logit error ($\beta=0.37$),
% and disagreement rate — fraction of examples where sparse and dense
% predictions disagree — ($\beta=0.46$)), 
(RMSE/pooling/logit/disagreement exponents:
$\beta=0.27,0.30,0.37,0.46$),
while agreement rate and accuracy
approach the dense model with diminishing returns.
}
\label{fig:downstream_classification}
\end{figure}
As shown in \cref{fig:slope_vs_H_stars}, convergence rates deteriorate systematically as the horizon $H$ increases. The rates also vary across text distributions at a fixed layer, reflecting the dependence of our bounds on the token geometry through $\Sigma$. In particular, $H=\|\Sigma^{1/2}A\Sigma^{1/2}\|_2$ does not measure anisotropy alone, but the interaction between the token distribution and the attention parameter $A$: convergence slows when the directions emphasized by attention align with high-variance directions of the token distribution. These results therefore reflect the combined influence of the model and the data. Analogous covariance results are reported in \cref{fig:bigbird_bert_additional} of Appendix~\ref{app:add_results}.

\cref{fig:convergence_mean} zooms in on the star-marked configurations. The linear behavior in log--log scale confirms a clear power-law convergence regime. 
Differences in slope across the three curves reveal the combined effect of layer depth and dataset anisotropy on convergence rates. 

\paragraph{Window-based sampling.}
We also examine whether the horizon-dependent slowdown persists under structured
key sampling. Keeping the same experimental protocol as above, we replace uniform
i.i.d.\ subsampling with two window-based rules: a local window around each query
supplemented with $r=50$ uniformly sampled out-of-window keys, and a hybrid rule
that splits the key budget evenly between local-window and out-of-window random
keys. These patterns are closer to practical sparse-attention mechanisms such as
BigBird. We again observe sub-parametric convergence (see \cref{fig:slope_vs_H_stars_window_cov} and \cref{fig:slope_vs_H_stars_window_rmse} in Appendix), with rates deteriorating as
the horizon increases, suggesting that the slowdown is not specific to i.i.d.\
subsampling but also appears under structured, window-based sparsification.

\paragraph{Downstream classification task} \label{sec:downstream_sec}
We finally test whether the finite-token convergence behavior observed at the attention-output level is reflected in end-to-end prediction quality. We fine-tuned BigBird--RoBERTa--base for $3$ epochs on the arxiv-classification task ($11$ categories, dense accuracy of $86.4\%$) using mean pooling, and evaluated inference with $n$ i.i.d.\ sampled keys per query, for $n$ ranging from $64$ to $4096$ ($K=5$ Monte Carlo repetitions, over $2400$ test documents). We compare sampled-key and dense inference through downstream errors, prediction agreement, defined as the fraction of examples where sparse and dense predictions agree, and classification accuracy. Figure~\ref{fig:downstream_classification} shows that the error metrics follow sub-parametric power laws $n^{-\beta}$ with $\beta<\tfrac12$, showing that the slowdown identified at the attention-output level also propagates through the full network to downstream prediction quality. When test examples are stratified by attention horizon, high-horizon groups converge more slowly, both in downstream errors and in the recovery of dense-model accuracy (see \cref{fig:downstream-practical}).
The sub-parametric convergence of the error curves can translate into
quantifiable insight into downstream accuracy: with only $25\%$ of keys
($n=1101$), the model already achieves $84.5\%$ accuracy, and doubling $n$
from $1101$ to $2124$ yields only $+0.7\%$ gain.

\section{Conclusion} 
In this paper, we introduced the notion of \emph{token sample complexity} to characterize the
convergence of the attention outputs as the sequence length $n$ increases. 
While parametric convergence rates can be derived for compactly supported token distributions,
these rates are asymptotic and fail to describe the finite-context regimes encountered
in practice. Encompassing compactly supported distributions, we extend our analysis to
sub-Gaussian tokens, allowing us to capture the combined effect of data anisotropy and attention
geometry on the convergence behavior of attention mechanisms. We derive slower finite-sample
convergence rates that match what is observed in both synthetic and real data. From an inference perspective, these rates quantify how quickly sampled-key
attention recovers dense-model behavior, with the attention horizon identifying
regimes where logits, predictions, and accuracy are slower to stabilize.
Our upper bound applies to a large class of distributions and describes the observed slow
convergence rate on real data. However, showing the tightness of our upper bound in the Gaussian
setting, as suggested by our numerical experiments, remains widely open.
More generally, the precise rate of convergence of attention crucially depends on the token distribution, and identifying classes of distributions and their corresponding tight rates is left for future work.

\section*{Acknowledgements}
This work was granted access to the HPC resources of IDRIS under the allocation 2025-[A0181016159] made by GENCI.
The work of Gabriel Peyr\'e was supported by the European Research Council (ERC project WOLF) and the French government under the management of Agence Nationale de la Recherche as part of the ``France 2030'' program, reference ANR-23-IACL-0008 (PRAIRIE-PSAI).
The work of Léa Bohbot was supported by the Fondation CFM, through the Jean-Pierre Aguilar fellowship.

\section*{Impact Statement}

% Authors are \textbf{required} to include a statement of the potential broader
% impact of their work, including its ethical aspects and future societal
% consequences. This statement should be in an unnumbered section at the end of
% the paper (co-located with Acknowledgements -- the two may appear in either
% order, but both must be before References), and does not count toward the paper
% page limit. In many cases, where the ethical impacts and expected societal
% implications are those that are well established when advancing the field of
% Machine Learning, substantial discussion is not required, and a simple
% statement such as the following will suffice:

This paper presents work whose goal is to advance the field of Machine
Learning. There are many potential societal consequences of our work, none
which we feel must be specifically highlighted here.

% The above statement can be used verbatim in such cases, but we encourage
% authors to think about whether there is content which does warrant further
% discussion, as this statement will be apparent if the paper is later flagged
% for ethics review.


\bibliography{main}
\bibliographystyle{icml2026}

%%%%%%%%%%%%%%%%%%%%%%%%%%%%%%%%%%%%%%%%%%%%%%%%%%%%%%%%%%%%%%%%%%%%%%%%%%%%%%%
%%%%%%%%%%%%%%%%%%%%%%%%%%%%%%%%%%%%%%%%%%%%%%%%%%%%%%%%%%%%%%%%%%%%%%%%%%%%%%%
% APPENDIX
%%%%%%%%%%%%%%%%%%%%%%%%%%%%%%%%%%%%%%%%%%%%%%%%%%%%%%%%%%%%%%%%%%%%%%%%%%%%%%%
%%%%%%%%%%%%%%%%%%%%%%%%%%%%%%%%%%%%%%%%%%%%%%%%%%%%%%%%%%%%%%%%%%%%%%%%%%%%%%%
\newpage
\appendix
\onecolumn

\section{Additional figures}\label{sec:add_figures}
\captionsetup[figure]{aboveskip=4pt,belowskip=0pt}
This section presents additional experimental results on both synthetic and real data in high dimension. All synthetic experiments were conducted using multi-head attention following Definition \ref{def:mh_attn}.

\begin{figure}[H]
  \centering
  \begin{subfigure}{0.485\linewidth}
    \centering
    \includegraphics[width=\linewidth]{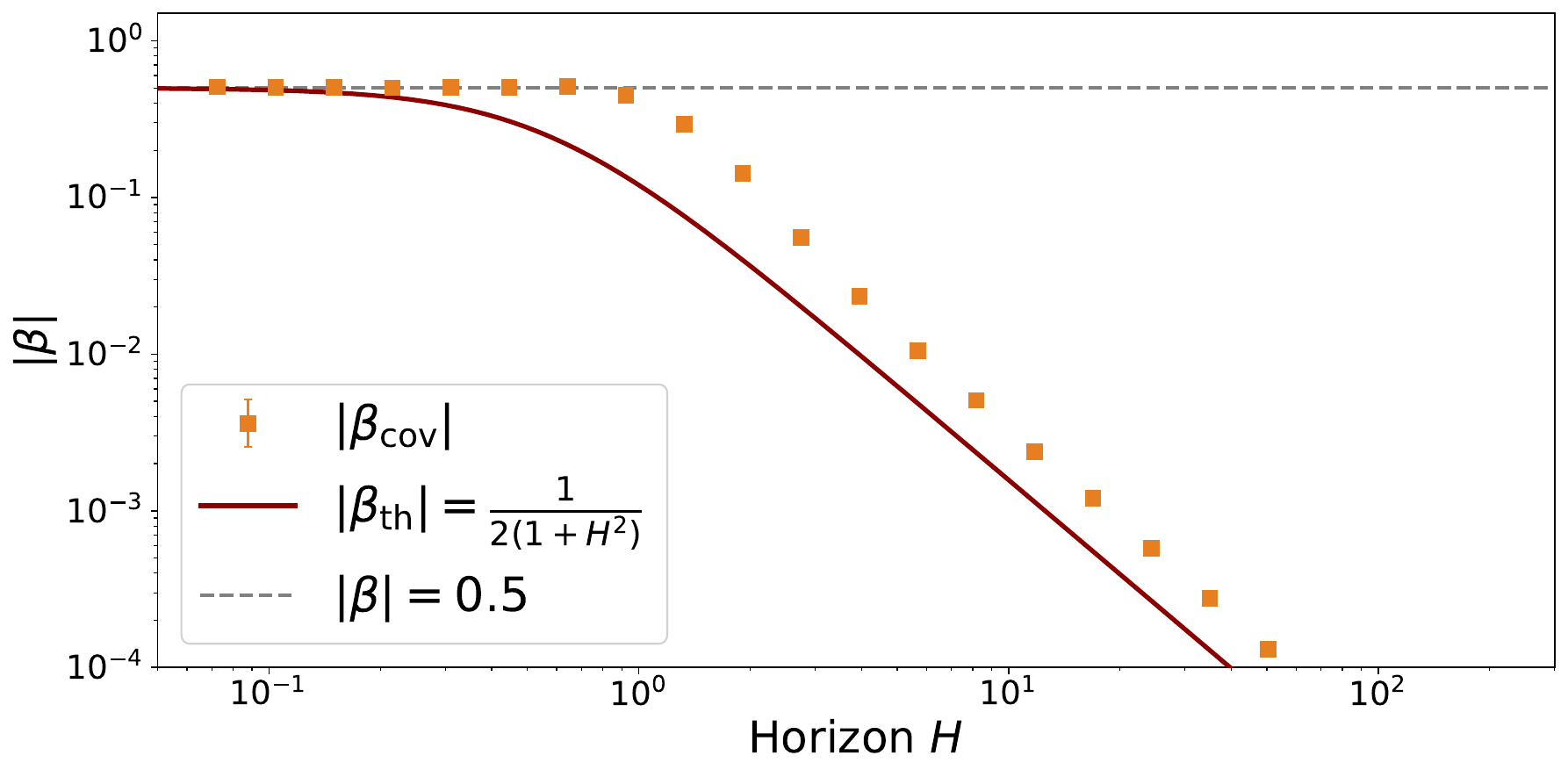}
    \caption{$d = 30$}
  \end{subfigure}\hfill
  \begin{subfigure}{0.485\linewidth}
    \centering
    \includegraphics[width=\linewidth]{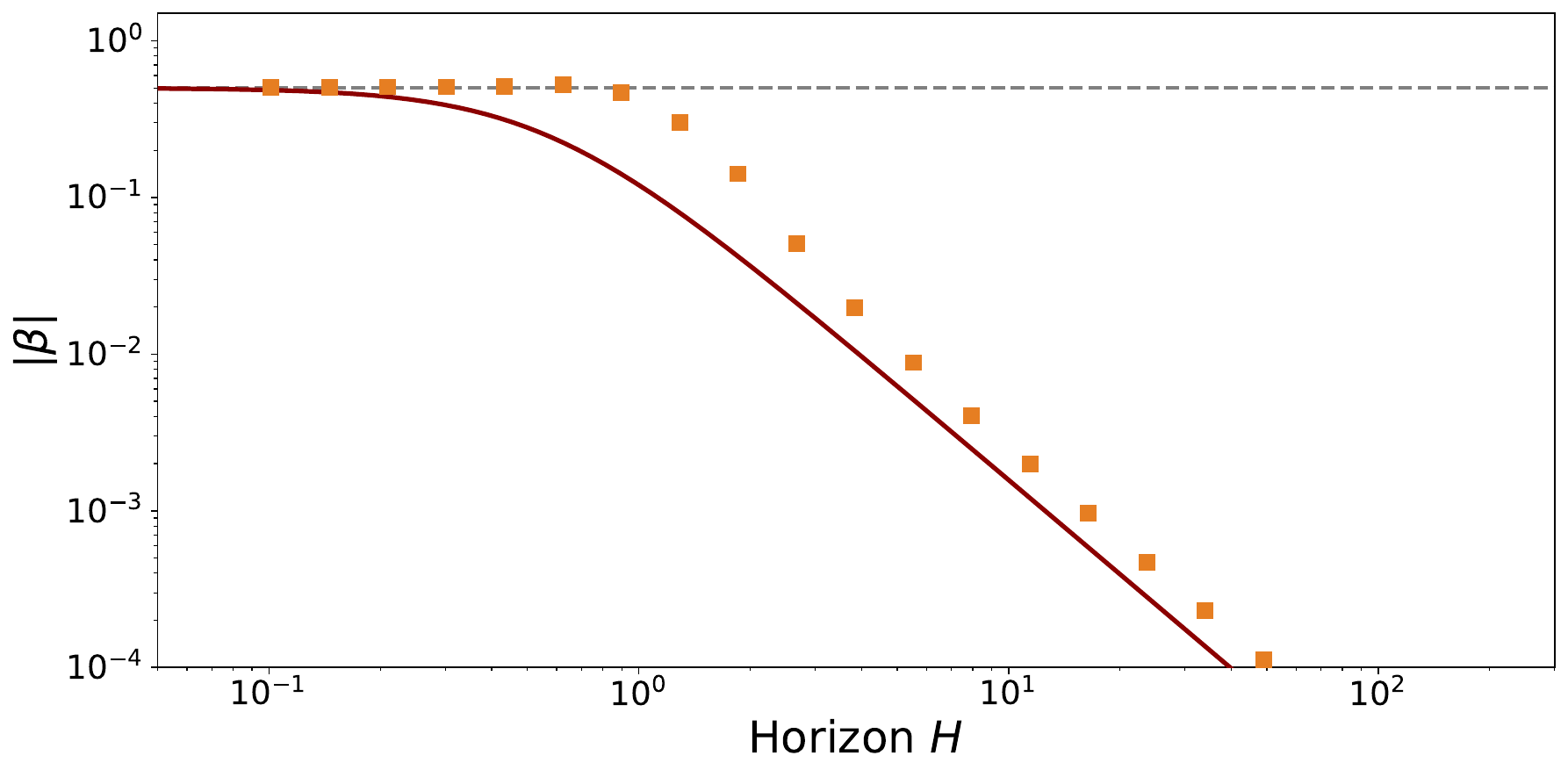}
    \caption{$d = 40$}
  \end{subfigure}
  \vspace{0.25em}

  \begin{subfigure}{0.485\linewidth}
    \centering
    \includegraphics[width=\linewidth]{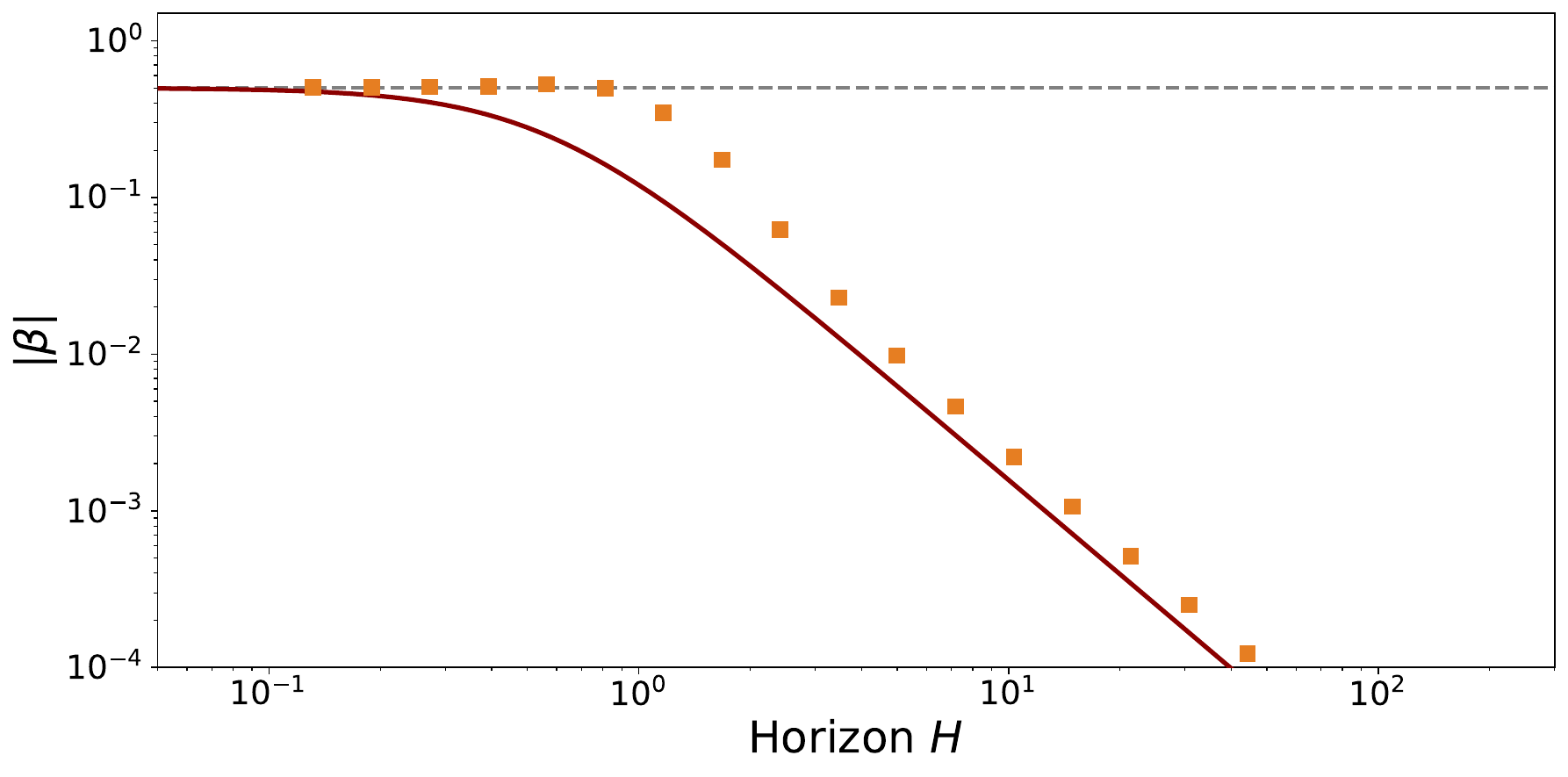}
    \caption{$d = 50$}
  \end{subfigure}\hfill
  \begin{subfigure}{0.485\linewidth}
    \centering
    \includegraphics[width=\linewidth]{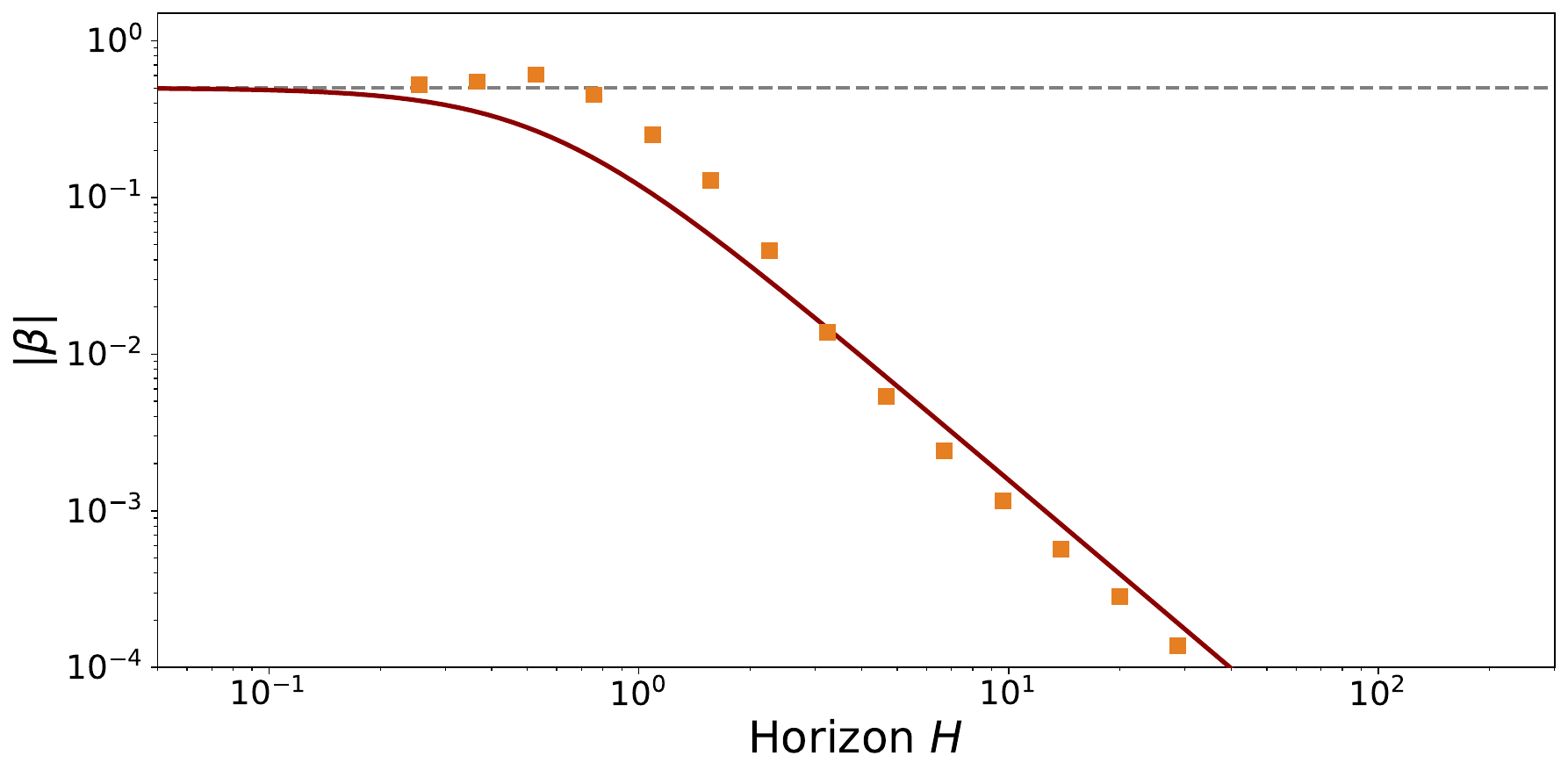}
    \caption{$d = 100$}
  \end{subfigure}
  \caption{\textbf{Covariance convergence rate $|\beta|$ vs horizon $H$ for Gaussian token} embeddings $\mathcal{N}(0, \Sigma)$, with $\Sigma =  \lambda_{max}\mathbf{I}$, across dimensions $d \in \{20, 30, 40, 50, 100\}$. We vary the horizon by rescaling a fixed matrix $A$, to capture both low and high-horizon regimes. Diagonal covariance matrices have identical spectral properties of $\Sigma$ across dimensions. The results demonstrate strong agreement with theoretical rate of Proposition \ref{prop:cov_emp_vs_pop}.}
  \label{fig:Gaussian_plots}
  \vskip -0.1in
\end{figure}

\begin{figure}[H]
  \vspace{-0.8em}
  \centering
  \begin{minipage}[c]{0.50\linewidth}
    \centering
    \includegraphics[width=\linewidth]{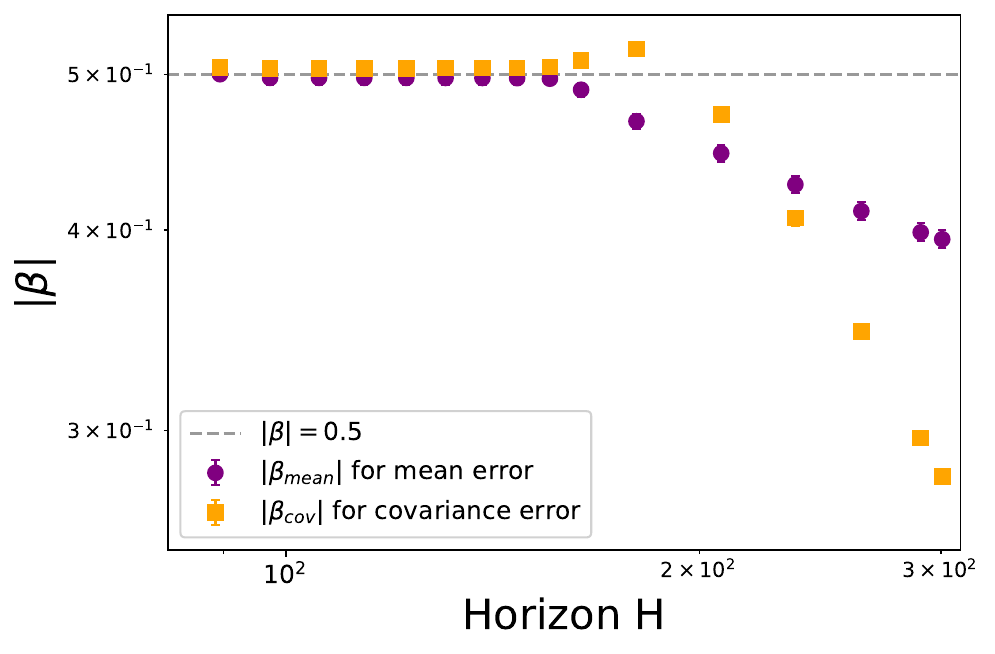}
  \end{minipage}\hfill
  \begin{minipage}[c]{0.44\linewidth}
    \caption{
      \textbf{Empirical mean and covariance convergence rate exponents $|\beta_{mean}|$ and  $|\beta_{cov}|$} as a function of attention horizon $H$ for uniform token embeddings on a unit sphere in dimension $50$.  Gray dotted line indicates the classical rate exponent $|\beta|=0.5$ corresponding to the parametric rate in $1/\sqrt{n}$.}
    \label{fig:uniform}
  \end{minipage}
  \vspace{-0.8em}
\end{figure}

\begin{definition}[Multi-head self-attention]\label{def:mh_attn}
Let $H_{\text{heads}}$ divide $d$ and set $k = d/H_{\text{heads}}$. For each head $h \in \{1,\ldots,H_{\text{heads}}\}$, let $Q^{(h)},K^{(h)},V^{(h)} \in \mathbb{R}^{k\times d}$ and an output projection $W^{(h)} \in \mathbb{R}^{d\times k}$. Denoting by $f_n^{(h)}$ the single-head map associated with $\big(Q^{(h)},K^{(h)},V^{(h)}\big)$, the multi-head operator is
\[
f_n^{\mathrm{MH}}(X) \;=\; \sum_{h=1}^{H_{\text{heads}}} W^{(h)} \, f_n^{(h)}(X) \;\in\; (\mathbb{R}^d)^n .
\]
\end{definition}
\begin{figure}[!t]
\centering

% ---------- Row 1: BigBird covariance ----------
\begin{subfigure}[t]{0.48\textwidth}
    \centering
    \includegraphics[width=\linewidth]{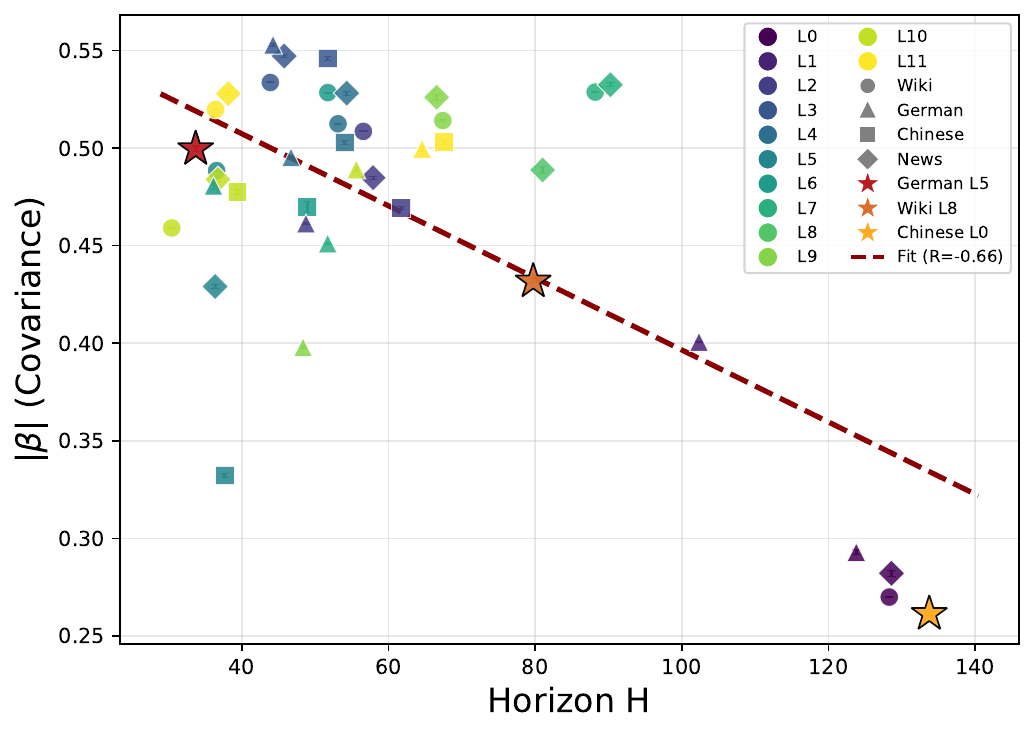}
    \caption{BigBird: rate vs horizon}
    \label{fig:bigbird_cov_rate}
\end{subfigure}
\hfill
\begin{subfigure}[t]{0.48\textwidth}
    \centering
    \includegraphics[width=\linewidth]{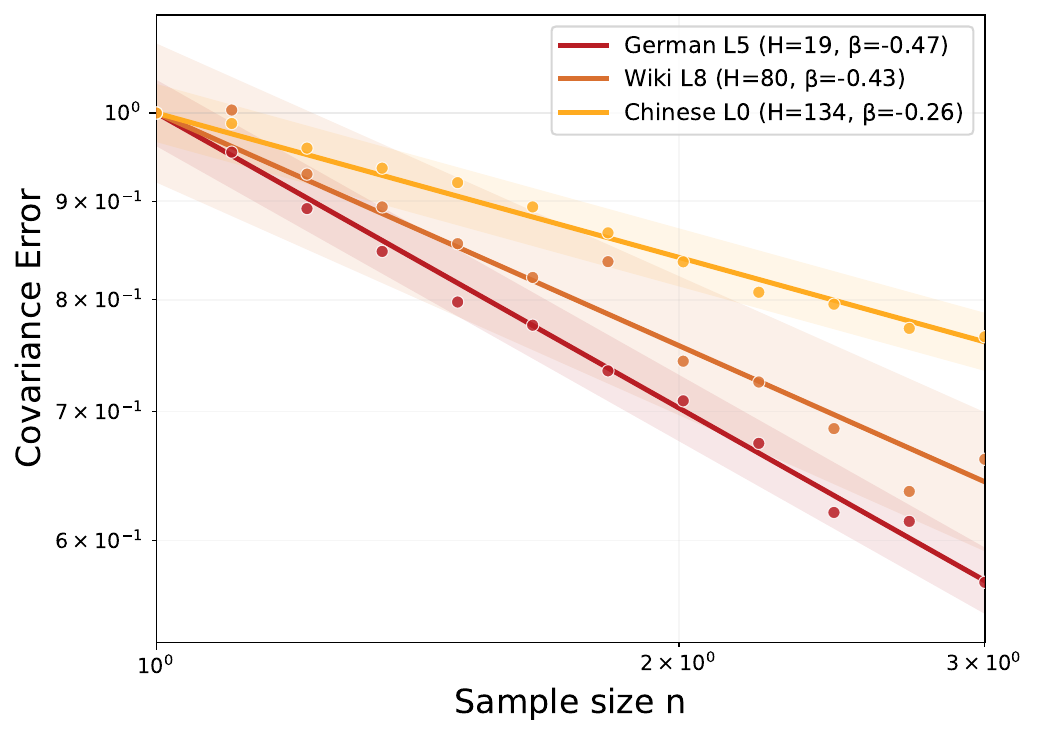}
    \caption{BigBird: convergence curves}
    \label{fig:bigbird_cov_curves}
\end{subfigure}

\vspace{1.5mm}

% ---------- Row 2: BERT mean/covariance ----------
\begin{subfigure}[t]{0.48\textwidth}
    \centering
    \includegraphics[width=\linewidth]{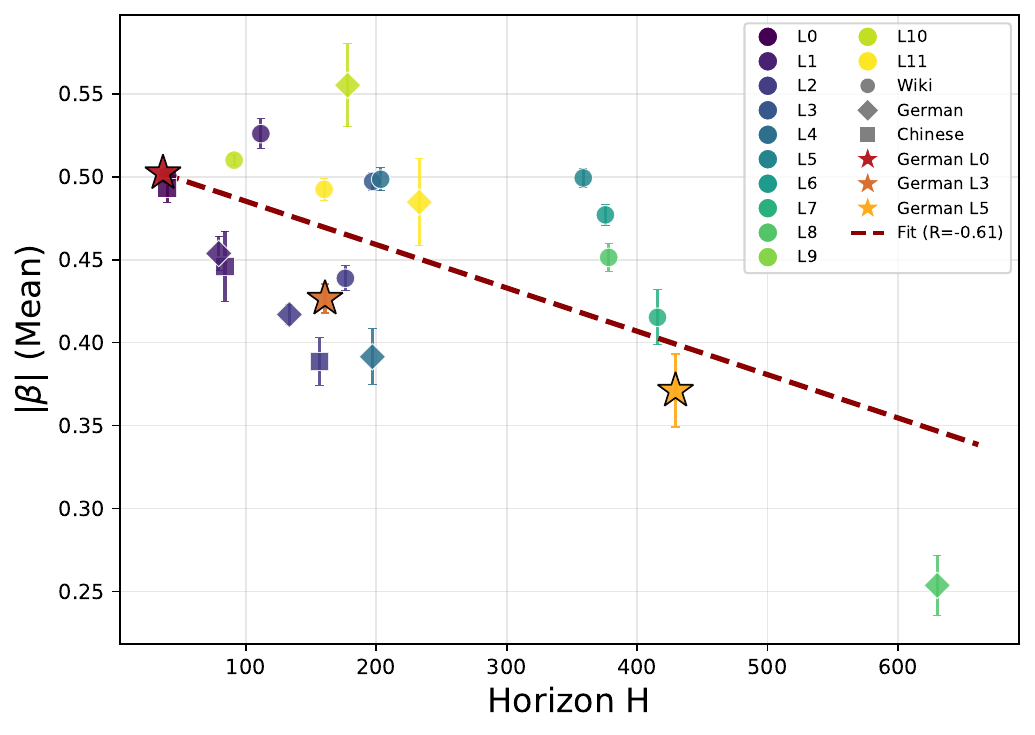}
    \caption{BERT: mean error}
    \label{fig:bert_mean_rate}
\end{subfigure}
\hfill
\begin{subfigure}[t]{0.48\textwidth}
    \centering
    \includegraphics[width=\linewidth]{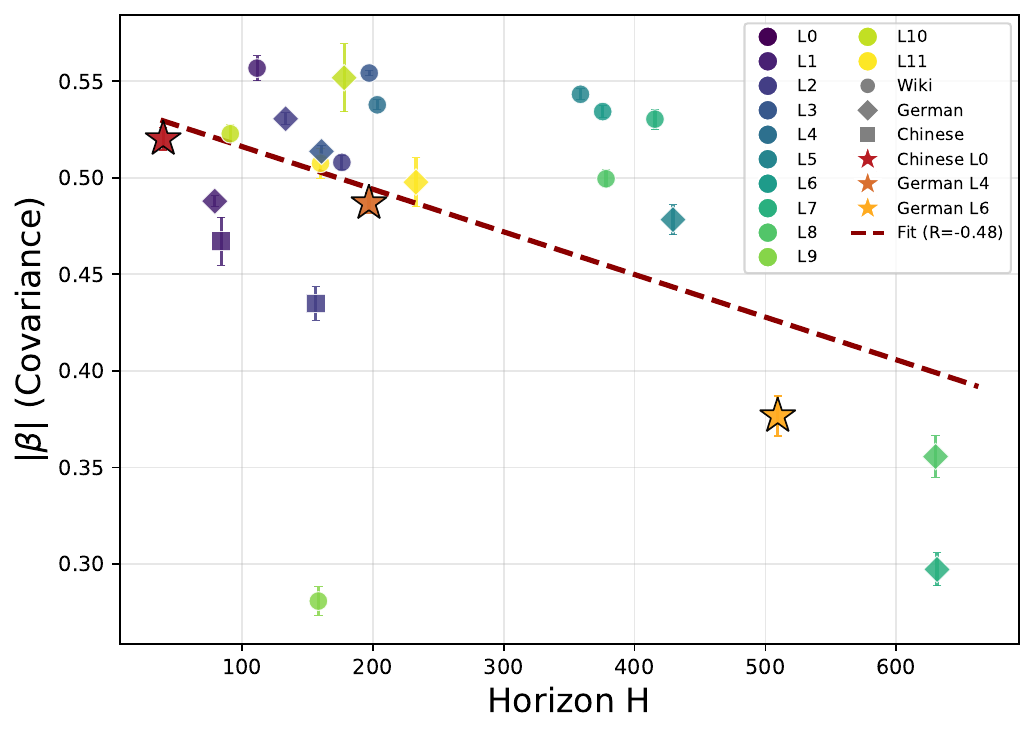}
    \caption{BERT: covariance error}
    \label{fig:bert_cov_rate}
\end{subfigure}

\vspace{-1mm}

\caption{
\textbf{Additional convergence-rate experiments on BigBird and BERT.}
Top row: covariance convergence for BigBird-RoBERTa across layers and text distributions.
The left panel reports the fitted convergence rate $|\beta|$ versus horizon $H$; the right panel shows the detailed covariance-error curves for the starred configurations.
Bottom row: convergence rate $|\beta|$ versus horizon $H$ for BERT-base-uncased, for mean error and covariance error across Wikipedia text distributions.
Markers indicate $(\text{layer},\text{text configuration})$ pairs, colors indicate layers, and larger $|\beta|$ means faster convergence.
}
\label{fig:bigbird_bert_additional}
\end{figure}

\begin{figure}[H]
\centering
\begin{subfigure}[b]{0.48\textwidth}
    \includegraphics[width=\textwidth]{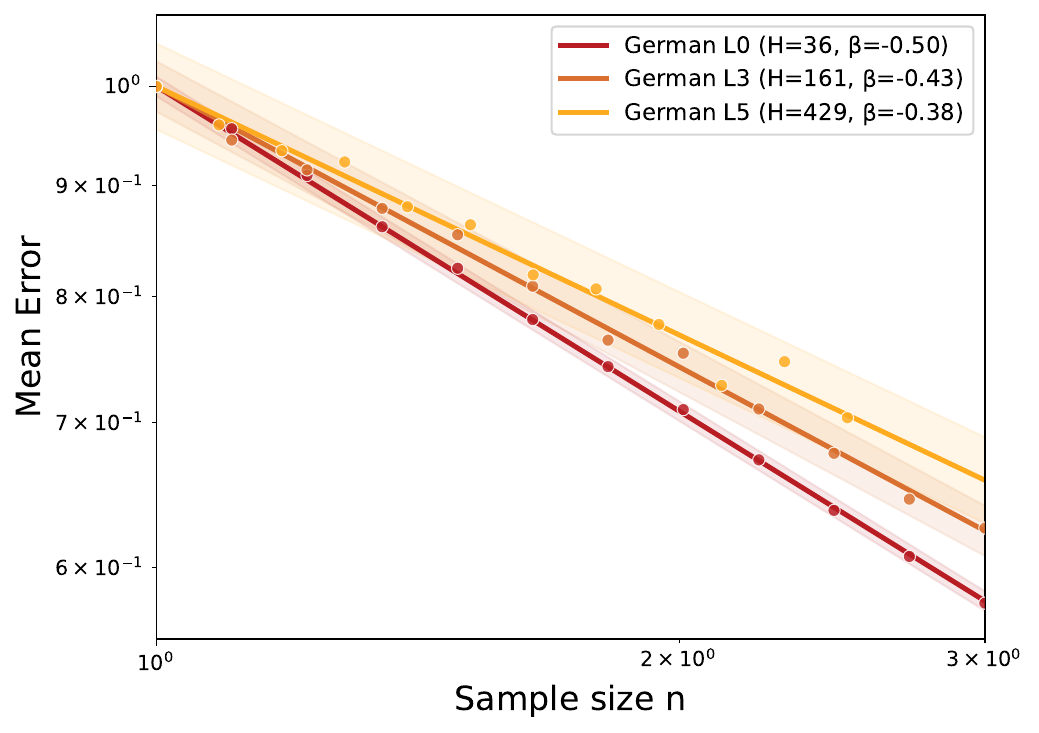}
    \caption{Mean error}
\end{subfigure}
\hfill
\begin{subfigure}[b]{0.48\textwidth}
    \includegraphics[width=\textwidth]{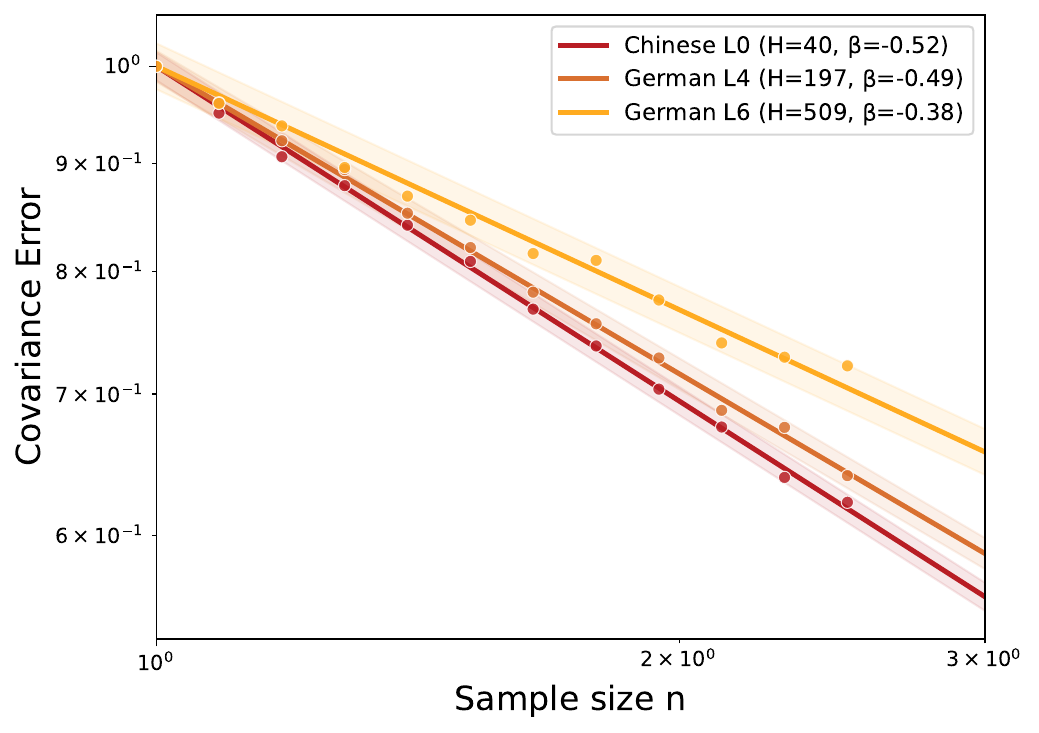}
    \caption{Covariance error}
\end{subfigure}
\caption{\textbf{Convergence curves for starred layers.}
  Mean error (left) and covariance error (right) versus subsample size $n$ for
  the three starred configurations from Figure~\ref{fig:bigbird_bert_additional}, using
  BERT-base-uncased (dense attention).
  Markers show the averaged $\ell_2$ errors over $k=500$--$1000$ subsamples,
  with $\pm 3$ standard-error bands.
  Log--log linear fits yield power-law exponents $|\beta|$ (legend), with error
  scaling as $n^{-|\beta|}$.
  From dark red to light orange as the horizon $H$ increases, confirming the
  slowdown predicted by Theorem~5.3.}
\label{fig:bert_convergence}
\end{figure}
\begin{figure}[H]
\centering
\begin{subfigure}[t]{0.49\linewidth}
    \centering
    \includegraphics[width=\linewidth]{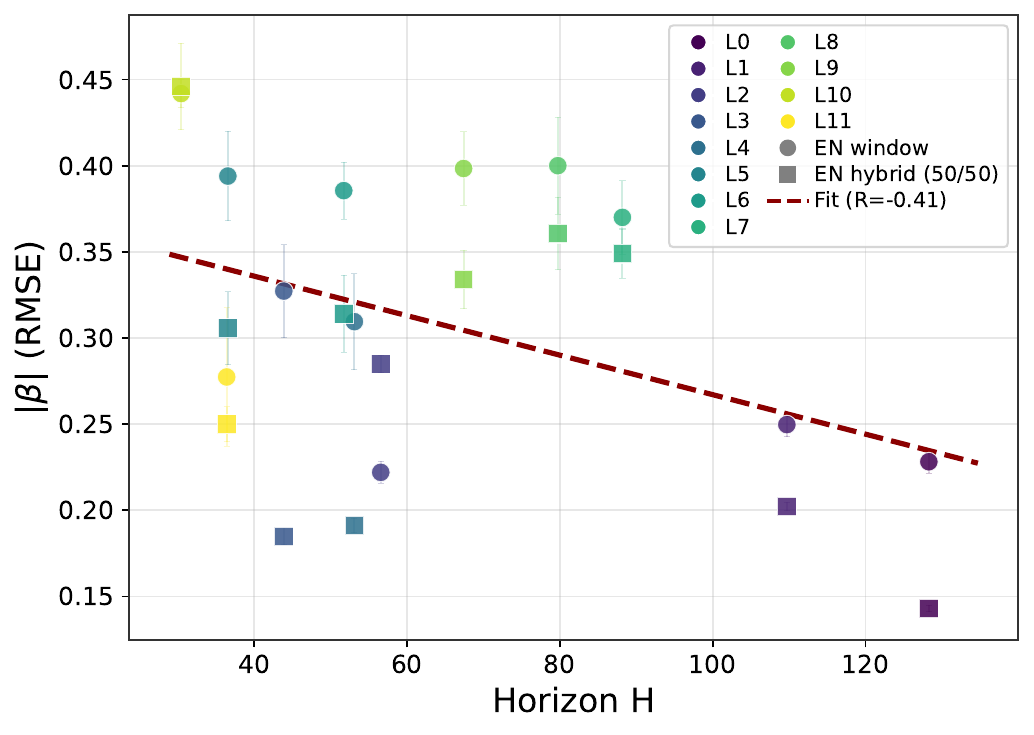}
    \caption{Mean/RMSE error}
    \label{fig:slope_vs_H_stars_window_rmse}
\end{subfigure}
\hfill
\begin{subfigure}[t]{0.49\linewidth}
    \centering
    \includegraphics[width=\linewidth]{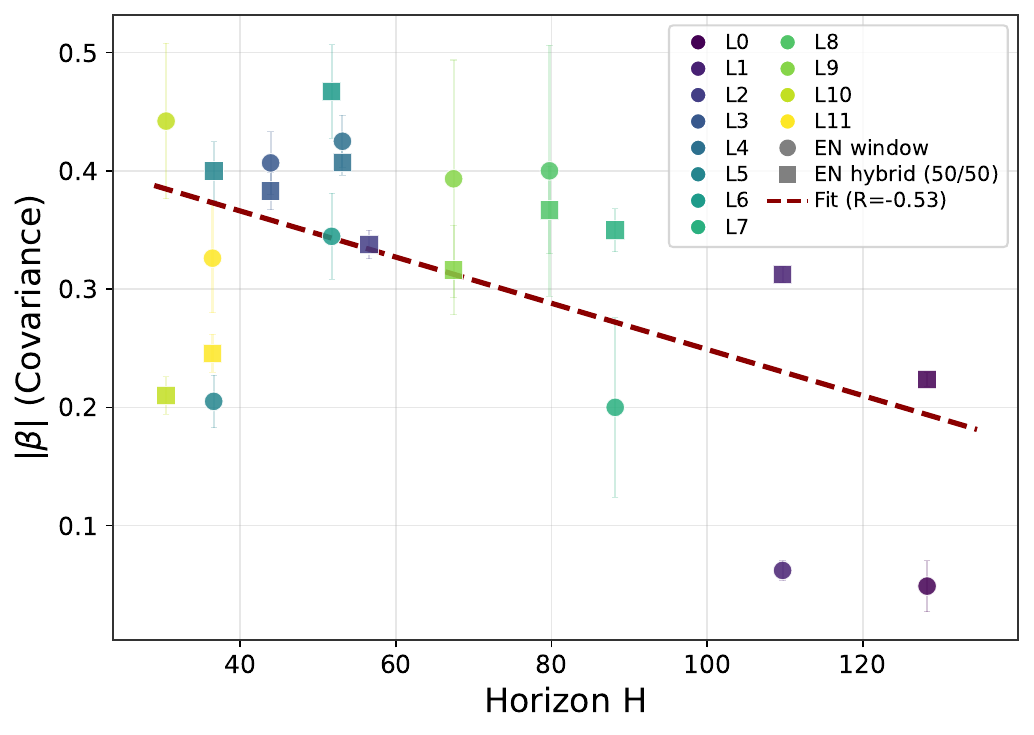}
    \caption{Covariance error}
    \label{fig:slope_vs_H_stars_window_cov}
\end{subfigure}

\vspace{-1mm}
\caption{\textbf{Convergence rate $|\beta|$ versus horizon $H$ under non-i.i.d. windowed sampling.}
BigBird--RoBERTa layers $0$--$11$ are evaluated on English Wikipedia with two structured schemes:
\emph{window} ($n-r-1$ local keys plus $r=50$ random keys) and
\emph{hybrid} (half local, half random). In both mean/RMSE and covariance errors, larger horizons still correspond to slower convergence, confirming that the slowdown persists beyond i.i.d.\ sampling.}
\label{fig:slope_vs_H_stars_window}
\end{figure}

\paragraph{Implementation details.}\label{app:xp_details}
All experiments use $N_{\max} \approx 5\times 10^5$ reference tokens of each dataset as the reference limit distribution, and pass them through BigBird--RoBERTa--base to compute the empirical mean and covariance of the attention outputs at each layer, which serve as approximations of the infinite-context limits. For subsample sizes $n \in [1000,3000]$, we randomly sample $n$ tokens from the full sequence and estimate the corresponding mean and covariance errors $\bigl\|\mathbb{E}_n[f_n(\hat X)] - \mathbb{E} [f(X)]\bigr\|_2$ and $\big\|\operatorname{Cov}_n(f_n(\hat X))-\operatorname{Cov}(f(X))\big\|_2$ of Corollaries  \ref{prop:cov_emp_vs_pop} and \ref{coro:mean_convergence_rate}
. Each Monte Carlo estimate is repeated $500$ times to reduce variance. 

\paragraph{Discussion on the i.i.d. assumption}\label{app:iid_assump}
Having tokens drawn from a distribution with Markovian dependencies does not contradict our theory: as long as tokens are sampled i.i.d. from that distribution — which is the case in our real-data experiments — the theorem applies directly. Nonetheless, in practice, subsampling is often structured, e.g. via windowing, and in that case the i.i.d. assumption no longer holds. Obtaining theoretical guarantees in this setting is significantly harder, which motivates our simplifying assumption. We addressed this concern empirically by conducting an additional experiment using windowing-based subsampling on BigBird (see \cref{sec:experiments} and \cref{fig:slope_vs_H_stars_window}), and observe the same slowdown in convergence, suggesting that our theoretical predictions remain informative beyond the i.i.d. setting. 

% \begin{figure}[H]
%   \centering
%   \begin{minipage}[c]{0.50\linewidth}
%     \centering
%     \includegraphics[width=\linewidth]{icml_figures/Window_RMSE_slope_vs_H.pdf}
%   \end{minipage}\hfill
%   \begin{minipage}[c]{0.44\linewidth}
%     \caption{\textbf{Convergence rate $|\beta|$ versus horizon $H$ under non-i.i.d. windowed sampling for BigBird-RoBERTa's layers}
%       $0$--$11$ on English Wikipedia, under two sampling
%     strategies: fixed-random (EN window, $r=50$ random tokens outside a sliding
%     window of size $n-r-1$) and hybrid (EN hybrid 50/50, half window and half
%     random tokens). Colors indicate layers. The negative trend (fit slope
%     $R=-0.65$) confirms that the slowdown driven by $H$ persists under structured,
%     non-i.i.d.\ sampling.}
%     \label{fig:slope_vs_H_stars_window_rmse}
%   \end{minipage}
% \end{figure}
% \begin{figure}[ht]
%   \vskip -0.08in
%   \begin{center}
%     \centerline{\includegraphics[width=\columnwidth]{icml_figures/Window_COV_slope_vs_H.pdf}}
%     \caption{
% \textbf{Covariance convergence rate $|\beta|$ versus horizon $H$ under structured sampling.}
% BigBird-RoBERTa layers $0$--$11$ are evaluated on English Wikipedia with two non-i.i.d. schemes:
% \emph{window} ($n-r-1$ local keys plus $r=50$ random keys) and
% \emph{hybrid} (half local, half random).
% Larger horizons still correspond to slower convergence.
% }
%     \label{fig:slope_vs_H_stars_window_cov}
%   \end{center}
%   \vskip -0.34in
% \end{figure}
\section{Tools from empirical process theory}\label{sec:empirical_proc_tools}

Before diving into the proof of Theorem \ref{thm:uniform_convergence}, we recall a few results and definitions from empirical process theory that will be useful to the proof.

First, recall the definition of the Rademacher complexity. 
\begin{definition}[Rademacher complexity in $\mathbb R^d$]
Let $\mathcal{F}\subset\{f:\mathcal{Y}\to\mathbb{R}^d\}$ be a class of (measurable) functions and 
$S=(Y_1,\dots,Y_n)\sim\nu^n$ an i.i.d.\ sample. 
Let $\varepsilon_1,\dots,\varepsilon_n$ be i.i.d.\ Rademacher variables, independent of $S$, with 
$\mathbb{P}(\varepsilon_i=1)=\mathbb{P}(\varepsilon_i=-1)=\tfrac12$. The \emph{empirical} Rademacher complexity of $\mathcal{F}$ with respect to\ $S$ is:
\[
\widehat{\mathfrak{R}}_{S}(\mathcal{F})
\;:=\;
\mathbb{E}_{\varepsilon}\!\left[
\sup_{f\in\mathcal{F}}
\left\|
\frac{1}{n}\sum_{i=1}^n \varepsilon_i\, f(Y_i)
\right\|_2
\right].
\]
The (expected) Rademacher complexity at sample size $n$ is
\[
\mathfrak{R}_n(\mathcal{F})
\;:=\;
\mathbb{E}_{S\sim\nu^n}\big[\widehat{\mathfrak{R}}_{S}(\mathcal{F})\big].
\]
If $\mathcal{F}=\{f_\theta:\theta\in\Theta\}$ is parameterized by $\Theta$, replace 
$\sup_{f\in\mathcal{F}}$ with $\sup_{\theta\in\Theta}$. 
For $d=1$, this reduces to the usual scalar definition.
\end{definition}

Intuitively, the Rademacher complexity quantifies the ability of the function class $\mathcal F$ to fit random noise on the specific sample $S \sim \nu$.
Notice that the expression resembles the empirical expectation we aim to bound. The symmetrization step consists in artificially introducing Rademacher samples $\varepsilon_i$ to relate our supremum over $\mathcal F$ to a measure of the richness of the class, that relies on measuring the alignment between function outputs $f(Y_i)$ and random noise. 

Next, we recall some useful definitions about sub-Gaussian processes and the Dudley's entropy integral, central in the proof Theorem \ref{thm:uniform_convergence}. 
\begin{definition}[Sub-Gaussian $\psi_2$ norm]\label{def:psi2-norm}
For a real-valued random variable $Z$, its $\psi_2$ (sub-Gaussian) norm is
\[
\|Z\|_{\psi_2}
:= \inf\Bigl\{\sigma>0:\; \mathbb{E}\!\left[\exp\!\left(\frac{Z^{2}}{\sigma^{2}}\right)\right]\le 2\Bigr\}.
\]
For a random vector $X\in\mathbb{R}^d$,
\[
\|X\|_{\psi_2}
:= \sup_{v\in \mathbb{S}^{d-1}} \|\langle X,v\rangle\|_{\psi_2},
\]
where $\mathbb{S}^{d-1}=\{v\in\mathbb{R}^d:\|v\|_2=1\}$. We say $X$ is sub-Gaussian if $\|X\|_{\psi_2}<\infty$.
\end{definition}

\begin{definition}[Sub-gaussian increments]\label{def:subgaussian}
Consider a random process $(X_t)_{t \in T}$ on a metric space $(T, d)$. We say that the process has \emph{sub-Gaussian increments} if there exists $K \geq 0$ such that
\begin{equation}
\|X_t - X_s\|_{\psi_2} \leq K d(t, s) \quad \text{for all } t, s \in T, \label{eq:subgaussian}
\end{equation}
where $\|\cdot\|_{\psi_2}$ denotes the sub-Gaussian norm. \\

\end{definition}

\begin{definition}[$\varepsilon$-net and covering numbers]\label{def:cov_nb}
Let $(T, d)$ be a metric space. Consider a set $T' \subset T$ and a number $\varepsilon > 0$. A subset $\mathcal N \subset T'$ is an $\varepsilon$-net of $T$ if balls of radius $\varepsilon$ centered at points in $\mathcal N$ cover $T'$.\\
The smallest cardinality of an $\varepsilon$-net of $T'$ is called the covering number of $K$ and is denoted $\mathcal N(\varepsilon, T', d)$. Equivalently, $\mathcal N(\varepsilon, T', d)$ is the smallest number of closed balls with centers in $T'$ and radius $\varepsilon$ whose union covers $T'$.
\end{definition}

\begin{theorem}[Dudley's entropy integral](\cite{Vershynin18}, see Theorem 8.1.3, p. 221, 2nd ed.) \label{thm:dudley}
Let $(X_t)_{t \in T}$ be a centered random process on a metric space $(T, d)$ with sub-Gaussian increments as in \eqref{eq:subgaussian}, with $K \ge 0$ its sub-Gaussian parameter. Then, there exists a constant $C \ge 0$ such that: 

\begin{equation}
\mathbb{E} [\sup_{t \in T} X_t] \leq CK \int_0^{\text{diam}(T)} \sqrt{\ln \mathcal N(\varepsilon, T, d)} \, d\varepsilon,
\end{equation}

where $\mathcal N(\varepsilon, T, d)$ denotes the covering number of  $T$ at scale $\varepsilon$ introduced in Definition \ref{def:cov_nb}.
\end{theorem}

\begin{definition}[Data-dependent distance]\label{def:data_distance}
Let $Y_1,\dots,Y_n$ be a sample of random vectors in $\mathbb R^d$ and let $\mathbb P_n := \frac1n\sum_{i=1}^n \delta_{Y_i}$ denote the empirical measure.
For any measurable $g$, define the empirical $L^2(\mathbb P_n)$ norm under the samples  $Y_1,\dots,Y_n$ by
\[
\|g\|_{L^2(\mathbb P_n)} \;:=\; \Big(\tfrac1n\sum_{i=1}^n \|g(Y_i)\|_2^2\Big)^{1/2}.
\]
For the indexed class $\mathcal F=\{f_x:\, x\in B_R\}$, the data-dependent distance between $f_x$ and $f_{x'}$ is
\[
\|f_x-f_{x'}\|_{L^2(\mathbb P_n)} \;=\; \Big(\tfrac1n\sum_{i=1}^n \|f_x(Y_i)-f_{x'}(Y_i)\|_2^2\Big)^{1/2}.
\]
\end{definition}

\begin{lemma}[Empirical Lipschitz control]\label{lem:emp_lip}
Assume there exists a measurable envelope $L:\mathbb R^d\to[0,\infty)$ such that for all $x,x'\in B_R$ and all $y \in \mathbb R^d$,
\[
\|f_x(y)-f_{x'}(y)\|_2 \;\le\; L(y)\,\|x-x'\|_2.
\]
Then, for any samples $Y_1,\dots,Y_n$,
\[
\|f_x-f_{x'}\|_{L^2(\mathbb P_n)} \;\le\; \|L\|_{L^2(\mathbb P_n)}\,\|x-x'\|_2.
\]
\end{lemma}

\begin{proof}
By the pointwise Lipschitz condition and the definition of $\|\cdot\|_{L^2(\mathbb P_n)}$,
\[
\|f_x-f_{x'}\|_{L^2(\mathbb P_n)}^2
= \frac1n\sum_{i=1}^n \big\|f_x(Y_i)-f_{x'}(Y_i)\big\|_2^2
\le \frac1n\sum_{i=1}^n L(Y_i)^2\,\|x-x'\|_2^2
= \|L\|_{L^2(\mathbb P_n)}^2\,\|x-x'\|_2^2.
\]
The claim follows by taking square roots.
\end{proof}

\begin{lemma}[Bound on empirical Lipschitz envelope]\label{lem:bound_exp_lipschitz}
Let $Y_1,\dots,Y_n$ be i.i.d.\ with law $\nu$. Assume there exists a function $L$ as defined in Lemma \ref{lem:emp_lip} such that $\mathbb E[L(Y_1)^2]<\infty$. Then
\[
\mathbb E\big[\|L\|_{L^2(\mathbb P_n)}\big]
\;\le\; \big(\mathbb E[\|L(Y_1)\|_2^2]\big)^{1/2}
\;=:\; \|L\|_{L^2(\mathbb \nu)}.
\]
\end{lemma}

\begin{proof}
By Cauchy--Schwarz applied to $Z=\|L\|_{L^2(\mathbb P_n)}$: 
\[
\mathbb E\big[\|L\|_{L^2(\mathbb P_n)}\big]
\le \Big(\mathbb E\big[\|L\|_{L^2(\mathbb P_n)}^2\big]\Big)^{1/2}.
\]
Using the fact that $Y_1,\dots,Y_n$ are i.i.d,
\[
\mathbb E\big[\|L\|_{L^2(\mathbb P_n)}^2\big]
= \mathbb E\!\left[\frac1n\sum_{i=1}^n L(Y_i)^2\right]
= \frac1n\sum_{i=1}^n \mathbb E[L(Y_i)^2]
= \mathbb E[L(Y_1)^2].
\]
\end{proof}

\section{Proofs}

\subsection{Proofs of technical Lemmas for Theorem \ref{thm:uniform_convergence} }\label{app:Lemmas_cov_nb}

\begin{lemma}[Bound on the covering number of a product space]\label{lem:cov_number1}
Let $\mathcal F := \{ f_x, x \in \Theta \}$ be a parametric class of Lipschitz functions with respect to $x$. Let $\|\cdot\|_{L^2(\mathbb{P}_n)}^2$ defined in Definition \ref{def:data_distance}. Consider the product space  $\mathcal G = \mathcal F  \times B_1$ and the associated distance defined in Lemma \ref{lem:subgaussian_hdp}. The covering number of the product space is bounded by:
$$\mathcal N(\varepsilon,  \mathcal F \times B_1, d_{\mathcal G}((z_1,x_1),(z_2,x_2))) \le \mathcal N\left(\frac{\varepsilon}{2},  \mathcal F , \| \cdot\|_{L^2(\mathbb P_n)}\right) \times \mathcal N\left(\frac{\varepsilon}{2\|F( \cdot)\|_{L^2(\mathbb P_n)}},  B_1 , \| \cdot\|_2\right).$$
\end{lemma}
\begin{proof}
Let $\mathcal N(\varepsilon,  \mathcal F \times B_1, d_{\mathcal G}((z_1,x_1),(z_2,x_2)))$ be the covering number of $\mathcal G$ for the distance $ d_{\mathcal G}$.

Consider an $\frac{\varepsilon}{2} $-cover of $\mathcal F$ and a $\frac{\varepsilon}{2\|F( \cdot)\|_{L^2(\mathbb P_n)}}$-cover of $B_1$. We can write,
\begin{align*}
    \|z_1^\top f_{x_1}( \cdot) - z_2^\top f_{x_2}( \cdot)\|_{L^2(\mathbb P_n)} &\le \|z_1^\top f_{x_1}( \cdot) - z_1^\top f_{x_2}( \cdot) + z_1^\top f_{x_2}( \cdot) - z_2^\top f_{x_2}( \cdot)\|_{L^2(\mathbb P_n)} \\
     &\le \|f_{x_1}( \cdot) - f_{x_2}( \cdot)\|_{L^2(\mathbb P_n)} + \|z_1 - z_2\|_2  \, \|F( \cdot)\|_{L^2(\mathbb P_n)} \\
     &\le \frac{\varepsilon}{2} + \frac{\varepsilon}{2\|F( \cdot)\|_{L^2(\mathbb P_n)}}\|F( \cdot)\|_{L^2(\mathbb P_n)} = \varepsilon.
\end{align*}
The conclusion of the lemma follows immediately.
\end{proof}
\begin{lemma}[Bound on the covering number]\label{lem:cov_number2}
Let $\mathcal F$ be a parametric class of function $\{ f_x, x \in \Theta \}$ such that any $f_x \in \mathcal F$ is Lipschitz with respect to its index parameter $x$. Let $\|\cdot\|_{L^2(\mathbb{P}_n)}^2$ be defined by Definition \ref{def:data_distance}, then:
$$\mathcal N(\delta, \mathcal F,\| \cdot\|_{L^2(\mathbb P_n)}) \leq \mathcal N\left(\frac{\delta}{\|L\|_{L^2(\mathbb{P}_n)}}, \Theta,  \| \cdot \|_2\right).$$
\end{lemma}
\begin{proof}
Let $x_1, \dots, x_M$ form an $\frac{\delta}{\|L\|_{L^2(\mathbb{P}_n)}}$ cover for $\Theta$. Then $\{f(x_i , \cdot) : i = 1, \dots , M\}$ form an $\delta$ cover for $\mathcal F$. Hence the result.
\end{proof}

The following Lemmas prove that the empirical processes $Z_h$ and $Z_{(z,x)}$ defined in the proof of Theorem \ref{thm:uniform_convergence} in Appendix \ref{app:proof_thm_unif_cv} are sub-Gaussian processes w.r.t. well-defined metrics. 

\begin{lemma}[Sub-gaussian increments in dimension $d$]\label{lem:subgaussian_hdp}
\
Conditionally on the $Y_i$, $Z_{(z,x)} = \frac{1}{n} \sum_{i=1}^{n} \varepsilon_i \langle z, f_x(Y_i)\rangle$ is a sub-Gaussian process with respect to the distance $d_{\mathcal G}((z_1,x_1),(z_2,x_2))$ of the metric space $\mathcal G = B_1  \times \mathcal F$ defined by:
$$d_{\mathcal G}((z_1,x_1),(z_2,x_2)) := \|f_{x_1}( \cdot) - f_{x_2}( \cdot)\|_{L^2(\mathbb P_n)} + \|z_1 - z_2\|_2  \, \|F( \cdot)\|_{L^2(\mathbb P_n)}.$$
\end{lemma}

\begin{proof}
 By Hoeffding's Lemma:
 \begin{align*}
&\mathbb E[\exp(\lambda(Z_{(z_1,x_1)} - Z_{(z_2,x_2)})) | Y_1, \ldots, Y_n] \\
&\qquad \qquad = \prod_{i=1}^{n} \mathbb E\left[\exp\left(\frac{\lambda}{\sqrt{n}} \varepsilon_i (\langle z_1, f_{x_1}( Y_i)\rangle - \langle z_2, f_{x_2}(Y_i)\rangle)\right) \Big| Y_1, \ldots, Y_n\right] \\
&\qquad \qquad \leq \exp\left(\frac{\lambda^2}{2} \|z_1^\top f_{x_1}(\cdot) - z_2^\top f_{x_2}(\cdot)\|_{L^2(\mathbb P_n)}^2\right).
\end{align*}
This norm admits the following expression:
\begin{align*}
    \|z_1^\top f_{x_1}(\cdot) - z_2^\top f_{x_2}(\cdot)\|_{L^2(\mathbb P_n)} &= \|z_1^\top f_{x_1}(\cdot) - z_1^\top f_{x_2}( \cdot) + z_1^\top f_{x_2}( \cdot) - z_2^\top f_{x_2}( \cdot)\|_{L^2(\mathbb P_n)} \\
    &\le \|z_1\|_2 \, \|f_{x_1}( \cdot) - f_{x_2}( \cdot)\|_{L^2(\mathbb P_n)} + \|z_1 - z_2\|_2 \, \|f_{x_2}(  \cdot)\|_{L^2(\mathbb P_n)} \\
     &\le \|f_{x_1}( \cdot) - f_{x_2}( \cdot)\|_{L^2(\mathbb P_n)} + \|z_1 - z_2\|_2  \, \|F( \cdot)\|_{L^2(\mathbb P_n)}.
\end{align*}
This leads to the announced distance on the product space $\mathcal G$.
\end{proof}
\begin{lemma}[Sub-gaussian increments in dimension 1]\label{lem:subgaussian_hdp2}
\
Conditionally on the $Y_i$, $Z_h = \frac{1}{\sqrt{n}} \sum_{i=1}^n \varepsilon_i h_x(Y_i)$ is a sub-Gaussian process with respect to the norm $\| \cdot \|_{\mathcal{L}_2(\mathbb{P}_n)}$.
\end{lemma}
\begin{proof}
The proof follows from Hoeffding's lemma, as for Lemma \ref{lem:subgaussian_hdp}.
\end{proof}

\subsection{Proof of Theorem \ref{thm:uniform_convergence}}\label{app:proof_thm_unif_cv}

The strategy of the proof of Theorem \ref{thm:uniform_convergence} is to analyze the numerator and denominator of the attention ratio separately, and then combine the results. We will prove this Theorem in three parts. We first establish a concentration in expectation and subsequently derive a high-probability bound via Markov’s inequality. These steps rely on classical tools from the empirical processes literature.

\paragraph{Part 1: Uniform concentration of the numerator of $f_n(x)$}
Recall that the continuous attention map is given by
$$f(x) = \frac{\mathbb E[Y \, e^{\langle Ax, Y\rangle}]}{\mathbb E[e^{\langle Ax, Y\rangle}]}.$$
Consider the numerator. It is the expectation of $y \mapsto y \, e^{\langle Ax, y\rangle}$, for an $x \in \mathbb R^d$ , over the sub-Gaussian random vector $Y$.

Define the function class $\mathcal F := \{f_x, \,  x \in B_R \}$, where,$\text{ for all } x\in\mathbb{R}^d, y\in\mathbb{R}^d$, \begin{equation}
    f_x : y \mapsto y\,e^{\langle A x,\; y\rangle} \in L^2(\nu;\mathbb{R}^d).\label{eq:def_f}
\end{equation}
Our goal is to prove a uniform concentration result over the function class $\mathcal F$, indexed by $x$ in the ball $B_R$.

Let $$\|\mathbb{P}_n - \mathbb{P}\|_{\mathcal F} := \sup_{f \in \mathcal F} \big\|\tfrac1n\sum_{i=1}^n f_x(Y_i)
\;-\;\mathbb{E}_{Y\sim\nu}[f_x(Y)]\big\|_2, $$
and define the envelope function of the class $\mathcal F$,
\begin{equation}
    F : y \mapsto \sup_{x \in B_R} \|f_x(y)\|_2 \leq \|y\|_2\, e^{R \|A^\top y\|_2}.\label{eq:F}
\end{equation}
\begin{proposition}[Bound on numerator] \label{prop:numerator}
With probability at least $1 - \delta$, 
$$\|\mathbb{P}_n - \mathbb{P}\|_{\mathcal{F}} \leq  \frac{C}{\delta}\,  d \, (R \|A\|_2 \sqrt{\|\Sigma\|_2d} + 5^{d/2})\cdot \frac{\sqrt{\|\Sigma\|_2} \,}{\sqrt{n}} \cdot e^{\,8R^2\,\|\Sigma^{1/2}A\|_2^2}.
$$
\end{proposition}

\begin{proof}
We will first bound $\mathbb E[\|\mathbb{P}_n - \mathbb{P}\|_{\mathcal F}]$, and then apply
Markov’s inequality to obtain a high-probability concentration bound.

The proof follows two main steps. First, we use a symmetrization lemma (Lemma \ref{lem:sym} below) to relate  $\mathbb E[\|\mathbb{P}_n - \mathbb{P}\|_{\mathcal F}]$ to the Rademacher complexity of the function class $\mathcal F$. Next, we use Dudley's entropy integral together with the Lipschitz property of \(f_x\) to bound this Rademacher complexity via covering numbers.

\step{1}{Symmetrization of $\mathbb E[\|\mathbb{P}_n - \mathbb{P}\|_{\mathcal F}]$}\\
We first derive a symmetrization result that will allow us to link $\mathbb E[\|\mathbb{P}_n - \mathbb{P}\|_{\mathcal F}]$  to the Rademacher complexity of the class $\mathcal F$ defined in Section \ref{sec:empirical_proc_tools}.

\begin{lemma}[Symmetrization of $\mathbb{E}\lbrack\|\mathbb{P}_n-\mathbb{P}\|_{\mathcal F}\rbrack$]\label{lem:sym}
\[
\mathbb{E}\!\left[\left\|\mathbb{P}_n-\mathbb{P}\right\|_{\mathcal F}\right]
\le 2\,\mathfrak{R}_n(\mathcal{F}).
\]
\end{lemma}

\begin{proof}

Let $(X_1, ..., X_n)$ i.i.d. copies of $Y$ independent of $Y$.  
\begin{align*}
\|\mathbb P_n - \mathbb P\|_{\mathcal F} &= \sup_{f \in \mathcal{F}} \left\| \frac{1}{n} \sum_{i=1}^{n} f(Y_i) - \mathbb{E}_Y[f(Y)] \right\|_2 \\
&= \sup_{f \in \mathcal{F}} \left\| \frac{1}{n} \sum_{i=1}^{n} (f(Y_i) - \mathbb{E}_{X_i}[f(X_i)]) \right\|_2 \\
&= \sup_{f \in \mathcal{F}} \left\| \mathbb{E}_X \left[ \frac{1}{n} \sum_{i=1}^{n} f(Y_i) - f(X_i) \right] \right\|_2.
\end{align*}
Using Jensen inequality, we get:
\begin{align*}
\mathbb{E} [\|\mathbb P_n - \mathbb P\|_{\mathcal F}] &= \mathbb{E}_Y \left[ \sup_{f \in \mathcal{F}} \left\| \mathbb{E}_X \left[ \frac{1}{n} \sum_{i=1}^{n} (f(Y_i) - f(X_i)) \right] \right\|_2 \right] \\
&\leq \mathbb{E}_{X,Y} \left[ \sup_{f \in \mathcal{F}} \left\| \frac{1}{n} \sum_{i=1}^{n} (f(Y_i) - f(X_i)) \right\|_2 \right].
\end{align*}
Let \(\varepsilon_1,\dots,\varepsilon_n\) be i.i.d.\ Rademacher signs, independent of \(S =(Y_1,\dots,Y_n)\). The random vector of components $\varepsilon_i (f(Y_i) - f(X_i))$ has the same
joint distribution as the vector with components
$f(Y_i) - f(X_i)$. Hence,
\begin{align*}
\mathbb{E}_{X,Y} \left[ \sup_{f \in \mathcal{F}} \left\| \frac{1}{n} \sum_{i=1}^{n} (f(Y_i) - f(X_i)) \right\|_2 \right] &= \mathbb{E}_{X,Y, \varepsilon} \left[ \sup_{f \in \mathcal{F}} \left\| \frac{1}{n} \sum_{i=1}^{n} \varepsilon_i (f(Y_i) - f(X_i)) \right\|_2 \right] \\
&\leq 2 \mathbb{E}_{Y, \varepsilon} \left[ \sup_{f \in \mathcal{F}} \left\| \frac{1}{n} \sum_{i=1}^{n} \varepsilon_i f(Y_i) \right\|_2 \right] := 2 \mathfrak{R}_n(\mathcal{F}).
\end{align*}
This yields $\mathbb E[\|\mathbb{P}_n - \mathbb{P}\|_{\mathcal F}] \le 2 \mathfrak{R}_n(\mathcal{F})$.
\end{proof}

\step{2}{Bounding the Rademacher complexity $\mathfrak{R}_n(\mathcal{F}$)}\\
Now that we have bounded $\mathbb E[\|\mathbb{P}_n - \mathbb{P}\|_{\mathcal F}]$ by the Rademacher complexity, the next step of the proof is to use Dudley’s entropy integral to bound this Rademacher complexity $\mathfrak{R}_n(\mathcal{F}$).

In order to do that, we will use some useful definitions introduced in Section \ref{sec:empirical_proc_tools}, as well as the Dudley's entropy integral Theorem \ref{thm:dudley}. 

Recall that the Rademacher complexity we want to bound is: 
$$\mathfrak{R}_{n}(\mathcal{F})\;:=\;\mathbb{E}_{S, \varepsilon}[\sup_{x\in B_R}\underbrace{\left\|\frac{1}{n}\sum_{i=1}^n \varepsilon_i\, f_x(Y_i)\right\|_2}_{:=\|Z_x\|_2}] =: \mathbb{E}_{S}[\underbrace{\mathbb{E}_{\varepsilon}\![\sup_{x\in B_R}\left\|Z_x\right\|_2]}_{(\star)}].$$
The strategy of the rest of the proof is to apply Dudley's Theorem \ref{thm:dudley} to $(\star)$. In the one-dimensional case, Dudley’s theorem can be applied directly to the scalar process $\frac{1}{\sqrt{n}}\sum_{i=1}^n \varepsilon_i\, f_x(Y_i)$, which is a sub-Gaussian process conditioned on the $Y_i$ with respect to the associated empirical norm $\|f\|_{L^2(\mathbb P_n)} \;:=\; \Big(\tfrac1n\sum_{i=1}^n f(Y_i)^2\Big)^{1/2}$ (see Lemma \ref{lem:subgaussian_hdp2}). In the present setting, however, the process takes values in $\mathbb R^d$. To use the same concentration inequality, we therefore project it onto scalar directions—that is, we use the dual norm considering: \begin{equation}  Z_{(z,x)} := \frac{1}{n} \sum_{i=1}^{n} \varepsilon_i \langle z, f_x(Y_i)\rangle := z^\top Z_x. \label{eq:sub gaussian process} \end{equation}
We must then verify that this scalar process $Z_{(z,x)}$ remains sub-Gaussian, though, now, with respect to an appropriate metric $d_{\mathcal G}$ defined on the product space $(\mathcal G, d_{\mathcal G})$, where $\mathcal G := \mathcal F \times B_1$ (Lemma \ref{lem:subgaussian_hdp}). Once this sub-Gaussian property is established, Dudley’s theorem \ref{thm:dudley} can be applied to $Z_{(z,x)}$ and the rest of the analysis will require bounds on:
\begin{itemize}
  \item[(i)] the moments of the envelope function~$F$ (Lemma \ref{lem:bound_envelope} ),
  \item[(ii)] the Lipschitz constant of functions in~$\mathcal{F}$, (Lemmas \ref{lem:bound_lipschitz1} and \ref{lem:bound_lipschitz2}), and
  \item[(iii)] the covering number of~$\mathcal{G}$ (Lemmas \ref{lem:cov_number1} and \ref{lem:cov_number2}).
\end{itemize}

In order to do this we will need the definition \ref{def:data_distance} of the empirical distance and Lemmas \ref{lem:emp_lip} and \ref{lem:bound_exp_lipschitz} introduced in Section \ref{sec:empirical_proc_tools}, which we will be referring to. 

\begin{lemma}[Bound on Lipschitz constant for numerator]\label{lem:bound_lipschitz1}
Let $f \in \mathcal F := \{f_x, \,  x \in B_R \}$ as defined in \eqref{eq:def_f}. With an abuse of notations, denote $f : (x,y) \mapsto y\, e^{\langle Ax, y \rangle}$. Then, $f(\cdot,y)$ is Lipschitz with respect to $x$. Let $y \mapsto L(y)$ be its Lipschitz constant. Consider $Y \sim \nu$, a sub-gaussian random vector in $\mathbb R^d$ of matrix parameter $\Sigma$, and scalar parameter $\sigma := \sqrt{\|\Sigma\|_2}$. Then, $L$ verifies
$$\|L\|_{L^2(\mathbb \nu)} \le C ||A||_2 \, 
\sigma^2\, d \, 
\exp\!\big(2\,R^2\|\Sigma^{1/2}A\|_2^2\big),$$
for some constant $C > 0$.
\end{lemma}

\begin{proof}
Consider $f_t(y) := y\, e^{\langle t, y\rangle}$ for $t \in \mathbb R^d$ - we will then apply the result to $t = Ax$ using $\nabla_x f_x(y) = A^\top (\nabla_t f_t)(Ax,y)$.
The Jacobian of $f(\cdot,y)$ at $t$ is
\[
\nabla_t f_t(y)\;= y\,
y^\top \;e^{\langle t, y\rangle}\, \in\mathbb{R}^{d\times d}\,.
\]
Let \[
\|\nabla_t f_t\|_{\mathrm{L^2}(\nu)}:=\Big(\mathbb{E}_{Y\sim\nu}\|\nabla_t f_t(Y)\|_{2}^2\Big)^{1/2} = \Big(\mathbb{E}_{Y\sim\nu}\big[\|Y\|_2^{4} \, e^{2\langle t, Y\rangle}\big]\Big)^{1/2},
\]
where we used the fact that $YY^\top$ is a rank one matrix, hence $\|Y Y^\top \| = \|Y\|_2^2$. 

By Cauchy--Schwarz and the sub-Gaussian property of $Y$,
\begin{align*} 
\|\nabla_t f_t\|_{\mathrm{L^2}(\nu)}
&\le
\Big(\mathbb{E}_{Y\sim\nu}\big[\|Y\|_2^{8}\big]\Big)^{\!1/4} \, \Big(\mathbb{E}_{Y\sim\nu}e^{4\langle t, Y\rangle}\Big)^{\!1/4}\\
&\le
\Big(\mathbb{E}_{Y\sim\nu}\big[\|Y\|_2^{8}\big]\Big)^{\!1/4} \,\exp\!\Big(2\,t^\top\Sigma t\Big).
\end{align*}
Use $t^\top\Sigma t = x^\top A^\top\Sigma A\,x \le \|\Sigma^{1/2}A\|_2^2 \|x\|_2^2$, and take the supremum over $\|x\|_2\le R$:
\[
\sup_{\|x\|_2\le R}\|\nabla_x f_x\|_{\mathrm{L^2}(\nu)}
\ \le\ ||A||_2
\Big(\mathbb{E}_{Y\sim\nu}\|Y\|_2^{8}\Big)^{\!1/4}
\ \exp\!\big(2\,R^2\|\Sigma^{1/2}A\|_2^2\big).
\]
To express the moment factor, 
\[
\Big(\mathbb{E}_{Y\sim\nu}\|Y\|_2^{8}\Big)^{\!1/4} =\Big(\Big(\mathbb{E}_{Y\sim\nu}\|Y\|_2^{8}\Big)^{\!1/8}\Big)^2 \underbrace{\le}_{\star} (c \, \sigma \, \sqrt{d})^2,
\]
where $\star$ is an extension to dimension $d$ of a classical bound on sub-Gaussian moments, with $c$ a universal constant (see Lemma 5.5 in \cite{Vershynin12a}).

Hence
\[
\|L\|_{L^2(\mathbb \nu)} :=\ \sup_{\|x\|_2\le R}\|\nabla_x f_x\|_{\mathrm{L^2}(\nu)}\
 \le C ||A||_2 \, 
\sigma^2\, d \, 
\exp\!\big(2\,R^2\|\Sigma^{1/2}A\|_2^2\big)
\]
which concludes the proof.
\end{proof}
Then, we will need a bound on the envelope function of the class $\mathcal F$.

\begin{lemma}[Bound on envelope function]\label{lem:bound_envelope}
Let $F$ be the envelope function defined in equation \eqref{eq:F}, and $Y \sim \nu$, a sub-gaussian random vector in $\mathbb R^d$ of parameter $\sigma = \sqrt{\|\Sigma\|_2}$. Then,
$$\mathbb E[F(Y)^2]^{1/2} \le  C 5^{d/2}\, \sqrt{d} \, \sigma \,e^{\,8R^2\,\|\Sigma^{1/2}A\|_2^2}.$$
\end{lemma}

\begin{proof}
    By Cauchy--Schwarz,
\[
\mathbb E[F(Y)^2]^{1/2} = \mathbb E\!\big[\|Y\|_2^2\,e^{\,2R\|A^\top Y\|_2}\big]^{1/2}
\ \le\
\big(\mathbb E\|Y\|_2^{4}\big)^{1/4}\;\big(\mathbb E e^{\,4R\|A^\top Y\|_2}\big)^{1/4}.
\]
The first term $\big(\mathbb E\|Y\|_2^{4}\big)^{1/4}$ is controlled by the same argument used for $\Big(\mathbb{E}_{Y\sim\nu}\|Y\|_2^{8}\Big)^{\!1/4} $ in Lemma \ref{lem:bound_lipschitz1}. Now, let $\mathbb S^{d-1}$ be the unit sphere and $U\subset\mathbb S^{d-1}$ an $\varepsilon$-net with
$|U|\le(1+2/\varepsilon)^d$. By a standard estimate (e.g. \cite{Vershynin18}, Lemma 4.4.1, p. 116, 2nd ed.) we have,
\[
\|g\|_2=\sup_{\|u\|=1}\langle g,u\rangle\ \le\ \frac{1}{1-\varepsilon}\max_{u\in U}\langle g,u\rangle,
\]
hence for $g=A^\top Y$, 
\[
e^{\,4R\|A^\top Y\|_2}
\ \le\
\sum_{u\in U}\exp\!\Big(\tfrac{4R}{1-\varepsilon}\,\langle A u, Y\rangle\Big).
\]
Taking expectations and using the sub-Gaussian moment generative function bound of Definition~\ref{def:subGaussian} with $t=\tfrac{4R}{1-\varepsilon}Au$,
\[
\mathbb E\,e^{\,4R\|A^\top Y\|_2}
\ \le\
(1+2/\varepsilon)^d\,
\exp\!\Big(\tfrac12\Big\|\tfrac{4R}{1-\varepsilon}\,\Sigma^{1/2}A\Big\|_2^2\Big)
=
(1+2/\varepsilon)^d\,
\exp\!\Big(\frac{8R^2}{(1-\varepsilon)^2}\,\|\Sigma^{1/2}A\|_2^2\Big).
\]
Finally, for $\varepsilon = \frac12$, 
\[
\big(\mathbb E[F(Y)^2]\big)^{1/2}
\ \le\
C 5^{d/2}\, \sqrt{d} \, \sigma \,e^{\,8R^2\,\|\Sigma^{1/2}A\|_2^2}
\]
which concludes the proof.
\end{proof}
Proposition  \ref{lem:emp_lip} holds for all $f \in \mathcal F$. That is for all $x, x' \in B_R$,
\[
\|f_x-f_{x'}\|_{L^2(\mathbb P_n)}^2
\ \le\ \|L\|_{ L^2(\mathbb{P}_n)}^2 \|x-x'\|_2^2.
\]
We formally prove in Lemma \ref{lem:subgaussian_hdp} that the empirical process defined in equation \eqref{eq:sub gaussian process} is a sub-Gaussian process, with respect to the distance associated to the product space $\mathcal G = (\mathcal F \times B_1)$ defined in Lemma \ref{lem:subgaussian_hdp}. We also need a bound on the covering number of $\mathcal G $. Corresponding Lemmas with their proofs can be found in Appendix \ref{app:Lemmas_cov_nb}.

Now that we have gathered all the necessary tools to bound the Rademacher complexity $\mathfrak{R}_n(\mathcal{F})$ we can end the proof of Proposition \ref{prop:numerator}. 

\textit{Proof of Proposition \ref{prop:numerator}}\\
Let $R_{L, F}=2R\|L\|_{L^2(\mathbb{P}_n)} + 2\|F\|_{L^2(\mathbb{P}_n)}$ and apply Theorem \ref{thm:dudley} to the centered sub-Gaussian process $Z_{(z,x)}$ with respect to the metric space $(\mathcal G, d_{\mathcal G})$, in order to bound $\mathbb E_{\varepsilon}[\sup_{(z,x) \in \mathcal G} Z_{(z,x)}]$,
\begin{align*}
    &\mathbb E_{\varepsilon}[\sup_{(z,x) \in \mathcal G} Z_{(z,x)}] \\
    &\le \frac{C}{\sqrt{n}} \int_{0}^{\text{diam}(\mathcal{G})} \sqrt{\ln \mathcal N(\delta, \mathcal G, d_{\mathcal{G}})} \, d \delta \\
   % &\le \frac{C}{\sqrt{n}} \int_{0}^{R_{L, F}}\sqrt{\ln \mathcal N(\frac{\delta}{2  \|L\|_{L^2(\mathbb{P}_n)}},  B_R , \| \cdot\|_2) \times \mathcal N(\frac{\delta}{2\|F( \cdot)\|_{L^2(\mathbb P_n)}},  B_1 , \| \cdot\|_2)} \, d \delta \\
    &\leq \frac{C}{\sqrt{n}} \int_{0}^{R_{L, F}} \sqrt{\ln \mathcal N\Bigl(\frac{\delta}{2 \|L\|_{L^2(\mathbb{P}_n)}},  B_R , \| \cdot\|_2\Bigr)}\, + \, \sqrt{\ln \mathcal N\Bigl(\frac{\delta}{2\|F( \cdot)\|_{L^2(\mathbb P_n)}},  B_1 , \| \cdot\|_2\Bigr)} \, d \delta,
\end{align*}
where we used Lemma \ref{lem:cov_number1} and \ref{lem:cov_number2}.

Taking the expectation over $Y_i$, and using the symmetrization Lemma \ref{lem:sym},
\begin{align*}
 \mathbb E[\|\mathbb P_n - \mathbb P\|_{\mathcal{F}}] &\leq  \mathfrak{R}_{n}(\mathcal{F}) := \mathbb E_Y \mathbb E_{\varepsilon}[\sup_{(z,x) \in \mathcal G} Z_{(z,x)} | Y_1,..., Y_n]\leq \frac{c}{\sqrt{n}}(\mathcal{I}+\mathcal{J})
\end{align*}
where 
$$
\mathcal{I}= \mathbb E_Y \Big[ \int_{0}^{R_{L, F}} \sqrt{\ln \mathcal N\Bigl(\frac{\delta}{2 \|L\|_{L^2(\mathbb{P}_n)}},  B_R , \| \cdot\|_2\Bigr)}\Big] d\delta,
$$
and
$$
\mathcal{J}=\mathbb E_Y \Big[ \int_{0}^{R_{L, F}} \sqrt{\ln \mathcal N\Bigl(\frac{\delta}{2\|F( \cdot)\|_{L^2(\mathbb P_n)}},  B_1 , \| \cdot\|_2\Bigr)} \, d \delta \Big]\Big).
$$
For the term $\mathcal{I}$, set the change of variable $u = \frac{\delta}{2R \|L\|_{L^2(\mathbb{P}_n)}} $ for $0 \leq u \leq 1$. For the second term $\mathcal{J}$, set $u = \frac{\delta}{2\|F\|_{L^2(\mathbb{P}_n)}}$ 
\[
\mathcal{I}\leq  \mathbb E\left[\int_0^{1 + \frac{\|F\|_n}{R \, \|L\|_n}} \sqrt{\ln \mathcal N\left(Ru,B_R, \|\cdot\|_2\right)} \,2 R \, \|L\|_{L^2(\mathbb P_n)} \, du\right],
\]
\[
\mathcal{J}\leq  \mathbb E\left[\int_0^{1 + \frac{R \, \|L\|_n}{ \, \|F\|_n}} \sqrt{\ln \mathcal N\left(u,B_1, \|\cdot\|_2\right)}  \, 2\|F\|_{L^2(\mathbb P_n)} \, du\right],
\]
where we can denote $\|F\|_n :=\|F\|_{L^2(\mathbb{P}_n)}$ and  $\|L\|_n :=\|L\|_{L^2(\mathbb{P}_n)}$ in the integrand to alleviate the notation. 

Recall that $\ln \mathcal N\left(\varepsilon,B_r, \|\cdot\| \right) \leq d \ln\left(1 + \frac{2 \, r}{\varepsilon}\right) $ (see Corollary 4.2.11. in \cite{Vershynin18}), hence 
\begin{align}
\mathcal{I} \leq  4\sqrt{2d}\, R\,\mathbb E[\|L\|_{L^2(\mathbb P_n)}] + 2\sqrt{ d \ln 3 }\, \mathbb E[\|F\|_{L^2(\mathbb P_n)}].\label{eq:star}
\end{align}
The same steps for the second term give,
\begin{align}
\mathcal{J} \leq  4\sqrt{2d}\,  \mathbb E[\|F\|_{L^2(\mathbb P_n)}] + 2R \, \sqrt{ d \ln 3 }\, \mathbb E[\|L\|_{L^2(\mathbb P_n)}].\label{eq:star star}
\end{align}
Detailed derivations of \eqref{eq:star} and \eqref{eq:star star} can be found in Appendix \ref{app:details_eq_star}.
Finally, 
\begin{equation}
    \mathbb E[\|\mathbb P_n - \mathbb P\|_{\mathcal{F}}] \leq  \frac{c \, \sqrt{d}}{\sqrt{n}} \Big(  R\, \mathbb E[\|L\|_{L^2(\mathbb P_n)}] + \, \mathbb E[\|F\|_{L^2(\mathbb P_n)}]\Big).\label{eq:bound_lip_envel}
\end{equation}
Applying Lemma \ref{lem:bound_exp_lipschitz}, $\mathbb E[\|L\|_{L^2(\mathbb P_n)}] \leq \left(\mathbb E\left[\|L\|_{L^2(\mathbb P_n)}^2\right]\right)^{1/2} = \left(\mathbb E[L(Y_1)^2]\right)^{1/2} = \|L\|_{L^2(\mathbb \nu)},$ and using Lemma \ref{lem:bound_lipschitz1}, along with Lemma \ref{lem:bound_envelope} it comes, 
% \textcolor{red}{Dans le lemme \ref{lem:bound_envelope} c'est $L^2(P)$, alors qu'ici c'est $L^2(\mathbb{P}_n)$, à revérifier.}\textcolor{olive}{oui je majore d'abord par $\|L\|_{L^2(\mathbb \nu)}$, puis j'applique les lemmes qui portent sur $\|\cdot\|_{L^2(\mathbb \nu)}$}
\begin{align*}
    \mathbb E[\|\mathbb P_n - \mathbb P\|_{\mathcal{F}}] &\leq  \frac{c \, \sqrt{d}}{\sqrt{n}} \Big( R\, C_1 \, \|A\|_2 
\sigma^2\,d
e^{2R^2\|\Sigma^{1/2}A\|_2^2} + 5^{d/2} C_2
\sigma\,\sqrt{d}\,e^{\,8R^2\,\|\Sigma^{1/2}A\|_2^2}\Big) \\
&\leq \frac{C \,d\,   (R \|A\|_2 \sigma \sqrt{d} + 5^{d/2})\sigma \,}{\sqrt{n}} \,e^{\,8R^2\,\|\Sigma^{1/2}A\|_2^2}=: \frac{C_0(\Sigma, \, A,  \, \delta,  \, R)}{\sqrt{n}}.
\end{align*}
for some positive constant $C$.

To conclude, we apply the Markov inequality, which yields, with probability at least $1 - \delta$, 
$\|\mathbb P_n - \mathbb P\|_{\mathcal{F}} \le \frac{1}{\delta}\, \mathbb E[\|\mathbb P_n - \mathbb P\|_{\mathcal{F}}] \le \frac{C_0(\Sigma, \, A,  \, \delta,  \, R)}{\delta \cdot \sqrt{n}}. $
This ends the proof of Proposition \ref{prop:numerator}.
\end{proof}

\paragraph{Part 2: Uniform concentration of the denominator empirical process} 
An analogous result as Proposition \ref{prop:numerator} holds for the denominator with an identical proof using the function class 
$$
\mathcal F' := \{ h_x : y \mapsto e^{\langle Ax, y\rangle}, x \in B_R\}.
$$
\begin{proposition}[Bound on denominator] \label{prop:denominator}
For all $R>0$ and $\delta \in (0,1)$, with probability at least $1 - \delta$, 
$$\|\mathbb P_n - \mathbb P\|_{\mathcal{F'}} \leq  \frac{C'}{\delta} \cdot \frac{\|A\|_2 \, \sigma \sqrt{d}}{\sqrt{n}} e^{2\,R^2\|\Sigma^{1/2}A\|_2^2}.
$$
\end{proposition}
\begin{proof}
The strategy of the proof follows the one of Proposition \ref{prop:numerator}, in dimension 1. The first step of symmetrization is identical. Then, we detail a few differences compared to the numerator. 

First, the Lipschitz constant calculation slightly differs, as shown in Lemma \ref{lem:bound_lipschitz2} below. 
\begin{lemma}[Bound on Lipschitz constant for denominator]\label{lem:bound_lipschitz2} 
Let $f \in \mathcal F'$, and $L$ defined in Lemma \ref{lem:emp_lip}
$$\|L\|_{L^2(\mathbb \nu)} \le C \|A\|_2
\sigma \sqrt{d}\,
\exp\!\big(2\,R^2\|\Sigma^{1/2}A\|_2^2\big),$$
for some constant $C > 0$.
\end{lemma}
\begin{proof}
Here, we apply the exact same proof as for Lemma \ref{lem:bound_lipschitz1}, with $h(t,y) :=  e^{\langle t, y\rangle}$, using $\Big(\mathbb{E}_{Y\sim\nu}\|Y\|_2^{4}\Big)^{\!1/4}
\ \le C
\sigma \sqrt{d}$.
\end{proof}
Then, the sub-Gaussian process under consideration is now
\begin{equation}
    Z_h = \frac{1}{\sqrt{n}} \sum_{i=1}^n \varepsilon_i h_x(Y_i),\label{eq:Zh}
\end{equation}
(see Lemma \ref{lem:subgaussian_hdp2} for justification).

In dimension 1, we apply Theorem \ref{thm:dudley} to the sub-Gaussian process $Z_h$ - defined in equation \eqref{eq:Zh} - with respect to the metric space $(\mathcal F', \| \cdot \|_{\mathcal{L}_2(\mathbb{P}_n)})$ to bound $\mathbb E_{\varepsilon}[\sup_{h \in \mathcal F'} Z_h]$. Note that Dudley's entropy bound extends to $\mathbb E_{\varepsilon}[\sup_{h \in \mathcal F'} |Z_h|]$ using that $\mathbb E_{\varepsilon}[\sup_{h \in \mathcal F'} |Z_h|] = \sup_{h \in \mathcal F'} \max \{Z_h, - Z_h \}$ (both statement can be found in the literature). Then,
\begin{align*}
    \mathbb E[\sup_{h \in \mathcal F'} |Z_h|] &\le \frac{C}{\sqrt{n}} \int_{0}^{\text{diam}(\mathcal{F'})} \sqrt{\ln N(\delta, \mathcal F', L^2(\mathbb P_n))} \, d \delta \\
    &\le \frac{C}{\sqrt{n}} \int_{0}^{\text{diam}(B_R) \|L\|_{L^2(\mathbb{P}_n)}} \sqrt{\ln N(\frac{\delta}{\|L\|_{L^2(\mathbb{P}_n)}}, B_R,  \| \cdot \|_2)} \, d \delta,
\end{align*}
where we used Lemma \ref{lem:cov_number2}.

Taking the expectation over $Y_i$, and using the symmetrization Lemma \ref{lem:sym} (see Appendix \ref{app:details_dudley} for details), 
\begin{align}
 E[\|\mathbb P_n - \mathbb P\|_{\mathcal{F'}}] \leq \frac{C \, \|A\|_2 \, 
\sigma \sqrt{d}}{ \delta \, \sqrt{n}} \, 
e^{2 \,R^2\|\Sigma^{1/2}A\|_2^2} := C_0'(\Sigma, A, R; \delta).\label{eq:dudley_den_eq}
\end{align}
Finally we also conclude using Markov inequality. 
\end{proof}

\paragraph{Part 3: Uniform concentration of the attention map (ratio bound)}
The last step of the proof is to combine Propositions \ref{prop:numerator} and \ref{prop:denominator} to bound the attention's ratio. 
% In the sequel, $x\in B_R$ is fixed, and thus omitted in the notation most of the time. 
For the sake of clarity, we 
% simplify the notation for both
denote the empirical and continuous attention's numerator and denominator: $N_n(x) := \frac1n \sum_{i=1}^n y_i e^{\langle Ax, y_i\rangle}$, $D_n(x):= \frac1n \sum_{i=1}^n e^{\langle Ax, y_i\rangle}$, and $N(x) := \int y e^{\langle Ax, y\rangle} d\nu(y)$, $D(x):= \int e^{\langle Ax, y\rangle} d\nu(y)$. In the sequel, we omit the dependence in $x\in B_R$ most of the time to alleviate notation. Then, we can write: $f_n(x) = \|V\|_2 \cdot \frac{N_n}{D_n}$ and $f(x) =  \|V\|_2 \cdot \frac{N}{D}$. 
Putting everything on the same denominator yields: 
\begin{align*}
\|\frac{N_n}{D_n} - \frac{N}{D}\|_2 &= \|\frac{N_nD - ND + ND - ND_n}{DD_n}\|_2\le \frac{\|N_n-N\|_2}{D_n} + \frac{\|N\|_2 \, |D_n - D|}{DD_n}.
\end{align*}
To bound below the denominator $D_n(x)$ uniformly in $x$, we define the following additional events:
\begin{align*}
    \mathcal{E}_3 &=  \{\sup_{x \in B_R} |D_n(x)-D(x)| \le \frac12\}\\
    \mathcal{E}_4 &= \{\forall x \in B_R,  D_n(x) \ge \frac12\} = \{\inf_{x \in B_R} D_n(x) \ge \frac12\}
\end{align*}
First, notice that $\mathcal{E}_3 \subseteq \mathcal{E}_4$. Let $\omega \in \mathcal{E}_3$. For all $x \in B_R$, $D_n(x, \omega) \ge D(x) - |D_n(x, \omega) - D(x)| \ge \frac12$, using $D(x) = \mathbb E[e^{\langle Ax, Y\rangle}] \ge 1$ by Jensen's inequality. Hence $\mathcal{E}_3 \subseteq \mathcal{E}_4$.

By Proposition \ref{prop:denominator}, if 
$$
C_2(R) := \frac{C'}{\delta_1} \cdot \frac{\|A\|_2 \, \sigma \sqrt{d}}{\sqrt{n}} e^{2\,R^2\|\Sigma^{1/2}A\|_2^2}
$$
then the event 
\begin{align*}
    \mathcal{E}_2 &= \{\sup_{x \in B_R} |D_n(x)-D(x)| \le C_2(R)\}
\end{align*}
is verified with probability $1-\delta_1$.
%,where $C_1(R)$ and $C_2(R)$ come from Propositions \ref{prop:numerator} and \ref{prop:denominator}.
Note that, for 
$$n \ge n_{\min}(\delta, \Sigma, A, R, d):= 4 \frac{C'^2}{\delta_1^2} \|A\|_2^2\sigma^2 d e^{4 R^2 \|\Sigma^{1/2}A\|_2^2}\,, $$
one has $C_2(R) \le \frac{1}{2}$, hence $\mathcal{E}_2 \subseteq \mathcal{E}_3$, and therefore $\mathbb{P}(\mathcal{E}_3) \ge \mathbb{P}(\mathcal{E}_2) \ge 1 - \delta_1$. 
On $\mathcal{E}_3$, the event $\mathcal{E}_4$ is also verified by the inclusion $\mathcal{E}_3 \subseteq \mathcal{E}_4$, which allows us to bound $D_n(x) \ge \frac{1}{2}$ uniformly in $x \in B_R$.
Therefore, with probability $1 - \delta_1$, and with $n \ge n_{min}(\delta_1, \Sigma, A, R, d)$, 
\begin{equation}
\sup_{x \in B_R}\|\frac{N_n}{D_n} - \frac{N}{D}\|_2 \le 2 (\sup_{x \in B_R}\|N_n - N\|_2 + \sup_{x \in B_R}\|N\|_2 \cdot C_2(R)). \label{eq:eq1}\end{equation}
where we used $\sup_{x \in B_R}(\|N\|_2 \cdot |D_n(x)-D(x)|) \le \sup_{x \in B_R}\|N\|_2 \cdot \sup_{x \in B_R}|D_n(x)-D(x)|$, and $\sup_{x \in B_R}|D_n(x)-D(x)| \le C_2(R)$ on $\mathcal{E}_2$.

By Proposition \ref{prop:numerator}, the event
$$
\mathcal{E}_1 =  \{\sup_{x \in B_R} \|N_n(x)-N(x)\|_2 \le C_1(R)\} 
$$
is verified with probability $1 - \delta_2$, where
\begin{equation}
 C_1(R) := \frac{C_1}{\delta_2} \, d\, (R \|A\|_2 \sqrt{\|\Sigma\|_2d} + 5^{d/2})\cdot \frac{\sqrt{\|\Sigma\|_2} \,}{\sqrt{n}} \cdot e^{\,8R^2\,\|\Sigma^{1/2}A\|_2^2}.\label{eq:eq2}\end{equation}
Then, to bound $\|N\|_2$, notice that: $\|N\|_2 = \|\mathbb E[Y \, e^{\langle Ax, Y\rangle}]\|_2 \le \operatorname{tr}(\Sigma)^{1/2} \, e^{2 R^2\, \|\Sigma^{\frac12} A\|_2^2}.$
Combining events $\mathcal{E}_1$ and $\mathcal{E}_2$, and for $n \ge n_{min}(\delta, \Sigma, A, R, d)$, we have with probability $1-\delta$ (taking $\delta_1 = \delta_2 =  \frac{\delta}{2})$, 
$$\sup_{x \in B_R}\|\frac{N_n}{D_n} - \frac{N}{D}\|_2 \le \|V\|_2 \cdot\frac{2 C \, \sqrt{d}\, (R \|A\|_2 \sqrt{\|\Sigma\|_2}d + \operatorname{tr}(\Sigma)^{1/2} \|A\|_2 + 5^{d/2}\sqrt{d})}{\delta } \cdot \frac{\sqrt{\|\Sigma\|_2}}{\sqrt{n}} \cdot e^{8 \,R^2\|\Sigma^{1/2}A\|_2^2}.
$$
This concludes the proof of Theorem \ref{thm:uniform_convergence}.

\subsection{Equations of Theorem \ref{thm:uniform_convergence}}
Below, we detail the derivation of some inequalities in the proof Theorem \ref{thm:uniform_convergence} (Appendix \ref{app:proof_thm_unif_cv}). We recall that $\ln \mathcal N\left(\varepsilon,B_r, \|\cdot\| \right) \leq d \ln\left(1 + \frac{2 \, r}{\varepsilon}\right) $.
\paragraph{Equation \eqref{eq:star}.}\label{app:details_eq_star}
Let $t = \frac{\|F\|_n}{R \, \|L\|_n}$, then
\begin{align*}
\mathcal{I} &\leq  2R \, \mathbb E\left[\|L\|_{L^2(\mathbb P_n)} \int_0^{1+t} \sqrt{ d \ln\!\left(1+\frac{2}{u}\right) }\,du \right]\nonumber \\
&= 2R \, \mathbb E\left[\|L\|_{L^2(\mathbb P_n)} \left( \int_0^{1} \sqrt{ d \ln\!\left(1+\frac{2}{u}\right) }\,du + \int_1^{1+t} \sqrt{ d \ln\!\left(1+\frac{2}{u}\right) }\,du \right) \right]\nonumber  \\
&\leq 2R \, \mathbb E\left[ \|L\|_{L^2(\mathbb P_n)} \left( \int_0^{1} \frac{\sqrt{2d}}{\sqrt u}\,du + \int_1^{1+t} \sqrt{ d \ln 3 }\,du \right) \right]\nonumber \\
&\leq 2R \, \mathbb E\left[\|L\|_{L^2(\mathbb P_n)} \left( 2\sqrt{2d} + t \sqrt{ d \ln 3 } \right) \right]\nonumber  \\
&=  4\sqrt{2d}\, R\,\mathbb E[\|L\|_{L^2(\mathbb P_n)}] + 2\sqrt{ d \ln 3 }\, \mathbb E[\|F\|_{L^2(\mathbb P_n)}].
\end{align*}
\paragraph{Equation \eqref{eq:star star}.}\label{app:details_eq_star_star}
Let $t' = \frac{R \, \|L\|_n}{ \, \|F\|_n}$, then by the same arguments as above,
\begin{align*}
\mathcal{J} &\leq   \, \mathbb E\left[2\|F\|_{L^2(\mathbb P_n)} \int_0^{1 + \frac{R \, \|L\|_n}{ \, \|F\|_n}} \sqrt{d \ln\left(1 + \frac{2}{u}\right)} \, du\right] \nonumber \\
&\leq  \mathbb E\!\left[  \|F\|_{L^2(\mathbb P_n)} \left( 4\sqrt{2d} + 2t' \sqrt{ d \ln 3 } \right) \right]\nonumber \\
&=4\sqrt{2d}\,  \mathbb E[\|F\|_{L^2(\mathbb P_n)}] + 2R \, \sqrt{ d \ln 3 }\, \mathbb E[\|L\|_{L^2(\mathbb P_n)}].
\end{align*}
\paragraph{Equation \eqref{eq:dudley_den_eq}.}
\label{app:details_dudley}
Taking the expectation over $Y_i$, and using the symmetrization Lemma \ref{lem:sym}, 
\begin{align*}
\mathbb{E}[\|\mathbb P_n - \mathbb P\|_{\mathcal{F}}] &\leq  \mathbb E_Y \mathbb E_{\varepsilon}[\sup_{h \in \mathcal F'}  |Z_h| \mid Y_1,..., Y_n] \nonumber \\
&\leq  \frac{C}{\sqrt{n}} \mathbb E_Y \left[\int_{0}^{2R \, \|L\|_{L^2(\mathbb{P}_n)}} \sqrt{\ln \mathcal{N}\Bigl(\frac{\delta}{\|L\|_{L^2(\mathbb{P}_n)}}, B_R,  \| \cdot \|_2\Bigr)} \, d \delta\right] \nonumber \\ 
&\le \frac{C}{\sqrt{n}} \mathbb E_Y\left[\int_0^2 \sqrt{\ln \mathcal{N}\left(Ru, B_R, \|\cdot\|_2\right)} \, R \, \|L\|_{L^2(\mathbb P_n)} \, du\right]\nonumber \\
&\le \frac{C}{\sqrt{n}} R \, \mathbb E_Y\left[\|L\|_{L^2(\mathbb P_n)} \int_0^2 \sqrt{d \ln\left(1 + \frac{2}{u}\right)} \, du\right]\nonumber  \\
&\le \frac{C}{\sqrt{n}} R \int_0^2 \sqrt{\frac{2d}{u}} \, du \, \mathbb E_Y[\|L\|_{L^2(\mathbb P_n)}]\nonumber  \\
&\le \frac{c}{\delta} \, \frac{\left(\mathbb E_Y[L(Y_1)^2]\right)^{1/2}}{\sqrt{n}}\\
&= \frac{C \, \|A\|_2 \, 
\sigma \sqrt{d}}{ \delta \, \sqrt{n}} \, 
\exp\!\big(2 \,R^2\|\Sigma^{1/2}A\|_2^2\big).
\label{eq:dudley_den_eq_dl}
\end{align*}
The last inequality follows directly from Lemma \ref{lem:bound_exp_lipschitz} and Lemma \ref{lem:bound_lipschitz1}.

\subsection{Generalization of \cref{thm:uniform_convergence}}\label{app:uniform_cv_B}

\begin{theorem}[Uniform convergence of the attention map for sub-Gaussian tokens over a general set]\label{thm:uniform_cv_B}
Let $X \sim \nu$ be centered and sub-Gaussian with parameter matrix $\Sigma \succ0$. Let $B$ an invertible matrix, let $R > 0$ and denote $K_R := \{x \in \mathbb{R}^d |  \|Bx\|_2 \le R \}$. For any $\delta > 0$, there exists a constant $C>0$, such that for $n \ge n_{min}(\delta, \Sigma, A, R) := 4e^{2 R^2 \|\Sigma^{1/2}AB^{-1}\|_2^2}\, \Bigl(\frac{1}{\delta} - 1\Bigr)$, with probability at least $1 - \delta$,
\begin{equation}\label{eq:uniform_convergence_B}
  \sup_{x \in K_R}\bigl\| f_n(x) -f(x) \bigr\|_2
\;\le\;  q_{(\Sigma, A, V, R, d, \delta)} \cdot \frac{e^{8 \,R^2\|\Sigma^{1/2}AB^{-1}\|_2^2}}{\sqrt{n}} ,
\end{equation}
where $q_{(\Sigma, A, V, R, d, \delta)}:= \frac{C \, \sqrt{d}}{\delta} \|V\|_2  \cdot \|\Sigma\|_2^{1/2} \cdot(R \|A\|_2 \|\Sigma\|_2^{1/2}  d + \operatorname{tr}(\Sigma)^{1/2} \|A\|_2 + 5^{d/2}\sqrt{d}).$ 
\end{theorem}
\begin{proof}
It suffices to apply Theorem 4.2 with $\tilde{A} := AB^{-1}$, and taking the supremum over $y \in \text{Im} B \cap B_R$, which is a compact set. 
\end{proof}

\subsection{Sharp bounds for two Dirac masses}\label{app:diracs}

In this Appendix, we show that when $\nu$ is a sum of two Dirac masses, the bound provided in Theorem \ref{thm:uniform_convergence} is not sharp with respect to the parameters $R$ and $R_{0}$, and one can actually prove a sharp dependence rate in these parameters. 
We consider the case $d=1$, $A=\beta{\rm Id}$ (i.e., $A=\beta\in\mathbb{R}$ is a scalar).
Let $\nu = p\delta_{-R_{0}}+(1-p)\delta_{R_{0}}$ for some  $p\in(0,1)$ and $R_{0}>0$. When $p=1/2$, the sub-Gaussian parameter $\Sigma$ is equal to $R_0^2$. Therefore Theorem \ref{thm:uniform_convergence} yields a bound of the form $\frac{e^{8\beta^2 R_0^2R^2}}{\sqrt{n}}$. We show below a sharp bound which is polynomial in $R_{0}$ and does not depend on $R$. Emphasising the dependence in $p$ and $R_{0}$, we write $f(p,R_{0},x)$ and $f_n(p,R_{0},x)$ instead of $f(x)$ and $f_n(x)$. 

\begin{theorem}
There exist $C,C'$ universal (in particular, independent of $p,\delta,R_{0}$) such that for any $\delta>0$, any $p\in (0,1), R, R_{0}>0$, and any $n\geq \frac{CR_0^2}{\delta\min(p^6,(1-p)^6)}$, there holds  with probability at least $1-\delta$
\begin{equation}\label{e:2Diracs}
C'\delta \sqrt{p(1-p)} \frac{R_{0}^2}{\sqrt{n}}\leq  \sup_{x\in B_{R}} |f(p,R_0,x)-f_n(p,R_0,x)|\leq  \sqrt{2\log(2/\delta)} \frac{R_{0}^2}{\sqrt{n}}.
\end{equation}
.for any $R>0$.
\end{theorem}
\begin{proof}
We have
\begin{equation}\label{e:fprx}
f(p,R_{0},x) =\frac{(1-p)e^{\beta R_{0}x}R_{0}-pe^{-\beta R_{0}x}R_{0}}{(1-p)e^{\beta R_{0}x}+pe^{-\beta R_{0}x}}=R_{0}-2R_{0}\frac{1}{1+\frac{1-p}{p}e^{2\beta R_{0}x}}.
\end{equation}
Let us now compute $f_n(p,R_{0},x)$ for some $n\in\mathbb{N}$. Since $\nu_n$ is obtained by taking $n$ random draws according to $\nu$, we denote by $m$ the (random) number of times that $-R_{0}$ appears, and thus $n-m$ is the number of times that $R_{0}$ appears. Thus $\nu_n=\frac{m}{n}\delta_{-R_{0}}+\frac{n-m}{n}\delta_{R_{0}}$. We let $p'=\frac{m}{n}$. It is not difficult to verify that
$$
f_n(p,R_{0},x)=f(p',R_{0},x).
$$
Our goal is to estimate the quantity $|f(p',R_{0},x)-f(p,R_{0},x)|$.

\textbf{Lower bound.}
We have $ \sup_{x\in B_{R}} |f(p,R_{0},x)-f_n(p,R_{0},x)|\geq |f(p,R_{0},0)-f_n(p,R_{0},0)|=2R_{0}|p-p'|$ according to \eqref{e:fprx}. By Berry-Esseen's theorem we have for some $c>0$ independent of $R_{0}$ and $p$
$$
\forall \varepsilon>0, \qquad \mathbb{P}\Bigl(|p'-p|\geq \varepsilon\frac{R_{0}\sqrt{p(1-p)}}{\sqrt{n}}\Bigr)\geq 1-c\Bigl(\varepsilon+\frac{1}{\sqrt{p(1-p)n}}\Bigr).
$$
Hence, with probability $\geq 1-\delta$, we have $|p'-p|\geq \frac{\delta}{c}\frac{R_{0}\sqrt{p(1-p)}}{\sqrt{n}}-\frac{R_{0}}{n}$. Our assumption on $n$ then implies that $|p'-p|\geq \frac{\delta}{2c}\frac{R_{0}\sqrt{p(1-p)}}{\sqrt{n}}$ (by choosing $C$ large) and finally the lower bound in \eqref{e:2Diracs} follows.

\textbf{Upper bound.} 
By Taylor's remainder formula,
$$
|f_n(p',R_{0},x)-f(p,R_{0},x)|\leq |p'-p||\frac{\partial f}{\partial p}(p,R_{0},x)|+ \frac12 |p'-p|^2 \|\frac{\partial^2 f}{\partial p^2}(\cdot,R_{0},x)\|_{L^\infty([p,p'])}.
$$
It follows from \eqref{e:fprx} that 
$$
\frac{\partial f}{\partial p}(p,R_{0},x)=-\frac{2R_{0}}{(pe^{-\beta R_{0}x}+(1-p)e^{\beta R_{0}x})^2}\qquad \text{and} \qquad \frac{\partial f^2}{\partial p^2}(q,R_{0},x)=\frac{4R_{0}e^{2\beta R_{0}x}(1-e^{2\beta R_{0}x})}{(q+(1-q)e^{2\beta R_{0}x})^3}.
$$
Hence
$$
|\frac{\partial f}{\partial p}(p,R_{0},x)|\leq R_{0} \qquad \text{and} \qquad |\frac{\partial f^2}{\partial p^2}(q,R_{0},x)|\leq 4R_{0}\max(q^{-3},(1-q)^{-3})
$$
(the second inequality follows by treating separately then cases $x\geq 0$ and $x\leq 0$).
Due to Hoeffding's inequality we have 
$$
\mathbb{P}(|p'-p|\geq \varepsilon)\leq 2\exp(-2n\varepsilon^2/R_{0}^2)
$$
hence on an event $A_\delta$ of probability $\geq 1-\delta$ we have 
\begin{equation}\label{e:p'-p}
|p'-p|\leq \frac{R_{0}\sqrt{\log(2/\delta)}}{\sqrt{2n}}.
\end{equation}
Our assumption on $n$ implies that $n\geq  \frac{2R_{0}^2\log(2/\delta)}{\min(p^2,(1-p)^2)}$ hence we have under the event $A_\delta$ that $p'\geq \frac{p}{2}$ and $1-p'\geq \frac{1-p}{2}$. We deduce easily that under $A_\delta$ there holds 
\begin{equation}\label{e:pfpp2est}
\|\frac{\partial^2 f}{\partial p^2}(\cdot,R_{0},x)\|_{L^\infty([p,p'])}\leq  32R_0\Bigl(\frac{1}{p^3} + \frac{1}{(1-p)^3}\Bigr).
\end{equation}
In particular, with probability $1-\delta$, up to choosing a larger $C_1$,
$$
|f(p',R_{0},x)-f(p,R_{0},x)|\leq \frac{R_{0}^2\sqrt{\log(2/\delta)}}{\sqrt{2n}}+8\frac{R_{0}^3\log(2/\delta)}{n}\Bigl(\frac{1}{p^3} + \frac{1}{(1-p)^3}\Bigr)
$$
By our assumption on $n$, we have $\frac{R_{0}^2\sqrt{\log(2/\delta)}}{\sqrt{2n}}\geq 8\frac{R_{0}^3\log(2/\delta)}{n}\Bigl(\frac{1}{p^3} + \frac{1}{(1-p)^3}\Bigr)$, thus 
$$
|f(p',R_{0},x)-f(p,R_{0},x)|\leq  \frac{R_{0}^2\sqrt{2\log(2/\delta)}}{\sqrt{n}}.
$$
This gives the upper bound in \eqref{e:2Diracs}.
\end{proof}

\begin{remark}
The same example of two Dirac masses can be considered in any dimension, and the lower and upper bounds are obviously the same, without any dimensional dependence.
\end{remark}

\subsection{Proof of Theorem \ref{thm:convergence_rate_lipschitz}}\label{app:proof_cv_lip}

\begin{proof}
 Let $h$ a Lipschitz function of parameter $L_0$. By considering $h - h(0)$ instead of $h$ if necessary, we can assume without loss of generality that $h(0) = 0$.

\step{1}{Decomposition of the error.}
We reformulate the mean error to facilitate our analysis.  
The quantity of interest is
\begin{equation*}
  \Bigl\| \mathbb{E}_n[h \circ f_n(\hat X)]
    -\mathbb{E}[h \circ f(X)] \Bigr\|_2 .
\end{equation*}
Applying the triangle inequality and adding–subtracting identical terms,
we obtain
\begin{align*}
  &\qquad\qquad \Bigl\| \mathbb{E}_n[h \circ f_n(\hat X)]
      -\mathbb{E}_x[h \circ f(X)] \Bigr\|_2 \\
      &\qquad\quad = \Bigl\|\mathbb{E}_n[h \circ f_n(\hat X)] - \mathbb{E}_n [h \circ f(X)] +   \mathbb{E}_n [h \circ f(X)] - \mathbb{E} [h \circ f(X)]\Bigr\|_2 \nonumber \\
  &\qquad\quad \le \mathcal{I}+\mathcal{J}\nonumber
\end{align*}
where 
$$
\mathcal{I}=\Bigl\|\mathbb{E}_n[h \circ f_n(\hat X)-h \circ f(X)]\Bigr\|_2
$$
and
$$
\mathcal{J}=\Bigl\|\mathbb{E}_n
                [h \circ f(X)]-\mathbb{E}
                [h \circ f(X)]\Bigr\|_2
$$

\step{2}{Analysis of the second term $\mathcal{J}$.}
To bound $\mathcal J$, we follow the same methodology as in the proof of Proposition \ref{prop:ptwise_convergence}: we apply the CLT, followed by the derivation of an asymptotic confidence bound using the quantile of the $\chi^2(d)$ law (see Appendix \ref{app:ptw_cv} for more details). We obtain that, with probability at least $1-\delta$,
$$\mathcal J \le \frac{q_{d,1- \delta}^{1/2} \, \lambda_{\max}( \operatorname{Cov}(h \circ f(X)))^{1/2}}{\sqrt n},$$
where the bound scales with the $\chi_2$ quantile $q_{d,1- \delta}$, and the maximum variance direction of $h \circ f(X)$. 

Let $\alpha := 2\,\|V\|_{2}\,
\|\Sigma^{1/2}\|_2\,
\bigl\|\Sigma^{1/2}A\bigr\|_{2}$, and let us show that 
\begin{equation*}
  \lambda_{\max}(\operatorname{Cov}(h \circ f(X)))
  \le
  \ \alpha^2\,L_0^{2} \cdot \|\Sigma\|_2,
\end{equation*}
from which $\mathcal{J}=\mathcal{O}(n^{-1/2})$ follows.

Using the at-most-linear growth of $\Gamma_\nu $ of Lemma \ref{lem:growth_gamma_sub}, we get,
\begin{align*}
  &\lambda_{\max}(\operatorname{Cov}(h \circ f(X)))\\
  &\qquad \qquad \le \operatorname{tr} (\operatorname{Cov}(h \circ f(X))) \\
&\qquad \qquad= \operatorname{tr} (\mathbb E[(h \circ f(X) - \mathbb E [h \circ f(X)])(h \circ f(X) -\mathbb E [h \circ f(X)] )^\top]) \\
  &\qquad \qquad= \mathbb E[\operatorname{tr} ((h \circ f(X) - \mathbb E [h \circ f(X)])^\top(h \circ f(X) -\mathbb E [h \circ f(X) ]))] \\
  &\qquad \qquad= \mathbb E[\|h \circ f(X) -\mathbb E [h \circ f(X)]\|_2^2 ] \\
  &\qquad \qquad\le \mathbb E[\|h \circ f(X)\|_2^2 ] \le \alpha^2\,L_0^{2} \cdot \mathbb{E}\!\bigl[\|X\|_2^{2}\bigr] =  \alpha^2\,L_0^{2} \cdot \|\Sigma\|_2.
\end{align*}

\step{3}{Analysis of the first term $\mathcal{I}$.}
To bound the first term $\mathcal{I}$, the key insight is to split the analysis based on whether $\|Bx_i\|_2$ exceeds a threshold $R$ or not, where $B$ is any matrix satisfying the hypothesis of \cref{thm:uniform_cv_B}. The threshold on $R$ will then be optimized later. We have
\begin{align}
  \mathcal{I}=
  \Bigl\|\frac1n\sum_{i=1}^n
          h \circ f_n(x_i)-h \circ f(x_i)\Bigr\|_2\le \mathcal{I}_1+\mathcal{I}_2+\mathcal{I}_3,\label{eq:first_bound}
\end{align}
where 
\begin{align*}
\mathcal{I}_1&=\frac1n\sum_{|Bx_i|<R} \Bigl\|h \circ f_n(x_i)-h \circ f(x_i)\Bigr\|_2\\
\mathcal{I}_2&=\frac1n\sum_{|Bx_i|>R}
\Bigl\|h \circ f_n(x_i)\Bigr\|_2\\
\mathcal{I}_3&=\frac1n\sum_{|Bx_i|>R}\Bigl\|h \circ f(x_i)\Bigr\|_2.
\end{align*}
For $\mathcal{I}_1$, we apply Theorem \ref{thm:uniform_convergence}, and get with probability at least $1 - \delta_1$,
For $\mathcal{I}_1$, we apply Theorem \ref{thm:uniform_cv_B}, and get with probability at least $1 - \delta_1$,
$$\mathcal{I}_1 \le \frac1n\sum_{|Bx_i|<R} L_0 \cdot \sup_{x \in B_R}\Bigl\|f_n(x)-f(x)\Bigr\|_2 \le L_0 \cdot \frac{
C_1(\Sigma, A, B, V, R, \delta_1, d)}{\sqrt{n}}.$$
For $\mathcal{I}_2$, we prove the following Proposition, 

\begin{proposition} \label{prop:tail_control}
    Recall $x_1,\dots,x_n$ are i.i.d.\ $\mathbb{R}^d$-valued sub-Gaussian random vectors, with parameter $\sqrt{\|\Sigma\|_2}$. Let $B$ an invertible matrix satisfying the hypothesis of \cref{thm:uniform_cv_B}, let $\sigma_{B} := \sqrt{\|B\Sigma B^{\top}\|_2}= \|\Sigma^{1/2} B^{\top}\|_2$ and $R>0$. For all $n > 0$, and some constants $c, C > 0$, with probability at least $1 - \delta$, 
\[
\frac{1}{n}\sum_{\|Bx_i\|_2>R}\bigl\|f_n(x_i)\bigr\|_2
\ \le\
\Biggl(e^{-c(\frac{R}{\sigma_{B} } - C\sqrt{d} )^2} +\sqrt{\frac{1}{2n}\ln\!\Bigl(\frac{4}{\delta}\Bigr)}\Biggr)
\cdot
c\sigma_{B} \Biggl(\sqrt{d} + \sqrt{\ln \frac{2n}{\delta}}\Biggr).
\]
\end{proposition}
\begin{proof}
Let
\[
N_R:=\sum_{i=1}^n \mathbf{1}_{\{\|Bx_i\|_2>R\}},
\qquad
M_n:=\max_{1\le i\le n}\|x_i\|_2,
\qquad
p_R:=\mathbb{P}(\|BX\|_2>R).
\]
Since $f_n(x)$ is a convex combination of $\{x_j\}_{j=1}^n$, 
\[
\Bigl\|f_n(x)\Bigr\|_2 \le  M_n.
\]
Hence,
\begin{equation}\label{eq:key-reduction}
\frac{1}{n}\sum_{\|Bx_i\|_2>R}\bigl\|f_n(x_i)\bigr\|_2
\;\le\; \frac{N_R}{n}\,M_n.
\end{equation}
Our task therefore reduces to finding high-probability bounds for:
(i) the maximum norm $M_n$, and (ii) the fraction $N_R/n$ of samples with norm above $R$.
We begin with a lemma that controls the norm of a sub-Gaussian vector $X$.

\begin{lemma}[\cite{Vershynin18} see Exercise 6.3.5 p.114, 1st ed.]\label{lem:bound norm subG}
    There exist universal constants $C, c>0$ such that for all $R >0$, if $X$ is a sub-Gaussian random vector in $\mathbb{R}^d$ with sub-Gaussian norm at most $\sqrt{\|\Sigma\|_2}$, then $BX$ is sub-Gaussian with sub-Gaussian norm at most $\sigma_{B} := \sqrt{\|B\Sigma B^{\top}\|_2}= \|\Sigma^{1/2} B^{\top}\|_2$ and
\begin{equation*}
p_R := \mathbb{P}(\|BX\|_2  > R) \leq e^{-c(\frac{R}{\sigma_B} - C\sqrt{d})^2}.
\end{equation*}
with explicitly, $c= \frac14$ and $C = 4$.
\end{lemma}
Then we bound separately $\frac{N_R}{n}$ and $M_n$.

\paragraph{Step 1: Concentration of $\frac{N_R}{n}$}
The indicators $\mathbf{1}_{\{\|Bx_i\|_2>R\}}$ are i.i.d.\ Bernoulli($p_R$).
By Hoeffding's inequality, for every $t>0$,
\begin{equation}\label{eq:hoeffding}
\mathbb{P}\!\left(\left|\frac{N_R}{n}-p_R\right|\ge t\right)
\ \le\ 2\,\exp(-2nt^2).\nonumber
\end{equation}
Equivalently, for any $\delta_1\in(0,1)$, with probability at least $1-\delta_1$,
\begin{equation}\label{eq:NR-bound}
\left|\frac{N_R}{n}-p_R\right|
\ \le\ \sqrt{\frac{1}{2n}\ln\!\Bigl(\frac{2}{\delta_1}\Bigr)}.\nonumber
\end{equation}
Then, we move on to bounding the maximum of the $x_i$'s norm.

\paragraph{Step 2: Union bound for $M_n=\max_{1\le i\le n}\|x_i\|_2$.}
The union bound gives
\begin{equation}\label{eq:union-Mn}
\mathbb{P}(M_n\ge t)
\ \leq \mathbb{P}\!\left(\bigcup_{i=1}^n\{\|x_i\|_2\ge t\}\right)
\ \le\ \sum_{i=1}^n \mathbb{P}(\|x_i\|_2\ge t)
\ =\ n\,\mathbb{P}(\|X\|_2\ge t).\nonumber
\end{equation}
Applying Lemma \ref{lem:bound norm subG} we get,
\begin{equation}\label{eq:Mn-tail}
\mathbb{P}(M_n\ge t)\ \le\ n\,e^{-\Big(\frac{t}{c\, \sigma_B} - \sqrt{d} \Big)^2}.\nonumber
\end{equation}
For any $\delta_2\in(0,1)$, solving $n\,e^{-(\frac{t}{c\, \sigma_B} - \sqrt{d} )^2}=\delta_2$ yields, with probability at least $1 - \delta_2$, 
$$M_n\le c\sigma_B(\sqrt{d} + \sqrt{\ln \frac{n}{\delta_2}}).$$
Combining this result with the step 1, we get with probability at least $1-\delta_1-\delta_2$,
\[
\frac{1}{n}\sum_{\|Bx_i\|_2>R}\bigl\|f_n(x_i)\bigr\|_2
\ \le\
\Biggl(e^{-c(\frac{R}{\sigma_B} - C\sqrt{d} )^2} +\sqrt{\frac{1}{2n}\ln\!\Bigl(\frac{2}{\delta_1}\Bigr)}\Biggr)
\cdot
c\sigma\Biggl(\sqrt{d} + \sqrt{\ln \frac{n}{\delta_2}}\Biggr).
\]
Taking $\delta_1 = \delta_2 = \frac{\delta}{2}$ concludes the proof of Proposition \ref{prop:tail_control}. 
\end{proof}
We deduce from Proposition \ref{prop:tail_control}, and using the same bound for $N_R/n$ as in Proposition \ref{prop:tail_control}, that with probability at least $1-\delta_2$
\begin{align*}
   \mathcal{I}_2&\le \frac{1}{n}\sum_{\|Bx_i\|_2>R}  L_0 \cdot \bigl\|f_n(x_i)\bigr\|_2  \\
   &\le L_0 \cdot  \frac{N_R}{n}  M_n\\
   &\le
L_0 \cdot C_2(\Sigma, R,B,\delta_2, d, n)
\end{align*}
where
$$
C_2(\Sigma, R, B,\delta_2, d, n)=\Biggl(e^{-c(\frac{R}{\sigma_B} - C\sqrt{d} )^2} +\sqrt{\frac{1}{2n}\ln\!\Bigl(\frac{4}{\delta_2}\Bigr)}\Biggr)
\cdot  c\sigma_B\Biggl(\sqrt{d} + \sqrt{\ln \frac{2n}{\delta_2}} \Biggr).
$$

 For $\mathcal{I}_3$, applying Lemma \ref{lem:growth_gamma_sub}:
\begin{align*}
    \label{eq:clt_bound}
\frac{1}{n}\sum_{i=1}^{n}\mathbf{1}_{\{\|Bx_i\|_2>R\}}\; \bigl\|h \circ f(x_i)\bigr\|_2 &\le  L_0 \cdot \underbrace{2\, \|V\|_{2}\, \sqrt{\|\Sigma\|_2}\,\bigl\|\Sigma^{1/2}A\bigr\|_{2}}_{\alpha}\cdot \frac{1}{n}\sum_{i=1}^{n}\mathbf{1}_{\{\|Bx_i\|_2>R\}}\; \bigl\|x_i\bigr\|_2,\\
&\le  L_0 \cdot \alpha \cdot \frac{N_R}{n}  M_n
\end{align*}
where we used the notations of Proposition \ref{prop:tail_control}. The same bound on $\frac{N_R}{n}M_n$ holds and gives, with probability $1 - \delta_3$
\begin{align*}
\frac{1}{n}\sum_{i=1}^{n}\mathbf{1}_{\{\|Bx_i\|_2>R\}}\; \bigl\|h \circ f(x_i)\bigr\|_2 & \le L_0 \cdot \alpha \cdot C_2(\Sigma, R,B, \delta_3, d, n).
\end{align*}
We return to our initial inequality~\eqref{eq:first_bound}, where, with probability $1 - \delta$ (taking $\delta_1 = \delta_2 = \delta_3 = \frac{\delta}{3}$, and denoting $\delta' = \frac{\delta}{3}$),
\begin{equation*}
\Bigl\|\mathbb{E}_n[h \circ f_n(\hat X) - h \circ f(X)]\Bigr\|_2  \leq \mathcal{I}_1+\mathcal{I}_2+\mathcal{I}_3 \leq f(R).
\end{equation*}
Here, $f(R)=  L_0 \cdot\frac{C_1(\Sigma, A, B, R, \delta',d)}{\sqrt{n}} +  L_0 \cdot (1 + \alpha) \cdot C_2(\Sigma, A, B, R, \delta',n)$ with,
\begin{align*}
C_1(\Sigma, A, B,  V, R, \delta', d) &= P_1(\Sigma, A, B, V, R, \delta', d) \cdot \exp\left\{8\|\Sigma^{1/2} A B^{-1}\|^2 R^2\right\} \\
C_2(\Sigma, R, B, \delta', d, n) &= \Biggl(e^{-c(\frac{R}{\sigma_B} - C\sqrt{d} )^2} +\sqrt{\frac{1}{2n}\ln\!\Bigl(\frac{4}{\delta'}\Bigr)}\Biggr)
\cdot  c\sigma_B\Biggl(\sqrt{d} + \sqrt{\ln \frac{2n}{\delta'}} \Biggr).
\end{align*}
where 
$$
P_1(\Sigma, A, V, R, \delta', d)=\frac{C \, \sqrt{d}}{\delta} \|V\|_2  \cdot \|\Sigma\|_2^{1/2} \cdot(R \|A\|_2 \|\Sigma\|_2^{1/2}  d + \operatorname{tr}(\Sigma)^{1/2} \|A\|_2 + 5^{d/2}\sqrt{d})
$$
is linear in $R$.
To make the two terms of $f(R)$ of the same order, we choose 
\begin{equation*}
R^{\star} = \sqrt{\frac{\ln(n)}{16\, \|\Sigma^{1/2} A B^{-1}\|^2 + \frac{2c}{\sigma_B^2}}}.
\end{equation*}
To be valid, our upper bound on $\mathcal{I}_1$ derived above requires (since we applied Theorem \ref{thm:uniform_convergence}) that $n\geq 4e^{2R^{\star 2}\|\Sigma^{1/2}AB^{-1}\|_2^2}\Bigl(\frac1\delta -1\Bigr)$. This is the case for the choice $R=R^\star$ for $n$ large enough since
$$
4e^{2R^{\star 2}\|\Sigma^{1/2}AB^{-1}\|_2^2}\Bigl(\frac1\delta -1\Bigr)\leq 4\Bigl(\frac1\delta -1\Bigr)n^{1/8}\ll n.
$$
Noting $\tau_B := \|\Sigma^{\frac12} A B^{-1}\|^2$, we get the bound 
$$f(R^{\star}) = \Bigl(K_1 + K_2\cdot \sqrt{\ln(n)}\Bigr)\, n^{-\frac{1}{2}( 1 - \frac{16\tau_B^2}{16\tau_B^2 + \frac{2c}{\sigma_B^2}})}. $$
Recalling that $c = \frac14$ (see Lemma \ref{lem:bound norm subG}), we get
$f(R^{\star}) = \Theta (n^{-\frac{1}{2(1 + 32 \cdot \rho_B)}}\sqrt{\ln n})$ with $\rho_B
:=
\|\Sigma^{1/2}AB^{-1}\|_2^2
\|\Sigma^{1/2}B^\top\|_2^2 .$ 
The final step is to optimize this result over B. 
Notice that, for any invertible $B$, 
\begin{align*}
    \rho_B \ge \|\Sigma^{\frac12} A \Sigma^{\frac12}\|_2^2 =: H^2
\end{align*}
with equality attained for $B = \Sigma^{-1/2}$, which is indeed invertible.
The final rate becomes $$f(R^{\star}) = \Theta (n^{-\frac{1}{2(1 + 32 \cdot H^2)}}\sqrt{\ln n})$$ with $H = \|\Sigma^{\frac12} A \Sigma^{\frac12}\|_2$, which we call the horizon parameter. This concludes the proof of Theorem \ref{thm:convergence_rate_lipschitz}.
\end{proof}
\subsection{Proof of Proposition \ref{prop:hardmax}}
We consider a set of tokens $X_1,\dots,X_n \stackrel{\mathrm{iid}}{\sim} \mathcal{N}(0,\sigma^{2})$. Let $M_n^{+}:=\max_{j\le n}X_j$ and $M_n^{-}:=\min_{j\le n}X_j$ be the maximum and minimum of the token set $(X_i)_{i \in [1, n]}$.

The hardmax-attention is defined by,
$$f_n(X_i) := \begin{cases}
M_n^{+} & \text{if } X_i>0,\\[2pt] 
M_n^{-} & \text{if } X_i<0.
\end{cases}$$
This corresponds to a limit behavior where $A=+\infty$, and thus the attention map $f_n(X_i)$ selects only the token with same sign as $X_i$ and largest modulus. 

The empirical mean is 
$$\bar{f}_n:=\frac{1}{n}\sum_{i=1}^{n}f_n(X_i) = \frac{I_{+}}{n}M_n^{+}+\frac{I_{-}}{n}M_n^{-}$$
where $I_{+}=\#\{i : X_i>0\}, $ and $I_{-}=n-I_{+}$ denote the number of positive and negative tokens $X_i$. Note that $I_+\sim\mathrm{Bin}(n,\tfrac{1}{2})$.

Our main result in the hardmax regime derives the bias and standard deviation of the absolute value of the empirical mean, both of which are of order $1/\sqrt{\ln(n)}$.

\begin{remark}
Notice that the symmetry of $X$ yields $\mathbb E [\bar{f}_n]=0$.
\end{remark}

In order to prove Proposition \ref{prop:hardmax}, let's first decompose the empirical mean into two components:
$$\bar{f}_n = \frac{I_{+}}{n}\,M_n^{+}+\frac{I_{-}}{n}\,M_n^{-} = \tfrac{1}{2}\bigl(M_n^{+}+M_n^{-}\bigr)\;+ \Bigl(\tfrac{2I_{+}-n}{2n}\Bigr)\bigl(M_n^{+}-M_n^{-}\bigr).$$
Setting
\begin{equation}
    V_n:=\tfrac{1}{2}(M_n^{+}+M_n^{-}), \quad U_n:=\Bigl(\tfrac{2I_{+}-n}{2n}\Bigr)\bigl(M_n^{+}-M_n^{-}\bigr), \label{un_vn}
\end{equation}
we get $\bar{f}_n = V_n+U_n$.

The key insight is that $V_n$ captures the main asymptotic behavior while $U_n$ represents a smaller-order deviation term.

Before proving Proposition~\ref{prop:hardmax}, we establish several technical results on extreme values of Gaussian random variables.

\subsubsection{Extreme value analysis}

In this subsection, we derive precise asymptotic rates for $V_n$ and establish bounds on $U_n$. Extreme value theory provides weak convergence of the maximum and minimum (Lemmas~\ref{lem:joint_convergence} and~\ref{lem:sum_convergence} in the Appendix). To obtain convergence in expectation, we establish uniform integrability of both extremes in Lemma \ref{lem:uniform_integrability}.

Let $a_n$ and $b_n$ be the scaling parameters of extreme values defined by  $$a_n:=\sigma(\sqrt{2\ln n} -\frac{1}{2} \sqrt{2\ln n}  \, (\log \log n + \log 4\pi)
) \qquad \text{and} \qquad b_n = \frac{\sigma}{\sqrt{2\ln n}}.$$

\begin{lemma}[Uniform Integrability of Extremes]\label{lem:uniform_integrability}
Let $Z_n = \frac{M_n^{+} - a_n}{b_n}$ and $Y_n = -\frac{M_n^{-} + a_n}{b_n}$. Then $\{Z_n\}_{n\geq 1}$ and $\{Y_n\}_{n\geq 1}$ are uniformly integrable.
\end{lemma}

The definition of uniform integrability is recalled in Appendix \ref{app:hardmax}. 

\begin{proof}[Proof of Lemma \ref{lem:uniform_integrability}]
We establish the result for $Z_n$ in several steps. The same argument applies to $Y_n$ by symmetry.

\textit{Step 1:} From equation (1.5) in \cite{Tanguy15}, we have the concentration inequality:
\begin{equation}\label{e:tanguy}
\mathbb{P} (|M_n^+- \mathbb{E}[M_n^+]| > t) \leq 6 e^{-c t \sqrt{\ln n}}
\end{equation}

\textit{Step 2:} We show that $\sup_n \mathbb{E}[|Z_n|] < \infty$.

Let $M_n'$ be an independent copy of $M_n^+$ and set $Z_n' = \frac{M_n' - a_n}{b_n}$. From Step 1:
\begin{equation}
    \sup_n \frac{1}{b_n}\mathbb{E}[|M_n^+ - \mathbb{E}[M_n^+]|] < \infty\,.\label{eq:sup_mn}
\end{equation}
Using the convexity of $|\cdot|$ and Jensen's inequality,
$$\mathbb{E}_{Z_n'}[|Z_n - Z_n'|] \geq |Z_n - \mathbb{E}[Z_n']| \,.$$
Therefore,
$$\mathbb{E}[|Z_n|] \leq \mathbb{E}[|Z_n - Z_n'|] + |\mathbb{E}[Z_n']|$$

From \eqref{eq:sup_mn}, we can deduce that $\sup_{n}\mathbb{E}[|Z_n - Z_n'|] < \infty$.
Indeed 
$$
|Z_n - Z_n'| = \frac{1}{b_n}|M_n - M_n'| \le \frac{1}{b_n}|M_n - \mathbb E[M_n]| + \frac{1}{b_n}|M_n '- \mathbb E[M_n']|.
$$
Since, on the other hand $\sup_n |\mathbb{E}[Z_n]| < \infty$, this gives $\sup_n \mathbb{E}[|Z_n|] < \infty$.

\textit{Step 3:} We transfer the concentration around $\mathbb{E}[M_n^+]$ to a concentration around $a_n$. 

Define $\delta_n = \mathbb{E}[M_n^+] - a_n$. Then $|\delta_n| = O(b_n)$, so $|\delta_n|\to 0$. From the event inclusion 
$$\{|M_n^+- a_n| > t\} \subset \{|M_n^+ - \mathbb{E}[M_n^+]| > t - |\delta_n|\}$$
and \eqref{e:tanguy}, we deduce
$$\mathbb{P}(|M_n^+ - a_n| > t) \leq 6 e^{-c (t -|\delta_n|) \sqrt{\ln n}}\,.$$
For $n$ large enough, it holds for all $t \ge 1$ that $t - |\delta_n| \geq t/2$, giving
$$\mathbb{P}(|Z_n| > t) \leq 6 e^{-c' t}\,.$$

\textit{Step 4:} Using the tail integral formula,
$$\mathbb{E}\!\big[|Z_n|\,\mathbf{1}_{\{|Z_n| > A\}}\big] = \int_{A}^{+\infty} \mathbb{P}(|Z_n| > x)\,dx \leq \int_{A}^{+\infty} 6 e^{-c' x}dx = \frac{6}{c'} e^{-c' A} \to 0$$
as $A \to \infty$, uniformly in $n$. 
\end{proof}

\begin{lemma}[Asymptotic behavior of  $V_n$]\label{lem:l1_convergence} There holds
$$\mathbb{E}|V_n| = \frac{\ln(4)\,\sigma}{2\,\sqrt{2\ln n}}\,[1+o(1)] \quad \text{and} \quad \mathbb{E}[V_n^2] = \frac{\sigma^{2}\pi^{2}}{24\,\ln n}\,[1+o(1)]\,.$$
\end{lemma}
\begin{proof}
The random variables $Z_n = \frac{M_n^{+} - a_n}{b_n}$ and $Y_n = -\frac{M_n^{-} + a_n}{b_n}$ jointly converge weakly to the joint distribution of two independent Gumbel random variables, which we denote $G_{+}$ and $G_{-}$ (see Lemma~\ref{lem:joint_convergence} in the Appendix).

By Lemma~\ref{lem:uniform_integrability}, both $|Z_n|$ and $|Y_n|$ are asymptotically uniformly integrable. Since $|Z_n - Y_n| \leq |Z_n| + |Y_n|$, it follows that $|Z_n - Y_n|$ is also asymptotically uniformly integrable. Combining the weak convergence of the sum (Lemma~\ref{lem:sum_convergence}) with the dominated convergence theorem (\cite{VanDerVaart98}, Theorem 2.20, Section 2.5), we conclude that $|Z_n - Y_n|$ converges in expectation.

Recall that $V_n = \frac{M_n^+ + M_n^-}{2}$, which can be rewritten as $V_n = \frac{\sigma(Z_n - Y_n)}{2\sqrt{2\ln n}}$. We can therefore deduce that
\begin{equation}
\mathbb{E}\left[\frac{2\sqrt{2\ln n}|V_n|}{\sigma}\right] \to \mathbb{E}\left[|G_+ - G_-|\right].
\end{equation}

Since $L := G_+ - G_-$ follows a centered logistic law, we have
$$\mathbb{E}[|L|] = \ln 4,$$
which gives:
$$\lim_{n\to\infty}\frac{2\sqrt{2\ln n}}{\sigma}\,\mathbb{E}|V_n|=\ln 4\,.$$

The variance result follows similarly using $\text{Var}(L) = \pi^2/3$.
\end{proof}

\begin{lemma}[Upper bound on $U_n$]\label{lem:deviation_bound}
Let $U_n$ and $V_n$ defined in \eqref{un_vn}, it holds:
$$\mathbb{E}[|U_n|] \leq \sigma\sqrt{\frac{\ln n}{n}}, \quad \text{and} \quad \mathbb{E}[U_n^2]=  O\Bigl(\frac{\ln n}{n}\Bigr)\,.$$
\end{lemma}
\begin{proof}
Using $\text{Var}(2I_{+}-n)=n$ and $\mathbb E[|M_n^{+}-M_n^{-}|^2] =O(a_n^2)$, we have:
$$\mathbb{E}[|U_n|] \leq \Bigl(\mathbb{E}\left(\left|\frac{2I_{+}-n}{2n}\right|^2\right) \cdot \mathbb{E}[|M_n^{+}-M_n^{-}|^2]\Bigr)^{1/2} \leq \sqrt{\frac{1}{4n}} \cdot 2a_n = \sigma\sqrt{\frac{\ln n}{n}}\,.$$
For the second moment, using that $\mathbb E[|M_n^{+}-M_n^{-}|^4] = O(a_n^4)$, we get
$$\mathbb{E}[U_n^2] \leq \Bigl(\mathbb{E}\left[\left|\frac{2I_{+}-n}{2n}\right|^4\right] \cdot \mathbb{E}[|M_n^{+}-M_n^{-}|^4]\Bigr)^{1/2} = O(\frac{\ln n}{n})
$$
which concludes the proof.
\end{proof}

\subsubsection{Conclusion}
We first establish the convergence of the mean. Using the triangle inequality,
$$|V_n|-|U_n|\leq|\bar{f}_n|\leq|V_n|+|U_n|\,.$$
Taking expectations and using $\mathbb{E}|U_n|=o(\mathbb{E}|V_n|)$ from Lemmas \ref{lem:l1_convergence} and \ref{lem:deviation_bound}:
$$(1-o(1))\,\mathbb{E}|V_n|\;\leq\;\mathbb{E}|\bar{f}_n|\;\leq\;(1+o(1))\,\mathbb{E}|V_n|$$
which provides the announced equivalent for the mean in Lemma \ref{lem:l1_convergence}.

For the variance, we decompose:
$$\text{Var}(\bar{f}_n)=\text{Var}(V_n)+ \text{Var}(U_n) + 2 \, \text{Cov}(U_n, V_n)\,.$$
By Cauchy--Schwarz,
\[
|\operatorname{Cov}(U_n,V_n)|
\;\leqslant\;
\sqrt{\operatorname{Var}(U_n)\,\operatorname{Var}(V_n)}.
\]
In addition, from Lemma \ref{lem:deviation_bound}, $\mathbb{E}[U_n^2] = O(\ln n / n)$, and from Lemma \ref{lem:l1_convergence}, $\text{Var}(V_n) = \Theta(1/\ln n)$. This gives $\text{Var}(U_n) = o(\text{Var}(V_n))$ and $|\operatorname{Cov}(U_n,V_n)| = |\operatorname{Cov}(U_n,V_n)|
\leq
\sqrt{O\!\left(\frac{\ln n}{n}\right)\,\Theta\!\left(\frac{1}{\ln n}\right)}
= O\!\left(\frac{1}{\sqrt{n}}\right)
= o\!\left(\frac{1}{\ln n}\right)$ from which $\text{Var}(\bar{f}_n) \sim \text{Var}(V_n)$ follows. Since $\mathbb{E}(\bar{f}_n)=\mathbb{E}(V_n)=0$, this concludes the proof.

\subsection{Lemmas for hardmax convergence rate}\label{app:hardmax}
In this section, we introduce some useful lemmas about extreme value theory used in the proof of Proposition \ref{prop:hardmax}.

\begin{definition}[Asymptotic uniform integrability]\label{def:uniform_integrability}
A family $(X_i)_{i\in I}$ of random variables is \emph{asymptotically uniformly integrable} if:
$$\lim_{A \rightarrow \infty}\sup_{i\in I}\;\mathbb{E}\!\left[\,|X_i|\,\mathbf{1}_{\{|X_i| > A\}}\right] = 0.$$
\end{definition}

\begin{lemma}[Joint Convergence]\label{lem:joint_convergence}
Let $Z_n = \tfrac{M_n^{+}-a_n}{b_n}$ and $Y_n = \tfrac{-M_n^{-}-a_n}{b_n}$.
Joint convergence holds:
$$\Bigl(Z_n,\;Y_n\Bigr)\xrightarrow{\mathcal{L}}(G_{+},G_{-})$$
with $G_{+},G_{-}$ independent, centered Gumbel random variables (variance $\pi^{2}/6$). 
\end{lemma}
\begin{proof}[Proof of Lemma \ref{lem:joint_convergence}]
A statement of the weak convergence of the marginals can be found in Theorem 1 of \cite{Tanguy15}.
For joint convergence, see Theorem 1.8.3 of \cite{Nadarajah00} applied to Gaussian distribution.
\end{proof}

\begin{lemma}[Sum Convergence]\label{lem:sum_convergence}
$$\frac{2\sqrt{2\ln n}}{\sigma}\,V_n = \frac{M_n^{+}+M_n^{-}}{\sigma}\,\sqrt{2\ln n}\;\xrightarrow{d}\;L$$
where $L:= G_{+}-G_{-}$ follows a standard logistic law with variance $\pi^{2}/3$.
\end{lemma}
\begin{proof}
This follows from Lemma \ref{lem:joint_convergence} and continuity of the mapping $(a,b) \mapsto a + b$.
\end{proof}
\subsection{Counter-examples for heavy-tailed distributions} \label{app:heavy_tail}
This subsection shows that the sub-Gaussian assumption is mandatory in order for the hardmax attention to converge. To prove so, we exhibit a counter-example of a heavy-tailed distribution. 

The key idea follows from the same decomposition as in the proof of \cref{prop:hardmax}: the empirical attention mean splits into a fluctuation term $\frac{M_n^+ - M_n^-}{\sqrt{n}}$
and a mean-shift term $\frac{M_n^+ + M_n^-}{2}.$
In the Gaussian case, symmetry forces the latter to zero. For heavy-tailed distributions, this term fails to converge
Consider a distribution with density
\[
\mu(x) = c_k(1+|x|)^{-k},
\qquad k \geq 4.
\]
Taking the same notation as in Appendix~B.6, for $x>0$,
\[
\mathbb{P}(M_n^+ > x)
=
1 -
\left(
1 - c_k' x^{-k+1}
\right)^n.
\]
Denoting by $p_n(\alpha,\beta)$ the probability that
\[
M_n^+
\in
\left[
(\alpha n)^{1/(k-1)},
(\beta n)^{1/(k-1)}
\right],
\]
we have
\[
p_n(\alpha,\beta)
=
\left(1-\frac{c_k'\alpha}{n}\right)^n
-
\left(1-\frac{c_k'\beta}{n}\right)^n
\underset{n\to+\infty}{\longrightarrow}
e^{-c_k'\alpha} - e^{-c_k'\beta}.
\]
In other words, $n^{-1/(k-1)}M_n^+$ converges in law to a non-trivial explicit distribution. Since $n^{-1/(k-1)}M_n^-$ converges to the same law,
\[
\frac{M_n^+ - M_n^-}{\sqrt{n}}
\to 0
\qquad
\text{a.s.}
\]
However,
\[
\frac{M_n^+ + M_n^-}{2}
\]
does not converge to zero: it has strictly positive variance, and the probability that it exceeds $1$ is positive. Therefore, studying the rate of convergence makes no sense in this setting: $\mathbb{E}\,\Gamma_n$ does not converge. This shows that the sub-Gaussian assumption is not merely a technical convenience; it is necessary for the convergence of the attention output to hold.

\section{Additional results and proofs}\label{app:add_results}

\subsection{At most linear growth of the attention map}\label{app:linear_growth}
We start with a preliminary result that quantifies the growth of the attention map norm with the query norm. Lemma \ref{lem:growth_gamma_sub} establishes that the attention map $f(x)$
cannot grow faster than linearly in the query norm $\|x\|_2$. This linear growth property is crucial for controlling the behavior of attention mechanisms and will be instrumental in establishing convergence rates.

Let us denote by $\|\Sigma\|_2 := \lambda_{\max}\!\bigl(\Sigma\bigr)$ the largest eigenvalue of $\Sigma$, where $\| \cdot \|_2$ denotes both the matrix norm and the Euclidean norm (clear from the context).
\begin{lemma}[At-most linear growth of $\Gamma$]\label{lem:growth_gamma_sub}
Let $X\in\mathbb{R}^d$ be centered and sub-Gaussian with parameter matrix $\Sigma \succ0$. For any matrices $A\in\mathbb{R}^{d\times m}$ and $V\in\mathbb{R}^{p\times d}$ and any $x,t\in\mathbb{R}^m$, define 
\[
f(x):=V\,\nabla K(Ax),\qquad
K(t):=\ln\mathbb{E}[e^{\langle t,X\rangle}].
\]
Then
\[
\bigl\|f(x)\bigr\|_{2}
\;\le\;
2\,\|V\|_{2}\,
\|\Sigma^{1/2}\|_2\,
\bigl\|\Sigma^{1/2}A\bigr\|_{2}\,
\|x\|_{2}\; .
\]
\end{lemma}

\begin{proof} Following the proof in \cite{Bobkov25}, we define the auxiliary function 
$P(t) = \frac 12 \langle t ,\Sigma t \rangle - K(t)$ which satisfies $P(t) \geq 0$ by definition of the sub-Gaussiannity. By Jensen's inequality, we have $K(t) \geq 0$ which implies $P(t) \leq \frac 12 \langle t ,\Sigma t \rangle$. Moreover, by convexity of $K(t)$, we have $P'' \leq \Sigma$. By Taylor expansion, this implies 
\begin{equation}
    0\leq P(h+t) = P(t) + \langle  \nabla P(t),h \rangle + \frac{1}{2} \langle h,\Sigma h\rangle\,.\nonumber
\end{equation}
Minimizing over $h$ the r.h.s. leads to 
\begin{equation}
    \frac{1}{2} \langle \nabla P(t),\Sigma^{-1} \nabla P(t)\rangle \leq P(t)\,.\nonumber
\end{equation}
Therefore, we get by the triangle inequality
\begin{equation}
    \| \nabla K(t) \|_{\Sigma^{-1}} = \|\Sigma t -\nabla P \|_{\Sigma^{-1}} \leq \|t\| + \sqrt{2 P(t)} \leq 2\|t\|_{\Sigma} \,,\nonumber
\end{equation}

where we used the Mahalanobis norm $\|\cdot\|_S$ defined for a positive definite matrix S by $\|x\|_S = x^\top S x = \|S^{1/2}x\|_2^2$.

Converting this estimate for the standard Euclidean norm leads to 
\begin{align*}
\|\nabla K(Ax)\|_2 = \left\|\Sigma^{1/2}\Sigma^{-1/2} \nabla K(Ax)\right\|_2\nonumber 
&\leq \left\|\Sigma^{1/2}\right\|_{\rm op} \left\|\Sigma^{-1/2} \nabla K(Ax)\right\|_2 \\ 
%&\leq\sqrt{\|\Sigma\|_2}\; \bigl\|\Sigma^{-1/2}\nabla K(Ax)\bigr\|_2 \\
&\leq 2\,\|\Sigma\|_2^{1/2}\,\|\Sigma^{1/2}Ax\|_2 .\label{eq:euclid}
\end{align*}
Coming back to the attention map, we have $f(x)=V\,\nabla K(Ax)$, which yields the claimed linear growth.
\end{proof}
\begin{remark}
The continuous self-attention map $f$ admits a probabilistic interpretation as an
exponentially tilted expectation, or equivalently as an importance-sampling transform.
More precisely, it can be rewritten as
\[
f(X)
\;=\;
V\,\frac{\mathbb{E}\!\left[\,X\,e^{\langle Ax, X\rangle}\right]}{\mathbb{E}\!\left[e^{\langle Ax, X\rangle}\right]}
\;=\;
V\,\mathbb{E}_{\mathbb{Q}_{Ax}}[X],
\qquad
d \mathbb{Q}_{Ax} =\frac{e^{\langle Ax,X\rangle}}{\mathbb{E}[e^{\langle Ax,X\rangle}]} \, d \mathbb P \,.
\]
where we use the \emph{Esscher transform} for $t\in\mathbb{R}^d$, defined by the tilted measure
\[
d\mathbb{Q}_t:=\frac{e^{\langle t,X\rangle}}{\mathbb{E}[e^{\langle t,X\rangle}]}\,d\mathbb{P}.
\]
Thus, the finite-$n$ self-attention $f_n$ is the Monte Carlo importance-sampling estimator of this tilted expectation.
\end{remark}

Gaussian distributions saturate the sub-Gaussian bound of Lemma \ref{lem:growth_gamma_sub} (see Lemma \ref{lem:Gaussian_linearity}): $K(t)=\tfrac{1}{2}t^\top \Sigma t$ exactly, so $\nabla K$ is linear and Lemma~\ref{lem:growth_gamma_sub} is tight up to constants. As a result, the class of Gaussian distributions is preserved layer-wise \cite{Castin25}, which allows to follow the evolution of tokens by tracking only means and covariances. This makes Gaussians a calibrating case for both growth bounds and deviation inequalities (see for example \cite{Wainwright19}, and \cite{Vershynin18}).  

Note that, when $\nu$ has a compact support, the transformed distribution of the tokens stays bounded.

\subsection{Pointwise convergence of the attention map}\label{app:ptw_cv}

Here we prove that the pointwise concentration of the attention map exhibits the classical $O(\frac{1}{\sqrt{n}})$ asymptotic rate for empirical processes, combined with an exponential dependence on $\|\Sigma^{1/2}Ax\|_2^2$ that captures how attention mechanisms amplify signals along high-variance directions. This pointwise analysis provides relevant insights into how individual attention outputs stabilize with increasing context length, with potential applications in attention sketching and approximation methods. 

\begin{proposition}[Pointwise convergence of the attention map for sub-Gaussian tokens]\label{prop:ptwise_convergence}
Let $X \sim \nu$ be centered and sub-Gaussian with parameter matrix $\Sigma \succ0$. For any fixed $x\in\mathbb{R}^m$, the empirical attention computed from $n$ i.i.d.\ tokens satisfies, for $n$ large enough, with probability at least $1 - 2\delta$,
\begin{equation}\label{eq:ptwise_convergence}
    \bigl\| f_n(x) -f(x) \bigr\|_2
\;\le\;
 q_{\Sigma, V, d, \delta}\cdot\frac{e^{\frac52 \,\|\Sigma^{1/2} A x\|_2^2 }}{\sqrt n}\,,
\end{equation}
where $q_{\Sigma, V, d, \delta} := (2 + \sqrt{3})\|V\|_2 \, \|\Sigma\|_2^{1/2}\cdot\;q_{d,1- \delta}^{1/2} .$

Here $q_{d,1-\delta}$ denotes the $(1-\delta)$-quantile of the $\chi^2(d)$ distribution, which satisfies in particular $q_{d,1-\delta} \le d + 2\sqrt{d\ln(1/\delta))} + 2\ln(1/\delta).$
\end{proposition}

The proof of this result combines classical tools of statistics with properties of sub-Gaussian distributions. 
\begin{proof}
Let $\nu$ be as in the statement of Proposition \ref{prop:ptwise_convergence}, and let $(x_1, \ldots, x_n)$ be i.i.d. samples from $\nu$. We let $A = K^\top Q$ where $K$ and $Q$ are the key and query matrices in the attention mechanism, and consider the random vector $Z_j(x) := \left( e^{\langle Ax, x_j \rangle} x_j, e^{\langle Ax, x_j \rangle} \right) \in \mathbb{R}^{d+1}$.

We introduce notation for the numerator and denominator of the attention maps $f_n(x)$ and $f(x)$:
\begin{align*}
f_n(x) &= \frac{N_n(x)}{D_n(x)} \quad\text{where} \quad
N_n(x) = \frac{1}{n} \sum_{j=1}^n e^{\langle Ax, x_j \rangle} x_j \in \mathbb{R}^d, \quad
D_n(x) = \frac{1}{n} \sum_{j=1}^n e^{\langle Ax, x_j \rangle} \in \mathbb{R} \\
f(x) &= \frac{N(x)}{D(x)} \quad\text{where} \quad
N(x) = \mathbb{E}_{X\sim\nu}\left[ e^{\langle Ax, X \rangle} X \right] \in \mathbb{R}^d, \quad
D(x) = \mathbb{E}_{X\sim\nu}\left[e^{\langle Ax, X \rangle} \right] \in \mathbb{R}
\end{align*}
where $X \sim \nu$.

\textbf{ Step 1: Joint Central Limit Theorem (CLT) and Delta Method.}
Under the assumptions that $\mathbb{E} \left[ \|e^{\langle Ax, X \rangle} X \|_2 \right] < \infty$ and $\mathbb{E}[e^{\langle Ax, X \rangle}] < \infty$, the strong law of large numbers ensures almost sure convergence
\begin{equation*}
N_n(x) \xrightarrow{\text{a.s.}} N(x), \qquad D_n(x) \xrightarrow{\text{a.s.}} D(x) \quad \text{as } n \to \infty.
\end{equation*}
The random vectors $Z_j(x) = (e^{\langle Ax, x_j \rangle} x_j,e^{\langle Ax, x_j \rangle}) \in \mathbb{R}^{d+1}$ are i.i.d. with mean $\mathbb{E}[Z_1(x)] = (N(x), D(x)) \in \mathbb{R}^{d+1}$. Assuming $\mathbb{E}[\|Z_1(x)\|^2] < \infty$, the multivariate central limit theorem yields
\begin{equation*}
\sqrt{n}\left( \begin{pmatrix} N_n(x) \\ D_n(x) \end{pmatrix} - \begin{pmatrix} N(x) \\ D(x) \end{pmatrix} \right)
\xrightarrow{d} \mathcal{N}_{d+1}\left( 0, \operatorname{Cov}(Z_1(x)) \right)
\end{equation*}
where $\operatorname{Cov}(Z_1(x)) = \begin{pmatrix}
\Sigma_{11}(x) & \Sigma_{12}(x) \\
\Sigma_{21}(x) & \Sigma_{22}(x)
\end{pmatrix} \in \mathbb{R}^{(d+1) \times (d+1)},$ and
\begin{align*}
    \Sigma_{11}(x) &= \operatorname{Cov}(e^{\langle Ax, X \rangle} X) = \mathbb E[X^\top X e^{2\langle Ax,X\rangle}]-\bigl(\mathbb E[Xe^{\langle Ax,X\rangle}]\bigr)^{2} \in \mathbb{R}^{d \times d}\\
    \Sigma_{22}(x) &= \operatorname{Var}(e^{\langle Ax, X \rangle}) = \mathbb E[e^{2 \langle Ax,X\rangle}]-\bigl(\mathbb E[e^{\langle Ax,X\rangle}]\bigr)^{2}\in \mathbb{R}\\
    \Sigma_{12}(x) &= \Sigma_{21}(x)^\top = \operatorname{Cov}(e^{\langle Ax, X \rangle} X, e^{\langle Ax, X \rangle}) = \mathbb E[Xe^{2tX}]-\mathbb E[Xe^{tX}]\,\mathbb E[e^{tX}] \in \mathbb{R}^{d}.
\end{align*}
Applying the multivariate delta method to $g: \mathbb{R}^{d} \times \mathbb{R}_*^+ \rightarrow \mathbb{R}^d$ defined by $g(a, b) = a/b$, we obtain,
\begin{equation*}
\sqrt{n} \left( \frac{N_n(x)}{D_n(x)} - \frac{N(x)}{D(x)} \right)
\xrightarrow{d} \mathcal{N}_d \left( 0, \operatorname{Cov}(x) \right),
\end{equation*}
where the asymptotic covariance is denoted $\operatorname{Cov}(x)$ and is equal to 
$$\operatorname{Cov}(x) = \nabla g_{(N(x), D(x))} \operatorname{Cov}(Z_1(x)) \nabla g_{(N(x), D(x))}^\top \in \mathbb{R}^{d \times d}$$
The gradient of $g$ at $(N(x), D(x))$ is
\begin{equation*}
\nabla g_{(N(x), D(x))} =
\begin{pmatrix}
\frac{1}{D(x)} I_d & -\frac{N(x)}{D(x)^2}
\end{pmatrix} \in \mathbb{R}^{d \times (d+1)},
\end{equation*}
where $I_d$ is the $d \times d$ identity matrix.
The variance $\operatorname{Cov}(x)$ expands as
\begin{equation*}
\operatorname{Cov}(x) = \frac{1}{D(x)^2} \Sigma_{11}(x)
- \frac{1}{D(x)^3} \left( N(x) \Sigma_{12}(x)^\top + \Sigma_{12}(x) N(x)^\top \right)
+ \frac{1}{D(x)^4} \Sigma_{22}(x) N(x) N(x)^\top.
\end{equation*}
\textbf{Step 2: Chi-Square Bound.}
From the asymptotic normality, the squared Euclidean norm converges in distribution to a chi-squared random variable
\begin{equation*}
\left\|\sqrt{n} \, \operatorname{Cov}(x)^{-1/2}(f_n(x) - f(x))\right\|_2^2 \xrightarrow{d} \chi^2(d).
\end{equation*}
For any $\delta \in (0,1)$, for $n$ large enough
% \textcolor{red}{il faut que cette hypothèse apparaisse dans l'énoncé. Et je pense que la borne \eqref{eq:chi_bound} ne découle pas de la convergence ci-dessous: il faut prendre un peu d'espace, $1-2\delta$ au lieu de $1-\delta$.}\lea{je n'ai pas eu le temps de reprendre davantage désolée...}
, we have
\begin{equation} \label{eq:chi_bound}
\mathbb{P}\left(\left\|\sqrt{n} \, \operatorname{Cov}(x)^{-1/2}(f_n(x) - f(x))\right\|_2^2 \leq q_{d, 1-\delta}\right) \geq 1 - \delta
\end{equation}
where $q_{d, 1-\delta} := F_{\chi^2(d)}^{-1}(1-\delta)$ is the $(1-\delta)$-quantile of the $\chi^2(d)$ distribution.

To derive the desired inequality for $\|f_n(x) - f(x)\|_2^2$, we use a spectral inequality.

\textbf{Step 3: Spectral Bound.}
Let $\tilde{z}_n = \sqrt{n}(f_n(x) - f(x))$. Applying the Rayleigh-Ritz theorem, we get
\begin{equation*}
\|\tilde{z}_n\|_2^2 \leq \lambda_{\max}(\operatorname{Cov}(x)) \|\operatorname{Cov}(x)^{-1/2} \tilde{z}_n\|_2^2.
\end{equation*}
This yields the event inclusion
\begin{equation} \label{eq:inclusion}
\left\{\|\operatorname{Cov}(x)^{-1/2} \tilde{z}_n\|_2^2 \leq q_{d,1-\delta}\right\} \subset \left\{\|\tilde{z}_n\|_2^2 \leq \lambda_{\max}(\operatorname{Cov}(x)) q_{d,1-\delta}\right\}.
\end{equation}
From \eqref{eq:chi_bound} and \eqref{eq:inclusion}, we obtain that for $n$ large enough,
\begin{equation*}
\mathbb{P}\left(\|\tilde{z}_n\|_2^2 \leq \lambda_{\max}(\operatorname{Cov}(x)) q_{d,1-\delta}\right) \geq 1 - \delta.
\end{equation*}
Therefore:
\begin{equation*}
\|f_n(x) - f(x)\|_2 \leq \frac{\lambda_{\max}(\operatorname{Cov}(x))^{1/2} q_{d,1-\delta}^{1/2}}{\sqrt{n}}
\quad \text{with probability } 1-\delta.
\end{equation*}
This spectral inequality allows us to convert bounds on quadratic forms back to Euclidean norm bounds, completing the bridge between the chi-squared concentration and our desired result.
% \textit{Improved Bound via Laurent-Massart Inequality}\\
The quantile $q_{d,1-\delta}$ can be upper bounded using the Laurent-Massart concentration inequality for chi-squared random variables (see \citep[Section~4.1, Lemma~1]{Laurent00}) stating that, for any $x > 0$,
\begin{equation*}
\mathbb{P}(\chi^2(d) \geq d + 2\sqrt{dx} + 2x) \leq e^{-x}.
\end{equation*}
Setting $x = \ln(1/\delta)$ in this inequality yields
\begin{equation*}
\mathbb{P}\left(\chi^2(d) \geq d + 2\sqrt{d\ln(1/\delta)} + 2\ln(1/\delta)\right) \leq \delta.
\end{equation*}
By definition of the quantile function,
\begin{equation*}
q_{d,1-\delta} \leq d + 2\sqrt{d\ln(1/\delta)} + 2\ln(1/\delta).
\end{equation*}
This yields our refined bound, with probability at least $1 - \delta$,
\begin{equation*}
\|f_n(x) - f(x)\|_2 \leq \frac{\lambda_{\max}(\operatorname{Cov}(x))^{1/2} \bigl(d + 2\sqrt{d\ln(1/\delta)} + 2\ln(1/\delta)\bigr)^{1/2}}{\sqrt{n}}.
\end{equation*}
\textbf{Step 4: Moment Generating Function Representation.}
We express the variance components using the moment generating function (MGF) $M(t) = \mathbb{E}[e^{t^\top X}]$ and its derivatives. For a centered sub-Gaussian $X$ with parameter $\Sigma $:
\begin{align*}
M(t) = \mathbb E[e^{\langle t, X\rangle}], \quad M'(t) = \mathbb E[Xe^{\langle t, X\rangle}], \quad M''(t) &= \mathbb E[XX^\top e^{\langle t, X\rangle}].
\end{align*}
The variance components become
\begin{align*}
\Sigma_{11}(x) &= M''(2Ax) - M'(Ax) M'(Ax)^\top .\\
\Sigma_{22}(x) &= M(2Ax) - M(Ax)^2. \\
\Sigma_{12}(x) &= M'(2Ax) - M'(Ax) M(Ax).
\end{align*}
Since $D(x) = M(Ax)$ and $N(x) = M'(Ax)$, the covariance $\operatorname{Cov}(x)$ can be written as
\begin{multline*}
\operatorname{Cov}(x)=
\frac{M''(2Ax)}{M(Ax)^{2}}
-\frac{ M'(Ax)M'(2Ax)^{\top}+M'(2Ax)M'(Ax)^{\top}}{M(Ax)^{3}}
+\frac{M(2Ax)}{M(Ax)^{4}}\,M'(Ax)M'(Ax)^{\top}.
\end{multline*}
Let $X$ be $\Sigma $-sub-Gaussian with $\|\Sigma\|_2^{1/2}  = \sqrt{\lambda_{\max}(\Sigma )}$. For any $t \in \mathbb{R}^d$ and unit vector $u$:
\begin{align}
|\langle u, M'(t) \rangle| &\leq \|\Sigma\|_2^{1/2}  e^{t^\top \Sigma t} .\label{eq:bound1}\\
\langle u, M''(t) u \rangle &\leq \sqrt{3} \|\Sigma\|_2 e^{t^\top \Sigma t}.
\label{eq:bound2}
\end{align}
Let us prove this result. Let $Y = \langle u, X \rangle$. By the Cauchy-Schwarz inequality,
\begin{equation*}
|\langle u, M'(t) \rangle| = |\mathbb{E}[Y e^{\langle t, X \rangle}]| \leq (\mathbb{E}[Y^2])^{1/2} (\mathbb{E}[e^{2\langle t, X \rangle}])^{1/2}.
\end{equation*}
Since $X$ is $\Sigma $-sub-Gaussian, $Y$ is $\sigma_u$-sub-Gaussian with $\sigma_u = \sqrt{u^\top \Sigma u} \leq \|\Sigma\|_2^{1/2}$. Thus $\mathbb{E}[Y^2] \leq \|\Sigma\|_2$ and $\mathbb{E}[e^{2\langle t, X \rangle}] \leq e^{2t^\top \Sigma t}$.

The second inequality follows similarly using the fourth moment bound $\mathbb{E}[Y^4] \leq 3 \|\Sigma\|_2^{2}$ for sub-Gaussian random variables.

\textbf{Step 5: Main spectral bound.}
We will prove the following spectral bound for attention variance.
For any $x \in \mathbb{R}^d$,
\begin{equation*}
\lambda_{\max}(\operatorname{Cov}(x)) \leq (2 + \sqrt{3})\|\Sigma\|_2e^{5(Ax)^\top \Sigma  Ax}.
\end{equation*}
Let $u$ be a unit vector and $t = Ax$. Using $M(t) \geq 1$, and $\frac{M(2t)}{M(t)^{4}}\le\frac1{M(t)^{2}}\le1$, 
\begin{equation*}
|u^\top \operatorname{Cov}(x) u| \leq |u^\top M''(2t) u| + |u^\top M'(t)|^2 + 2|u^\top M'(2t)| |u^\top M'(t)|.
\end{equation*}
Substituting the bounds from~\eqref{eq:bound1} and~\eqref{eq:bound2}, with $p(t) := t^\top \Sigma t$ ,
\begin{equation*}
|u^\top \operatorname{Cov}(x) u| \leq \|\Sigma\|_2 \left(\sqrt{3} e^{4p(t)} +  e^{2p(t)} + 2 e^{5p(t)}\right).
\end{equation*}
Since the exponential terms are dominated by $e^{5p(t)}$, we obtain
\begin{equation*}
|u^\top \operatorname{Cov}(x) u| \leq (2 + \sqrt{3}) \|\Sigma\|_2 e^{5p(t)}.
\end{equation*}
Since this holds for all unit vectors $u$, the result follows.

\textbf{Step 6: Conclusion.}
Combining our results, we obtain that, for any $x \in \mathbb{R}^d$ and $\delta \in (0,1)$, with probability at least $1-\delta$,
\begin{equation*}
\|f_n(x) - f(x)\|_2 \leq \frac{C(\Sigma , A, V, x, \delta, d)}{\sqrt{n}},
\end{equation*}
where
\begin{equation*}
C(\Sigma , A, V, x, \delta, d) =(2+\sqrt{3})\|V\|_2\|\Sigma^{1/2}\|_2 \, e^{\frac{5}{2}(Ax)^\top \Sigma  Ax} \bigl(d + 2\sqrt{d\ln(1/\delta)} + 2\ln(1/\delta)\bigr)^{1/2}.
\end{equation*}
This concludes the proof.
\end{proof}
In the particular case of Gaussians, a specific estimation can be found in Appendix \ref{app:ptwise_gaussian}.

\subsection{Mean convergence rate}\label{app:mean_cv}

We highlight the particular case of the mean of the distribution, which is a corollary of the previous result (with $f={\rm id}$).

\begin{corollary}[Mean convergence rate for sub-Gaussian tokens]\label{coro:mean_convergence_rate}
Let $X$ be centered and sub-Gaussian with parameter matrix $\Sigma\succ0$ and define $H$ as in \cref{def:spectral_reach_arc}. Let us denote by $\mathbb{E}$ the expectation w.r.t. $\nu$ and $\mathbb{E}_n$ the expectation w.r.t. the empirical measure.  For $n$ i.i.d.\ tokens, with $ n \ge 4 \, (\frac{1}{\delta}-1) \, n^{1/8}$, with probability at least $1 - \delta$,
\[
\bigl\|\mathbb{E}_n[f_n(x)] - \mathbb{E} [f(x)]\bigr\|_2
\;=\;
P(\sqrt{\ln n})\cdot  O\, \!\bigl( n^{-\frac{1}{2(1+32 \cdot H^2)}}\bigr).
\]
where $P$ is a polynomial function. 
\end{corollary}
The same result holds for Gaussian distribution, with a sharper rate (see \cref{sec:mean_gaussian})
\begin{proposition}[Mean convergence rate for Gaussian tokens]\label{thm:Gaussian_convergence_rate}
For Gaussian tokens using the same notation we have, 
\[
\bigl\|\mathbb{E}_n[f_n(x)] - \mathbb{E} [f(x)]\bigr\|
=O\!\left( (\ln n)^{\frac{d+1}{2}}\; n^{-\frac{1}{2(1 + H^2)}}\right).
\]
\end{proposition}

\subsection{Mean squared error (MSE) convergence rate }\label{app:mse_cv}

\begin{corollary}[Mean-squared convergence rate for Lipschitz observables under sub-Gaussian tokens]
\label{coro:mse_convergence_rate_lipschitz}
Let $X$ be centered and sub-Gaussian with parameter matrix $\Sigma\succ0$, and define $H$ as in Definition \ref{def:spectral_reach_arc}. Let $h$ be an $L_0$-Lipschitz function, squared integrable with respect to $\nu$. Let $\mathbb{E}$ denote expectation with respect to $\nu$. For $n$ i.i.d.\ tokens, assuming $n \geq 4\Bigl(\frac1\delta -1\Bigr)n^{1/8},$
with probability at least $1-\delta$,
\[
\mathbb{E}\Bigl[
\bigl\|f_n(X)-f(X)\bigr\|_2^2
\Bigr]
\leq
\frac{P(\sqrt{\ln n},L_0)}{n^{\beta}},
\]
where $P$ is a polynomial function, and
\[
\beta := \frac{1}{2(1+32H^2)}.
\]
\end{corollary}
 The proof follows from the same truncation and bounded-region versus tail decomposition argument used to establish the moment bound.
 
\subsection{Deep composition of layers }\label{app:multi_layer}
    Let
\[
    F := f^L \circ \cdots \circ f^1
    \qquad\text{and}\qquad
    F_n := f_n^L \circ \cdots \circ f_n^1
\]
be respectively a stack of $L$ attention layers and its sparse-attention
approximation. For each $\ell\in\{1,\dots,L\}$, assume that $f_n^\ell$ is
constructed from $n$ i.i.d.\ subsampled tokens drawn from the layer-$\ell$
token distribution $\mu^\ell$.

Provided that the subsampling remains i.i.d.\ at each layer, local sketching errors accumulate through the network, with amplification governed by the Lipschitz constants of downstream attention layers.
\begin{corollary}\label{coro:multi_layer}
Let $F $ and $F_n$ be defined as above, and assume that all token distributions remain supported in a ball of radius $R$ after each layer $l \in [1, L]$. Then,
\[
\|F - F_n\|_{L^2(\mu^1)}
=
O\!\left(
\sum_{\ell=1}^L
\frac{R^{2(L-\ell)}}{n^{\beta_\ell}}
\right).
\]
\end{corollary}
\begin{proof}
Consider a stack of $L$ attention layers
\[
F = f^L \circ \cdots \circ f^1
\]
and its sparse attention approximation
\[
F_n = f_n^L \circ \cdots \circ f_n^1,
\]
where each $f_n^\ell$ is constructed from $n$ i.i.d.\ subsampled tokens drawn from the layer-$\ell$ token distribution $\mu^\ell$.
Applying the single-layer convergence bound given by \ref{thm:convergence_rate_lipschitz} at each depth yields
\[
\|f^\ell - f_n^\ell\|_{L^2(\mu^\ell)}
=
O(n^{-\beta_\ell}),
\]
where $0 < \beta_\ell < 1/2$ is the convergence exponent associated with layer $\ell$.
If all token distributions remain supported in a ball of radius $R$, each attention map is $O(R^2)$-Lipschitz, and a composition argument gives
\[
\|F - F_n\|_{L^2(\mu^1)}
=
O\!\left(
\sum_{\ell=1}^L
\frac{R^{2(L-\ell)}}{n^{\beta_\ell}}
\right).
\]
This concludes the proof.
\end{proof}

\subsection{Covariance convergence rate for sub-Gaussian tokens}\label{app:cv_cov}

A similar convergence rate holds for the covariance matrix. 
Let
\[
\operatorname{Cov}_n(f(\hat X)):=\frac1n\sum_{i=1}^n(f_n(x_i)-\bar{f}_n\big)(f_n(x_i)-\bar{f}_n\big)^\top,\quad
\bar{f}_n:=\frac1n\sum_{i=1}^n f_n(x_i),
\]
and
\[
\operatorname{Cov}(f(X)):=\mathbb E\big[f(X)f(X)^\top\big]-\mathbb E[f(X)]\mathbb E[f(X)^\top].
\]
Our main result regarding the convergence rate of the covariance is the following.

\begin{proposition}[Covariance convergence rate for sub-Gaussian tokens]\label{prop:cov_emp_vs_pop}
With the notation  $\operatorname{Cov}_n(f_n(\hat X))$ and $\operatorname{Cov}(f(X))$ defined above, for $n$ i.i.d.\ tokens with $ n \ge4\, (\frac{1}{\delta}-1)\,n^{1/8}$, with probability at least $1 - \delta$,
\begin{equation}\label{eq:final_rate_cov}
\big\|\operatorname{Cov}_n(f_n(\hat X))-\operatorname{Cov}(f(X))\big\|_2
\ \le\
P(\sqrt{\ln n})\,O\, \!\bigl( n^{-\frac{1}{2(1+16 \cdot H^2)}}\bigr).
\end{equation}
\end{proposition}

The proof follows the same steps as the one of Theorem \ref{thm:convergence_rate_lipschitz}. 
\begin{proof}
To alleviate the notation, set $g_i := \Gamma_{\nu}(x_i)$ and $\hat{g}_i := f_n(x_i)$ for all $i \in \{1,\ldots,n\}$, as well as the empirical means $\bar{g}_n = \frac{1}{n}\sum_{i=1}^n g_i$ and $\bar{\hat{g}}_n = \frac{1}{n}\sum_{i=1}^n \hat{g}_i$. We also denote $g := f(x)$ and $\hat{g} := f_n(x)$. Now, decompose
\begin{align*}
&\quad \quad \|\operatorname{Cov}_n(\hat{g}) - \operatorname{Cov}(g)\|_2\\
&\quad \le \|\operatorname{Cov}_n(\hat{g}) - \operatorname{Cov}_n(g)\|_2 + \|\operatorname{Cov}_n(g) - \operatorname{Cov}(g)\|_2 \\
&\quad = \left\|\frac{1}{n} \sum_{i=1}^n \big(\hat{g}_i \hat{g}_i^\top - g_i g_i^\top\big) - \big(\bar{\hat{g}}_n \bar{\hat{g}}_n^\top - \bar{g}_n \bar{g}_n^\top\big)\right\|_2 + \|\operatorname{Cov}_n(g) - \operatorname{Cov}(g)\|_2 \\
&\quad \le \mathcal{I} + \mathcal{J},
\end{align*}
where 
\begin{align*}
    \mathcal{I} &= \left\|\frac{1}{n} \sum_{i=1}^n \big(\hat{g}_i \hat{g}_i^\top - g_i g_i^\top\big) - \big(\bar{\hat{g}}_n \bar{\hat{g}}_n^\top - \bar{g}_n \bar{g}_n^\top\big)\right\|_2,\\
    \mathcal{J} &= \|\operatorname{Cov}_n(g) - \operatorname{Cov}(g)\|_2.
\end{align*}

\textbf{Step 1: Analysis of the second term $\mathcal J$.}
Recall from Lemma \ref{lem:growth_gamma_sub} that,
\begin{equation}
    \|f(x)\|_2 \le \alpha\|X\|_2,\qquad  \alpha:=2\,\|V\|_2\,\|\Sigma^{1/2}\|_2\,\|\Sigma^{1/2}A\|_2.\label{eq:def_alpha}
\end{equation}
As $X$ follows a sub-Gaussian distribution with scalar parameter $\sigma = \|\Sigma^{1/2}\|_2$, $\|X\|_2$ is sub-Gaussian with parameter $\sigma \sqrt{d}$ (see Lemma $1$ in \cite{Jin19}). Using the characterization of sub-Gaussianity via the tail bound $\mathbb{P}(\|X\|_2 \ge t) \le 2e^{-t^2/(2\sigma^2)}$, we deduce from Lemma \ref{lem:growth_gamma_sub} that $\|f(x)\|_2$ is sub-Gaussian. Moreover, since for every
$u \in \mathbb{S}^{d-1},$ $|\langle u, f(x)\rangle| \le \|f(x)\|_2 \le \alpha\|X\|_2$, it follows that 
$f(x)$ is a sub-Gaussian random vector.

From Proposition $2.1$ in \cite{Vershynin12b} applied to the sub-Gaussian vector  $f(x)$, 
$$\mathcal J  \lesssim_{\sigma,\delta} \sqrt{\frac{d}{n}}.$$

\textbf{Step 2: Analysis of the first term $\mathcal I$.}
Now bound $\mathcal I$, 
\begin{align*}
   \mathcal I&\le \Big\|\frac1n \sum_{i=1}^n \big(\hat g_i \hat g_i^\top - g_i g_i^\top\big) \Big\|_2 +\Big\| \bar{\hat g}_n \bar{\hat g}_n^\top -  \bar{g}_n \bar{g}_n^\top\Big\|_2 \\
&\le \Big\|\frac1n \sum_{i=1}^n \big(\hat g_i \hat g_i^\top - g_i \hat g_i^\top + g_i \hat g_i^\top - g_i g_i^\top\big) \Big\|_2 +\Big\| \bar{\hat g}_n \bar{\hat g}_n^\top - \bar{\hat g}_n  \bar{g}_n^\top + \bar{\hat g}_n  \bar{g}_n^\top -  \bar{g}_n  \bar{g}_n^\top\Big\|\\
&\le \frac1n \sum_{i=1}^n [\|\hat g_i - g_i \|_2 (\|\hat g_i\|_2 + \|g_i\|_2 )]+(\| \bar{\hat g}_n \|_2 + \|\bar{g}_n\|_2) \|\bar{\hat g}_n - \bar{g}_n \|_2 \\
&\le \mathcal I_1 + \mathcal I_2.
\end{align*}
where 
\begin{align*}
\mathcal{I}_1&=\frac1n \sum_{i=1}^n [\|\hat g_i - g_i \|_2 (\|\hat g_i\|_2 + \|g_i\|_2 )]\\
\mathcal{I}_2&=(\| \bar{\hat g}_n \|_2 + \|\bar{g}_n\|_2) \|\bar{\hat g}_n - \bar{g}_n \|_2.
\end{align*}
For $\mathcal{I}_1$,
\begin{align*}
    \mathcal{I}_1 &\le \frac1n \sum_{\|Bx_i\|_2 \le R} [\|\hat g_i - g_i \|_2 (\|\hat g_i\|_2 + \|g_i\|_2 )] + \frac1n \sum_{\|Bx_i\|_2 > R} (\|\hat g_i\|_2 + \|g_i\|_2 )^2 \\
     &\le  \mathcal{I}_{1,\le R} + \mathcal{I}_{1,> R}
\end{align*}
where
\begin{align*}
 \mathcal{I}_{1,\le R}&=\frac1n \sum_{\|Bx_i\|_2 \le R} [\|\hat g_i - g_i \|_2 (\|\hat g_i\|_2 + \|g_i\|_2 )], \\
\mathcal{I}_{1,> R}&=\frac1n \sum_{\|Bx_i\|_2 > R} (\|\hat g_i\|_2 + \|g_i\|_2 )^2.
\end{align*}
For $\|Bx_i\|_2 \le R$,  using the notation $N_R:=\sum_{i=1}^{n} \mathbf{1}_{\{\|x_i\|_2>R\}},$ and $
M_n:=\max_{1\le i\le n}\|x_i\|_2$, we have $\|\hat g_i\|_2 \le M_n \le R$, and $\|g_i\|_2 \le \alpha M_n \le \alpha R$, where $\alpha$ is defined in \eqref{eq:def_alpha}. Also, from Proposition \ref{prop:ptwise_convergence}, $\|\hat g_i - g_i \|_2 \le \frac{C_1}{\sqrt{n}}$. Hence,
$$\mathcal{I}_{1,\le R} \le (1 + \alpha)R\frac{C_1}{\sqrt{n}}.$$
For $\|Bx_i\|_2 > R$, 
$$\mathcal{I}_{1,> R} \le \frac{N_R}{n}M_n^2(1+ \alpha)^2 \le \Biggl(e^{-c(\frac{R}{\sigma} - C\sqrt{d} )^2} +\sqrt{\frac{1}{2n}\ln\!\Bigl(\frac{4}{\delta}\Bigr)}\Biggr) \cdot 
c^2\sigma^2\Biggl(\sqrt{d} + \sqrt{\ln \frac{2n}{\delta}}\Biggr)^2 (1 + \alpha)^2,$$
where we used the result of Proposition \ref{prop:tail_control}.

For $\mathcal I_2$,
\begin{align*}
    \mathcal I_2 &= \Bigl(\|\bar{\hat g}_n \|_2 + \|\bar{g}_n\|_2\Bigr) \|\bar{\hat g}_n - \bar{g}_n \|_2 \\ 
    &\le \Bigl(\frac1n \sum_{i=1}^{n} \|\hat g_i \|_2 + \frac1n \sum_{i=1}^{n} \|g_i\|_2\Bigr) \frac1n \sum_{i=1}^{n}\|\hat g_i - g_i \|_2 \\ 
&\le \Bigl(\frac1n \sum_{\|x_i\|_2 \le R} \|\hat g_i \|_2 + \frac1n \sum_{\|x_i\|_2 \le R} \|g_i\|_2\Bigr) \frac1n \sum_{\|x_i\|_2 \le R}\|\hat g_i - g_i \|_2  \\
&+  \Bigl(\frac1n \sum_{\|x_i\|_2 > R} \|\hat g_i \|_2 + \frac1n \sum_{\|x_i\|_2 > R} \|g_i\|_2\Bigr) \frac1n \sum_{\|x_i\|_2 \le R}\|\hat g_i - g_i \|_2 \\
&+ \Bigl(\frac1n \sum_{\|x_i\|_2 \le R} \|\hat g_i \|_2 + \frac1n \sum_{\|x_i\|_2 \le R} \|g_i\|_2\Bigr)\Bigl( \frac1n \sum_{\|x_i\|_2 > R}\|\hat g_i \|_2 + \frac1n \sum_{\|x_i\|_2 > R}  \|g_i\|_2\Bigr)\\
&+ \Bigl(\frac1n \sum_{\|x_i\|_2 > R} \|\hat g_i \|_2 + \frac1n \sum_{\|x_i\|_2  > R} \|g_i\|_2\Bigr)^2 \\
& \le \mathcal I_{2,\le R, \le R} + \mathcal I_{2,> R, \le R} + \mathcal I_{2,\le R, > R} + \mathcal I_{2,> R, > R}.
\end{align*}
where
\begin{align*}
    \mathcal I_{2,\le R, \le R} &= \Bigl(\frac1n \sum_{\|x_i\|_2 \le R} \|\hat g_i \|_2 + \frac1n \sum_{\|x_i\|_2 \le R} \|g_i\|_2\Bigr) \frac1n \sum_{\|x_i\|_2 \le R}\|\hat g_i - g_i \|_2 \\
    \mathcal I_{2,> R, \le R} &= \Bigl(\frac1n \sum_{\|x_i\|_2 > R} \|\hat g_i \|_2 + \frac1n \sum_{\|x_i\|_2 > R} \|g_i\|_2\Bigr) \frac1n \sum_{\|x_i\|_2 \le R}\|\hat g_i - g_i \|_2 \\
    \mathcal I_{2,\le R, > R} &= \Bigl(\frac1n \sum_{\|x_i\|_2 \le R} \|\hat g_i \|_2 + \frac1n \sum_{\|x_i\|_2 \le R} \|g_i\|_2\Bigr)\Bigl(\frac1n \sum_{\|x_i\|_2 > R}\|\hat g_i \|_2 + \frac1n \sum_{\|x_i\|_2 > R}  \|g_i\|_2\Bigr) \\
    \mathcal I_{2,> R, > R} &= \Bigl(\frac1n \sum_{\|x_i\|_2 > R} \|\hat g_i \|_2 + \frac1n \sum_{\|x_i\|_2  > R} \|g_i\|_2\Bigr)^2.
\end{align*}
The first term $\mathcal I_{2,\le R, \le R} $ is bounded by
$$\mathcal I_{2,\le R, \le R} \le (1 + \alpha)R\frac{C_1}{\sqrt{n}} + (\frac{N_R}{n}\, M_n (1 + \alpha))^2.$$
The crossed terms $ \mathcal I_{2,> R, \le R} $ and $\mathcal I_{2,\le R, > R}$ are bounded by,
$$\mathcal I_{2,> R, \le R}  \le \frac{N_R}{n} M_n (1 + \alpha) \cdot \frac{C_1}{\sqrt{n}}.$$
$$\mathcal I_{2,\le R, > R} \le (1 + \alpha)^2 M_n \frac{N_R}{n} \cdot \frac{C_1}{\sqrt{n}}.$$
And the last term is again bounded by
$$\mathcal I_{2,> R, > R} \le \frac{N_R}{n}M_n^2(1+ \alpha)^2. $$
Recall that $\frac{N_R}{n}M_n \le \Biggl(e^{-c(\frac{R}{\sigma} - C\sqrt{d} )^2} +\sqrt{\frac{1}{2n}\ln\!\Bigl(\frac{4}{\delta}\Bigr)}\Biggr) \cdot 
c\sigma\Biggl(\sqrt{d} + \sqrt{\ln \frac{2n}{\delta}}\Biggr)$. \\
Therefore, as in Theorem \ref{thm:convergence_rate_lipschitz}, optimizing over $R$ yields the optimal radius $R^{\star}$,
\begin{equation*}
R^{\star} \approx \sqrt{\frac{\ln(n)}{16\, \|\Sigma^{1/2} A B^{-1}\|^2 + \frac{4c}{\|\Sigma B^\top\|_2}}}.
\end{equation*}
which again we optimize in $B$ by choosing $B := \Sigma^{-1/2}$. Finally, 
$\mathcal I +  \mathcal J = P\bigl(\sqrt{\ln(n)}\bigr) \cdot O (n^{-\frac{1}{2(1 + 16 \cdot H^2)}})$ where $H := \|\Sigma^{1/2}A\Sigma^{1/2}\|_2,$ is the horizon parameter and $P$ is a polynomial function of $\ln n$, which concludes the proof. 
\end{proof}

\begin{proposition}[Uniform convergence of the attention map for compactly supported tokens]\label{prop:unif_convergence_compact}
Let $X \sim \nu$ be centered, with $\operatorname{Supp}(\nu) \subset B_{R_0}$. For any $R > 0$, $\delta > 0$, there exists a constant $C>0$, such that for $n \ge n_{min}(\delta, A, R, R_0) := 4e^{2 RR_0 \|A\|_2}\, \Bigl(\frac{1}{\delta} - 1\Bigr)$, with probability at least $1 - \delta$,
\begin{equation}\label{eq:unif_convergence_compact}
  \sup_{x \in B_R}\bigl\| \Gamma_{\hat \nu_n}(x ) -\Gamma_{ \nu}(x) \bigr\|_2
\;\le\;  \frac{q_{V,A,  R_0,d, \delta}}{\sqrt{n}}\cdot  e^{2RR_0 \|A\|_2},
\end{equation}
where $q_{V,A, R_0,d, \delta} :=   \|V\|_2 \cdot\frac{ c \, \sqrt{d}\, R_0 (2\, R_0^2 \|A\|_2 + R_0 \, \|A\|_2 + 2)}{\delta }.$ 
\end{proposition}

The proof of Proposition \ref{prop:unif_convergence_compact} is the same as the proof of Theorem \ref{thm:uniform_convergence} in Appendix \ref{app:proof_thm_unif_cv}, except for the step 2 of part 1, where
bounds on the Lipschitz constant and envelope function are derived differently. More precisely, Lemma \ref{lem:bound_lipschitz1},  Lemma \ref{lem:bound_lipschitz2} and Lemma \ref{lem:bound_envelope} needs to be replaced by Lemma \ref{lem:bound_lipschitz_compact1},  Lemma \ref{lem:bound_lipschitz_compact2}, and Lemma \ref{lem:bound_envelope_compact} respectively. 

\begin{lemma}[Bound on Lipschitz constant of numerator for compactly supported $\nu$]\label{lem:bound_lipschitz_compact1}
Let $f \in \mathcal F$, and $L$ defined in Lemma \ref{lem:bound_lipschitz1}, for $Y \sim \nu$, with support $\nu \subset B_{R_0}$, and $x \in B_R$, 
\begin{equation*}
    \|L\|_{L^2(P)} \le R_0^2\,e^{R R_0\|A\|_2}.\end{equation*}
\end{lemma}

\begin{proof}
Consider $f(t,y) := y\, e^{\langle t, y\rangle}$. Then, with $t = Ax$, $\nabla_t f(t,y) = A^\top \nabla_x f(x,y) $. The Jacobian of $f(\cdot,y)$ at $t$ is
\[
\nabla_t f(t,y)\;= y\,
y^\top \;e^{\langle t, y\rangle} \, \in\mathbb{R}^{d\times d},
\]
\begin{align*}\|\nabla_t f(t,y)\|_{\mathrm{L^2}(\nu)}&:=\Big(\mathbb{E}_{Y\sim\nu}\|\nabla_t f(t,Y)\|_{2}^2\Big)^{1/2} \\
&=\, \Big(\mathbb{E}_{Y\sim\nu}\big[\|Y\|_2^{4} \, e^{2\langle t, y\rangle} \big]\Big)^{1/2} \\
&\le R_0^2 \Big(\mathbb{E}_{Y\sim\nu}\big[ e^{2\langle Ax, Y\rangle} \big]\Big)^{1/2}\\
&\le R_0^2 \, e^{R R_0 \|A\|_2}
\end{align*}
where we used Cauchy-Schwarz to bound 
$\langle t, z\rangle \le \|Ax\|_2 \|Y\|_2 \le  R R_0  \|A\|_2 $.
\end{proof}

\begin{lemma}[Bound on Lipschitz constant of denominator for compactly supported $\nu$]\label{lem:bound_lipschitz_compact2}
Let $f \in \mathcal F'$, and $L$ defined in  Lemma \ref{lem:bound_lipschitz2}, and let $Y \sim \nu$, with support $\nu \subset B_{R_0}$, and $x \in B_R$. Then, 
$$\|L\|_{L^2(P)} \le \|A\|_2 \, R_0 \, e^{R R_0 \|A\|_2}.$$
\end{lemma}
\begin{proof}
Consider $f(t,y) :=  e^{\langle t, y\rangle}$. Then, with $t = Ax$, $\nabla_x f(x,y) = A^\top \nabla_t f(Ax,y) $.\\
The Jacobian of $f(\cdot,y)$ at $t$ is
\[
\nabla_t f(t,y)\;= y\,e^{\langle t, y\rangle} \, \in\mathbb{R}^{d\times d},
\]
\begin{align*}\|\nabla_t f(t,y)\|_{\mathrm{L^2}(\nu)}&:=\Big(\mathbb{E}_{Y\sim\nu}\|\nabla_t f(t,Y)\|_{2}^2\Big)^{1/2} \\
&=\, \Big(\mathbb{E}_{Y\sim\nu}\big[\|Y\|_2^{2} \, e^{2\langle t, y\rangle} \big]\Big)^{1/2} \\
&\le R_0 \, e^{R R_0 \|A\|_2}.
\end{align*}
\end{proof}
\begin{lemma}[Bound on envelope function for compactly supported $\nu$]\label{lem:bound_envelope_compact}
For $Y \sim \nu$, with support $\nu \subset B_{R_0}$, and $x \in B_R$, 
$$\big(\mathbb E[F(Y)^2]\big)^{1/2}
\ \le R_0 \,e^{R R_0 \|A\|_2}.$$
\end{lemma}
\begin{proof}
\[
\mathbb E[F(Y)^2] = \mathbb E\!\big[\|Y\|_2^2\,e^{\,2R\|A^\top Y\|_2}\big]
\ \le\
R_0^2 \,e^{2 R R_0 \|A\|_2} .
\]
\end{proof}
To apply those results to prove Proposition \ref{prop:unif_convergence_compact}, recall equation \eqref{eq:bound_lip_envel} for the numerator of attention in the proof of Theorem  \ref{thm:uniform_convergence} in Appendix \ref{app:proof_thm_unif_cv},
\begin{equation*}
    \mathbb E[\|\mathbb P_n - \mathbb P\|_{\mathcal{F}}] \leq  \frac{c \, \sqrt{d}}{\sqrt{n}} \Big(  R_0\, \mathbb E[\|L\|_{L^2(\mathbb P_n)}] + \, \mathbb E[\|F\|_{L^2(\mathbb P_n)}]\Big).
\end{equation*}
Applying Lemma \ref{lem:bound_lipschitz_compact1} and Lemma \ref{lem:bound_envelope_compact} yields, 
\begin{align*}
    \mathbb E[\|\mathbb P_n - \mathbb P\|_{\mathcal{F}}] &\leq  \frac{c \, \sqrt{d}}{\sqrt{n}} \Big(  \|A\|_2 \, R_0^3 \, e^{RR_0 \|A\|_2} + \, R_0 \,e^{RR_0 \|A\|_2 }\Big)\\
    &\leq  \frac{c \, \sqrt{d}}{\sqrt{n}} R_0 \, (  \|A\|_2 \, R_0^2  + \, 1 )\, e^{RR_0 \|A\|_2}.
\end{align*}
To control the denominator of attention, we apply Lemma \ref{lem:bound_lipschitz_compact2} to equation \eqref{eq:dudley_den_eq}, 
\begin{align}
 E[\|\mathbb P_n - \mathbb P\|_{\mathcal{F'}}] \leq \frac{c \, }{ \delta \, \sqrt{n}} \, 
\|A\|_2 \, R_0 \, e^{RR_0 \|A\|_2} .
\end{align}
Combining those results, the conclusion of part 3 in Appendix \ref{app:proof_thm_unif_cv} becomes, 
$$\|N\|_2 = \|\mathbb E[Y e^{\langle Ax, Y\rangle}]\|_2 \le R_0 \, e^{RR_0 \|A\|_2},$$
$$\|\frac{N_n}{D_n} - \frac{N}{D}\|_2 \le \|V\|_2 \cdot\frac{ c \, \sqrt{d}\, R_0 (2\, R_0^2 \|A\|_2 + R_0 \, \|A\|_2 + 2)}{\delta } \cdot \frac{1}{\sqrt{n}} \cdot e^{2RR_0 \|A\|_2}.
$$

\subsection{Compactly supported tokens}\label{app:compact_lip_cv}

From Proposition \ref{prop:unif_convergence_compact}, we deduce the following estimation in the compact support setting. 

\begin{proposition}[Lipschitz functional convergence rate for compactly supported tokens]\label{prop:compact_convergence_rate_mean}
Let $X \sim \nu$ centered with $\operatorname{Supp}(\nu) \subset B_{R_0}$, and let $h : \mathcal L^2(\nu) \rightarrow \mathcal L^2(\nu)$  be a Lipschitz function with Lipschitz constant $L_0$. Recall the definition of the empirical distribution $\nu_n = \frac1n \sum_{i=1}^{n} \delta_{Y_i}$. Let us denote by $\mathbb{E}$ the expectation w.r.t. $\nu$, and $\mathbb{E}_n$ the expectation w.r.t. the empirical measure. For $n$ i.i.d.\ tokens, with probability at least $1 - \delta$,
\[
\left\|\mathbb{E}_n[h \circ f_n(\hat X)] - \mathbb{E}[h \circ f(X)]\right\|_2 
\leq \frac{1}{\sqrt{n}} \cdot L_0 \cdot \left(q_{V,A, R_0,d, \delta} \, e^{2\cdot  R_0^2\|A\|_2 } +  q'_{V, R_0, d, \delta}\right)
\]
where $q_{V,A, R_0,d, \delta}$ is defined in Proposition \ref{prop:unif_convergence_compact}, and $q'_{V, R_0, d, \delta} := \|V\|_2 \, q_{d,1-\delta}^{1/2} R_0^2$.
\end{proposition}

\subsection{Gaussian case}

This section provides additional results for the Gaussian token distribution, including pointwise convergence rates, one-dimensional analysis, and detailed proofs of auxiliary lemmas used in the main text.

\begin{lemma}[Exact linearity under Gaussian inputs (Castin and al. 2025)]\label{lem:Gaussian_linearity}
If $X\sim\mathcal N(0,\Sigma)$, then for all $x$, the log-MGF is quadratic and the population attention map is \emph{exactly linear}:
\[
f(x)=V\,\nabla K(Ax)=V\,\Sigma A x .
\qquad
K(t):= \ln \mathbb E[e^{\langle t, X \rangle}] = \tfrac12\,t^\top\Sigma t.
\]
Hence the Gaussian family is preserved by attention, and first/second moments determine the dynamics.
\end{lemma}
\subsubsection{Pointwise convergence} \label{app:ptwise_gaussian}
We begin with concentration bounds for fixed query vectors under Gaussian token distribution.
\begin{lemma}[High-probability concentration (Gaussian constants)]\label{lem:Gaussian_convergence}
For any fixed query $x \in \mathbb R^d$ and $\delta\in(0,1)$, with probability at least $1-\delta$,
\[
\bigl\|f_n(x)-f(x)\bigr\|_2
\;\le\;
\frac{C(\Sigma,A,x,\delta,d)}{\sqrt n},
\]
where
\[
C(\Sigma,A,x,\delta,d)
= \|V\|_{2}\cdot q_{A, x, d, \delta}^{\mathcal N} \cdot e^{\frac12\,\|\Sigma^{1/2}Ax\|_2^2}  .
\]
$q_{A, x, d, \delta}^{\mathcal N} := \sqrt{\,\|\Sigma\|_2+\|\Sigma Ax\|_2^{\,2}\,}\;\; q_{d,1-\delta}^{1/2}.$
\end{lemma}

\subsubsection{Dimension 1}
In the one-dimensional case, sharper convergence rates can be obtained with explicit constants.
\begin{lemma}[Concentration in $d=1$]\label{lem:dim1_convergence}
With probability at least $1-\delta$,
\[
\bigl\|f_n(x)-f(x)\bigr\|_2
\;\le\;
\frac{C(\sigma,a,x,\delta)}{\sqrt n},
\qquad
C(\sigma,a,x,\delta)
= |V|\; q_{1-\delta/2}\; \sigma\;
\sqrt{1+\sigma^2 a_x^{2}}\; \exp\!\Bigl(\tfrac12\sigma^2 a_x^{2}\Bigr)\;  .
\]
\end{lemma}

\begin{proposition}[Mean convergence rate in $d=1$]\label{thm:dim1_convergence_rate}
Let $H=(a\sigma) \cdot\sigma$. Then,
\[
\bigl\|\mathbb E_n[f_n(x)]-\mathbb E_{X \sim \hat{\nu_n}}[f(x)]\bigr\|
=\Theta\!\left(\ln n\; n^{-\frac{1}{2(1+H^2)}}\right).
\]\end{proposition}

\subsubsection{Uniform convergence}\label{uniform_gaussian}
\begin{proposition}[Uniform convergence of the attention map for Gaussian tokens]\label{prop:unif_cv_gaussian}
Let $X \sim \nu$ be centered and sub-Gaussian with parameter matrix $\Sigma \succ0$. For any $R > 0$, $\delta > 0$, there exists a constant $C>0$, such that for $n \ge n_{min}(\delta, \Sigma, A, R) := 4e^{2 R^2 \|\Sigma^{1/2}A\|_2^2}\, \Bigl(\frac{1}{\delta} - 1\Bigr)$, with probability at least $1 - \delta$,
\begin{equation}\label{eq:unif_cv_gaussian}
  \sup_{x \in B_R}\bigl\| f_n(x) -f(x) \bigr\|_2
\;\le\;  q_{(\Sigma, A, V, R, d, \delta)}^{\mathcal N}\cdot \frac{e^{R^2\|\Sigma^{1/2}A\|_2^2}}{\sqrt{n}},
\end{equation}
where $q_{(\Sigma, A,V, R, d, \delta)}^{\mathcal N}$ is defined in Appendix \ref{app:gaussian_derivations}. 
\end{proposition}

\subsubsection{Proof of Proposition \ref{prop:unif_cv_gaussian}}\label{app:gaussian_derivations}
The proof of Proposition  \ref{prop:unif_cv_gaussian} is the same as the one of Theorem \ref{thm:uniform_convergence} in Appendix \ref{app:proof_thm_unif_cv}, except for step 2 of part 1, where bounds on the Lipschitz constant and envelope function are tighter for Gaussian distributions. More precisely, Lemma \ref{lem:bound_lipschitz1} , Lemma \ref{lem:bound_lipschitz2} and Lemma \ref{lem:bound_envelope} needs to be replaced by Lemma \ref{lem:bound_lipschitz_gaussian1} , Lemma \ref{lem:bound_lipschitz_gaussian2} and Lemma \ref{lem:bound_envelope_gaussian} respectively.

\begin{lemma}[Bound on Lipschitz constant for numerator for Gaussian distribution]\label{lem:bound_lipschitz_gaussian1}
Let $f \in \mathcal F$, and $L$ defined in  Lemma \ref{lem:emp_lip}, for $Y \sim \mathcal N(0, \Sigma)$, and $x \in B_R$,
$$\|L\|_{L^2(P)} \le c(\Sigma, A, R) \exp\!\big(R^2\|\Sigma^{1/2}A\|_2^2\big),$$
where $c(\Sigma, A, R)$ is a polynomial function in $\|A\|_2$, $\|\Sigma\|_2$, $R$, and $\operatorname{tr}(\Sigma)$.
\end{lemma}
\begin{proof}
Consider $f(t,y) := y\, e^{\langle t, y\rangle}$ - we will then apply the result to $t = Ax$ using $\nabla_t f(t,y) = A^\top \nabla_x f_x(y) $.\\
The Jacobian of $f(\cdot,y)$ at $t$ is
\[
\nabla_t f(t,y)\;= y\,
y^\top \;e^{\langle t, y\rangle} \, \in\mathbb{R}^{d\times d},
\]
Let $\|\nabla_t f(t,y)\|_{\mathrm{L^2}(\nu)}:=\Big(\mathbb{E}_{Y\sim\nu}\|\nabla_t f(t,Y)\|_{2}^2\Big)^{1/2} =\, \Big(\mathbb{E}_{Y\sim\nu}\big[\|Y\|_2^{4} \, e^{2\langle t, y\rangle} \big]\Big)^{1/2},$
where we used the fact that $YY^\top$ is a rank one matrix, hence $\|Y Y^\top \| = \|Y\|_2^2$. 

Notice that, for any measurable function $g$, 
\begin{align*}
    \mathbb E[g(Y) e^{\langle t, y\rangle} ] &= \int g(y) e^{\langle t,y\rangle} \phi_\Sigma(y) \, dy = \int g(y) \exp\left(\langle t, y\rangle - \frac{1}{2}y^\top \Sigma^{-1}y\right) \frac{dy}{(2\pi)^{d/2} \det(\Sigma)^{1/2}}\\
&= \int g(y) \exp\left(-\frac{1}{2}(y - \Sigma t)^\top \Sigma^{-1}(y - \Sigma t) + \frac{1}{2}t^\top \Sigma t\right) \frac{dy}{(2\pi)^{d/2} \det(\Sigma)^{1/2}} \\
&= e^{\frac{1}{2}t^\top \Sigma t} \int g(y) \phi_\Sigma(y - \Sigma t) \, dy = e^{\frac{1}{2}t^\top \Sigma t} \mathbb{E}\left[g(Y + \Sigma t)\right].
\end{align*}
Using this property, we can write: 
\[
\|\nabla_t f(t,y)\|_{\mathrm{L^2}(\nu)}=e^{t^\top \Sigma t} \, \Big(\mathbb{E}_{Y\sim\nu}\big[\|Y + \Sigma t\|_2^{4} \big]\Big)^{1/2},
\]
Note that $t^\top\Sigma t = x^\top A^\top\Sigma A\,x \le \|\Sigma^{1/2}A\|_2^2 \|x\|_2^2$, and take the supremum over $\|x\|_2\le R$:
\[
\|L\|_{L^2(P)} :=\sup_{\|x\|_2\le R}\|\nabla_x f_x(y)\|_{\mathrm{L^2}(\nu)}
\ \le\ c(\Sigma, A)
\ \exp\!\big( \,R^2\|\Sigma^{1/2}A\|_2^2\big),
\]
where $c(\Sigma, A)$ is a polynomial function in $\|A\|_2$, $\|\Sigma\|_2$, and $\operatorname{tr}(\Sigma)$.
\end{proof}
\begin{lemma}[Bound on Lipschitz constant for denominator for Gaussian distribution]\label{lem:bound_lipschitz_gaussian2}
Let $f \in \mathcal F$, and $L$ defined in  Lemma \ref{lem:emp_lip}, for $Y \sim \mathcal N(0, \Sigma)$, and $x \in B_R$
$$\|L\|_{L^2(P)} \le c'(\Sigma, A,R) \exp\!\big(R^2\|\Sigma^{1/2}A\|_2^2\big),$$
where $c'(\Sigma, A, R)$ is a polynomial function in $\|A\|_2$, $\|\Sigma\|_2$, $R$, and $\operatorname{tr}(\Sigma)$.
\end{lemma}
\begin{proof}
Consider $f(t,y) :=  e^{\langle t, y\rangle}$ - we will then apply the result to $t = Ax$ using $\nabla_x f(x,y) = A^\top \nabla_t f(Ax, y) $. The Jacobian of $f(\cdot,y)$ at $t$ is
\[
\nabla_t f(t,y)\;= y\,e^{\langle t, y\rangle} \, \in\mathbb{R}^{d\times d},
\]
Let $\|\nabla_t f(t,y)\|_{\mathrm{L^2}(\nu)}:=\Big(\mathbb{E}_{Y\sim\nu}\|\nabla_t f(t,Y)\|_{2}^2\Big)^{1/2},$

As derived in the proof of Lemma , for any mesurable function $g$, 
\begin{align*}
    \mathbb E[g(Y) e^{\langle t, y\rangle} ] = e^{\frac{1}{2}t^\top \Sigma t} \mathbb{E}\left[g(Y + \Sigma t)\right].
\end{align*}
Using this property, 
\[
\|\nabla_t f(t,y)\|_{\mathrm{L^2}(\nu)}=e^{t^\top \Sigma t} \, \Big(\mathbb{E}_{Y\sim\nu}\big[\|Y + \Sigma t\|_2^{2} \big]\Big)^{1/2},
\]
Use $t^\top\Sigma t = x^\top A^\top\Sigma A\,x \le \|\Sigma^{1/2}A\|_2^2 \|x\|_2^2$, and take the supremum over $\|x\|_2\le R$:
\[
\|L\|_{L^2(P)} :=\sup_{\|x\|_2\le R}\|\nabla_x f_x(y)\|_{\mathrm{L^2}(\nu)}
\ \le\ c'(\Sigma, A)
\ \exp\!\big( \,R^2\|\Sigma^{1/2}A\|_2^2\big),
\]
where $c'(\Sigma, A, R)$ is a polynomial function in $\|A\|_2$, $\|\Sigma\|_2$, $R$, and $\operatorname{tr}(\Sigma)$.
\end{proof}
\begin{lemma}[Bound on envelope function for Gaussians]\label{lem:bound_envelope_gaussian}
For $Y \sim \mathcal N(0, \Sigma)$, and $x \in B_R$,
$$\big(\mathbb E[F(Y)^2]\big)^{1/2}
\ \le \|\Sigma^{1/2}\|_2 \, \Bigl( 2^{d+1}d\, + \frac{\sqrt{2 \pi} \, 2^{\frac{d}{2}+1}}{\Gamma(\frac{d}{2})} \, (2 R \|\Sigma^{1/2}A\|_2)^{d+1} \Bigr)\, e^{R^2 \|\Sigma^{1/2}A\|_2^2}.$$
\end{lemma}
\begin{proof}
Write $Y := \Sigma^{1/2}Z$, where $Z \sim \mathcal N(0, I_d)$, and notice that $2R\|A^\top Y\|_2 \le 2 R \|\Sigma^{1/2}A\|_2 \|Z\|_2 := t \|Z\|_2 $, where $t := 2 R \|\Sigma^{1/2}A\|_2$. Then:
\[
\mathbb E[F(Y)^2] = \mathbb E\!\big[\|Y\|_2^2\,e^{\,2R\|A^\top Y\|_2}\big]
\ \le\
\|\Sigma^{1/2}\|_2^{2} \, \mathbb E[\|Z\|_2^{2}e^{\,t \|Z\|_2}] = \|\Sigma\|_2 \, \mathbb E[\|Z\|_2^{2}e^{\,t \|Z\|_2}].
\]
As $\|Z\|_2 \sim \chi_d$, writing $c_d := \frac{2^{1 - \frac{d}{2}}}{\Gamma(\frac{d}{2})}$, one can explicitly compute: 
\begin{align*}
    E[\|Z\|_2^{2}e^{\,t \|Z\|_2}] &= c_d\,\int_0^\infty r^{d+1}e^{-r^2/2}e^{tr}\,dr =c_d\,e^{t^2/2}\int_0^\infty r^{d+1}e^{-(r-t)^2/2}\,dr.
\end{align*}
Making the change of variable $r-t=s$, 
\begin{align*}
    \int_0^\infty r^{d+1}e^{-(r-t)^2/2}\,dr &= \int_{-t}^{\infty}(s+t)^{d+1}e^{-s^2/2}\,ds \\
&\le \int_{-\infty}^{\infty}(|s|+t)^{d+1}e^{-s^2/2}\,ds \\
&\le  \int_{-\infty}^{\infty}\underbrace{2^{d}(|s|^{d+1}+t^{d+1})}_{(\star)}e^{-s^2/2}\,ds,
\end{align*}
where we use $(a+b)^{d+1} \le 2^{d}(a^{d+1}+b^{d+1})
$ in $(\star)$. 

Then use symmetry to compute
\[
\int_{-\infty}^{\infty}|s|^{d + 1} e^{-s^2/2}\,ds
= 2\int_0^\infty s^{d+1} e^{-s^2/2}\,ds
= 2\cdot 2^{\frac{d}{2}}\Gamma\!\left(\frac{d+2}{2}\right)
=2^{\frac{d+2}{2}}\Gamma\!\left(\frac{d+2}{2}\right).
\]
Combining these results, we obtain that,
\begin{align*}
    E[\|Z\|_2^{2}e^{\,t \|Z\|_2}] &\le c_d\,2^d(2^{\frac{d+2}{2}}\Gamma\!\left(\frac{d+2}{2}\right) + \sqrt{2 \pi}\, t^{d+1})\, e^{t^2/2}= ( 2^{d+1}d\, + \frac{\sqrt{2 \pi} \, 2^{\frac{d}{2}+1}}{\Gamma(\frac{d}{2})} \, t^{d+1} )\, e^{t^2/2}.
\end{align*}
Finally, with $t = 2 R \|\Sigma^{1/2}A\|_2$,
\[
\big(\mathbb E[F(Y)^2]\big)^{1/2}
\ \le \|\Sigma^{1/2}\|_2 \, \Bigl( 2^{d+1}d\, + \frac{\sqrt{2 \pi} \, 2^{\frac{d}{2}+1}}{\Gamma(\frac{d}{2})} \, (2 R \|\Sigma^{1/2}A\|_2)^{d+1}\Bigr)\, e^{R^2 \|\Sigma^{1/2}A\|_2^2},
\]
which is exactly the stated inequality.
\end{proof}
Recall equation \eqref{eq:bound_lip_envel} for the numerator of attention in the proof of Theorem  \ref{thm:uniform_convergence} in Appendix \ref{app:proof_thm_unif_cv},
\begin{equation*}
    \mathbb E[\|\mathbb P_n - \mathbb P\|_{\mathcal{F}}] \leq  \frac{c \, \sqrt{d}}{\sqrt{n}} \Big(  R\, \mathbb E[\|L\|_{L^2(\mathbb P_n)}] + \, \mathbb E[\|F\|_{L^2(\mathbb P_n)}]\Big).
\end{equation*}
Applying Lemma \ref{lem:bound_lipschitz_gaussian1} and Lemma \ref{lem:bound_envelope_gaussian} yields, 
\begin{align*}
    \mathbb E[\|\mathbb P_n - \mathbb P\|_{\mathcal{F}}] \leq  \frac{c \, \sqrt{d}}{\sqrt{n}} \Big(  R \, c(\Sigma, A) + C(d, \Sigma, A, R) \Big)\, e^{R^2 \|\Sigma^{1/2}A\|_2^2}, 
\end{align*}
where $C(d, \Sigma, A , R) := \|\Sigma^{1/2}\|_2 \, ( 2^{d+1}d\, + \frac{\sqrt{2 \pi} \, 2^{\frac{d}{2}+1}}{\Gamma(\frac{d}{2})} \, (2 R \|\Sigma^{1/2}A\|_2)^{d+1}).$

To control the denominator of attention, applying Lemma \ref{lem:bound_lipschitz_gaussian2} to equation \eqref{eq:dudley_den_eq}, 
\begin{align}
 E[\|\mathbb P_n - \mathbb P\|_{\mathcal{F'}}] \leq \frac{c'(\Sigma, A) \, }{ \delta \, \sqrt{n}} \, 
e^{R^2 \|\Sigma^{1/2}A\|_2^2}.
\end{align}
Combining those results, the conclusion of part 3 in Appendix \ref{app:proof_thm_unif_cv} becomes, 
$$\frac{\|N\|_2 }{D}= \frac{\|\mathbb E[Y e^{\langle Ax, Y\rangle}]\|_2  }{\mathbb E[e^{\langle Ax, Y\rangle}] }= \|\Sigma A x\|_2 \le \|\Sigma A \|_2 \, R, \text{ and } \|\frac{N_n}{D_n} - \frac{N}{D}\|_2 \le \|V\|_2 \cdot\frac{  c(\Sigma, A, R, d)}{\delta } \cdot \frac{1}{\sqrt{n}} \cdot e^{R^2 \|\Sigma^{1/2}A\|_2^2 },
$$
where 
$$q_{(\Sigma, A,V, R, d, \delta)}^{\mathcal N} = \frac{2 \, c \|V\|_2 \sqrt{d}}{\delta}  (R \, c(\Sigma, A, R) + C(d, \Sigma, A , R) + \|\Sigma A \|_2 \, R \, c'(\Sigma, A, R) ).$$

\subsubsection{Uniform convergence}\label{uniform_gaussian_B}
\begin{proposition}[Uniform convergence of the attention map for Gaussian tokens]\label{prop:unif_cv_gaussian_B}
Let $X \sim \nu$ be centered and sub-Gaussian with parameter matrix $\Sigma \succ0$. For any $R > 0$, $\delta > 0$, and invertible matrix $B$, there exists a constant $C>0$, such that for $n \ge n_{min}(\delta, \Sigma, A, R) := 4e^{2 R^2 \|\Sigma^{1/2}AB^{-1}\|_2^2}\, \Bigl(\frac{1}{\delta} - 1\Bigr)$, with probability at least $1 - \delta$,
\begin{equation}\label{eq:unif_cv_gaussian_B}
  \sup_{x \in B_R}\bigl\| f_n(x) -f(x) \bigr\|_2
\;\le\;  q_{(\Sigma, A, V, R, d, \delta)}^{\mathcal N}\cdot \frac{e^{R^2\|\Sigma^{1/2}AB^{-1}\|_2^2}}{\sqrt{n}},
\end{equation}
where $q_{(\Sigma, A,V, R, d, \delta)}^{\mathcal N}$ is defined in Appendix \ref{app:gaussian_derivations}. 
\end{proposition}
The proof follows from the same argument as for \cref{thm:uniform_cv_B}.

\subsubsection{Mean convergence rate}\label{sec:mean_gaussian}
\begin{proposition}[Mean convergence rate for Gaussian tokens]\label{prop:Gaussian_convergence_rate}
For Gaussian tokens using the same notation with $H = \|\Sigma^{1/2}A\Sigma^{1/2}\|_2$we have, 
\[
\bigl\|\mathbb{E}_n[f_n(x)] - \mathbb{E} [f(x)]\bigr\|
=O\!\left( (\ln n)^{\frac{d+1}{2}}\; n^{-\frac{1}{2(1 + H^2)}}\right).
\]
\end{proposition}
The proof follows from applying  \cref{thm:uniform_cv_B} with the same arguments as in \cref{thm:convergence_rate_lipschitz}.

\section{Improving the bounds}

Our goal in this section is to show the following improvement of Theorem \ref{thm:uniform_convergence} in the compactly supported setting, where the exponential dependency of the bound in the radius $R$ is replaced by a polynomial dependency. 
\begin{theorem}\label{thm:uniform_convergence_improved}
Let $Y \sim \nu$ be centered and with compact support $K\subset\mathbb{R}^d$, such that $\|Y\| \le R_0$ a.s. for some $R_0 > 0$. We assume that $x \mapsto \sup_{y\in K}\langle Ax,y\rangle$ is differentiable.
% $\partial K$ is $C^1$. 
Let $\varphi: \mathbb{R}^d \rightarrow \mathbb{R}$ defined by
$$ 
\varphi(t)=\sup_{y\in K} \langle t,y\rangle -\log\mathbb{E}[e^{\langle t,Y\rangle}]. 
$$
We assume  
\begin{equation}\label{e:phi(t)}
\sup_{|t|\leq t_0} \varphi(t) \leq C_d \log(t_0)
\end{equation}
for some $C_d >0$
as $t_0\rightarrow +\infty$. Then
for any $R > 0$, $\delta > 0$, there exists a constant $C_d >0$, such that for $n \ge n'_{min}(\delta, \Sigma, A, R)$, with probability at least $1 - \delta$, 
\begin{equation}\label{eq:uniform_convergence_improved}
  \sup_{x \in B_R}\bigl\| f_n(x) -f(x) \bigr\|_2
\;\le\;  \frac{q'_{(\Sigma, A, V, R, R_0, d, \delta)}}{\sqrt{n}}  ,
\end{equation}
where $q'_{(\Sigma, A, V, R, R_0, d, \delta)}:= \|V\|_2 \frac{C \,\|\Sigma\|_2^{1/2} \, \sqrt{d} \, M}{\delta }\, \bigl[(R + M)(\|\Sigma\|_2^{1/2} \, \sqrt{d}\, \|A\|_2 + \sup_{x\in B_R}\|\nabla_x  \sup_{y\in K}\langle Ax,y\rangle\|_2) + 1 \bigr],$  with $M := (\|A\|_2 R R_0)^{C_d }$
\end{theorem}
\begin{remark}
The condition \eqref{e:phi(t)} means that in each direction $v$, the pushforward of $\nu$ onto $\mathbb{R} v$ has a density close to its edges, and that this density has at most polynomial decay close to these edges. This is the case in most concrete examples, for example when $\nu$ is uniform on a regular open set $\Omega$, in any dimension. Note also that $x \mapsto \sup_{y\in K}\langle Ax,y\rangle$ is differentiable if, for instance $\partial K$ is $C^1$.
\end{remark}
\begin{proof}[Proof of Theorem \ref{thm:uniform_convergence_improved}]
The main idea is to write the continuous attention map as
$$f(x) = \frac{\mathbb E[Y \, e^{\langle Ax, Y\rangle-\alpha(x)}]}{\mathbb E[e^{\langle Ax, Y\rangle-\alpha(x)}]}$$
with $\alpha(x)$ defined for any $x\in \mathbb{R}^d$ as
$$
\alpha(x)= \sup_{y\in K}\langle Ax,y\rangle.
$$
Then instead of \eqref{eq:def_f} and \eqref{eq:F} we let
\begin{equation}\label{e:fxApp}
f_x:y\mapsto ye^{\langle Ax,y\rangle-\alpha(x)} 
\end{equation}
and 
$$
F:y\mapsto \sup_{x\in B_R} \|f_x(y)\|_2\leq \|y\|_2.
$$
Lemma \ref{lem:sym} is unchanged. Lemma \ref{lem:bound_lipschitz1} becomes:
\begin{lemma}[Bound on Lipschitz constant for numerator]\label{lem:bound_lipschitz1_improved}
Let $R>0$. Let $f \in \mathcal F := \{f_x, \,  x \in B_R \}$ as defined in \eqref{e:fxApp}. With an abuse of notation, denote $f : (x,y) \mapsto y\, e^{\langle Ax, y \rangle-\alpha(x)}$. Then, $f(\cdot,y)$ is Lipschitz with respect to $x$. Let $y \mapsto L(y)$ be its Lipschitz constant. Consider $Y \sim \nu$, a sub-Gaussian random vector in $\mathbb R^d$ of matrix sub-Gaussian parameter $\Sigma$, and scalar sub-Gaussian parameter $\sigma := \sqrt{\|\Sigma\|_2}$. Then, $L$ verifies
$$\|L\|_{L^2(\nu)} = \sup_{x\in B_R}\|\nabla_xf_x\|_{L^2(\nu)}\leq C\sigma\sqrt{d}\, (\,\sigma \sqrt{d}\|A\|_2+\sup_{x\in B_R}\|\nabla \alpha(x)\|_2),$$
for some constant $C > 0$.
\end{lemma}
\begin{proof} The proof is essentially the same as that of Lemma \ref{lem:bound_lipschitz1}. The only difference is that 
$$
\nabla_x f_x(y)=y(A^\top y-\nabla\alpha(x))^\top e^{\langle Ax,y\rangle -\alpha(x)}.
$$
Using that $\alpha(x)\geq \langle Ax,y\rangle$ and the Cauchy-Schwarz inequality, we get
\begin{align*}
    \|\nabla_xf_x\|_{L^2(\nu)} &\leq \sqrt{2\bigl( \mathbb{E}_{Y\sim \nu}[\|Y\|_2^4]\cdot\|A\|_2 ^2 + \mathbb{E}_{Y\sim \nu}[\|Y\|^2_2]\cdot \|\nabla \alpha(x)\|_2^2\bigr)} \\ 
    &\leq \sqrt{2}\bigl( \mathbb{E}_{Y\sim \nu}[\|Y\|_2^4]^{1/2}\cdot\|A\|_2  + \mathbb{E}_{Y\sim \nu}[\|Y\|^2_2]^{1/2}\cdot \|\nabla \alpha(x)\|_2\bigr) \\
    &\leq C\bigl( \sigma^2 d\|A\|_2  + \sigma \sqrt{d}\cdot \|\nabla \alpha(x)\|_2\bigr)
\end{align*}
Hence the statement follows.
\end{proof}

Lemma \ref{lem:bound_envelope} is replaced by the bound $\mathbb{E}[F(Y)^2]^{1/2}\leq c\sigma\sqrt{d}$.
Finally Proposition \ref{prop:numerator} becomes:

\begin{proposition}[Bound on numerator] \label{prop:numerator_improved}
With probability at least $1 - \delta$,
$$\|\mathbb{P}_n - \mathbb{P}\|_{\mathcal{F}} \leq  C'_1(R) := c \cdot \frac{\sigma\sqrt{d} }{\delta \sqrt{n}}\cdot \bigl[ R\,( \sigma\sqrt{d}\, \|A\|_2 \, +\, \sup_{x\in B_R}\|\nabla\alpha(x)\|_2) + 1 \Bigr] .
$$
\end{proposition}
This is a consequence of \eqref{eq:bound_lip_envel}. 

We turn to bounding the denominator $\mathbb{E}(e^{\langle Ax,y\rangle-\alpha(x)})$. 
For the same reasons we find,
\begin{proposition}[Bound on denominator] \label{prop:denominator_improved}
With probability at least $1 - \delta$,
$$\|\mathbb P_n - \mathbb P\|_{\mathcal{F'}} \leq  C'_2(R) := \frac{C_2}{\delta \, \sqrt{n}}(\sigma \sqrt{d}\|A\|_2+\sup_{x\in B_R}\|\nabla \alpha(x)\|_2).
$$
\end{proposition}
As in Part 3 of Appendix \ref{app:proof_thm_unif_cv}, we can define 
\begin{align*}
    \mathcal{E}_1 &=  \{\sup_{x \in B_R} \|N_n(x)-N(x)\|_2 \le C'_1(R)\} \\
    \mathcal{E}_2 &= \{\sup_{x \in B_R} |D_n(x)-D(x)| \le C'_2(R)\}.
\end{align*}
Let
\[c_{\min}
 = \exp \left(
\sup_{x \in B_R} \varphi(Ax)\right)
\]
where $\varphi$ is introduced in the statement.  Let also
\begin{align*}
    \mathcal{E}_3 &=  \{\sup_{x \in B_R} |D_n(x)-D(x)| \le \frac{1}{2c_{\min}}\}\\
    \mathcal{E}_4 &= \{\forall x \in B_R,  D_n(x) \ge \frac{1}{2c_{\min}}\} = \{\inf_{x \in B_R} D_n(x) \ge \frac{1}{2c_{\min}}\}.
\end{align*}
In particular, with this definition of $c_{\min}$, we have for all $ x \in B_R$,
$$D(x) = \mathbb{E}[e^{\langle Ax,y\rangle -\alpha(x)}] \ge \frac{1}{c_{\min}}.$$
By hypothesis, for $C_d  > 0$, 
$$
\sup_{|t|\leq t_0} \varphi(t) \leq C_d  \log(t_0)
$$
Hence, for $t_0 = \|A\|_2 R R_0$, we have
$$c_{\min} \le (\|A\|_2 R R_0)^{C_d} .$$
By Proposition \ref{prop:denominator_improved}, if 
$$
C'_2(R) := \frac{C_2}{\delta_1 \, \sqrt{n}}(\sigma \sqrt{d}\|A\|_2+\sup_{x\in B_R}\|\nabla \alpha(x)\|_2).
$$
then the event 
\begin{align*}
    \mathcal{E}_2 &= \{\sup_{x \in B_R} |D_n(x)-D(x)| \le C'_2(R)\}
\end{align*}
is verified with probability $1-\delta_1$.
%,where $C_1(R)$ and $C_2(R)$ come from Propositions \ref{prop:numerator} and \ref{prop:denominator}.
Then, for 
$$n \ge n'_{\min}(\delta, \Sigma, A, R, d):= 4 \frac{C^2_2\, (\|A\|_2 R R_0)^{2 C_d }}{\delta_1^2}(\sigma \sqrt{d}\|A\|_2+\sup_{x\in B_R}\|\nabla \alpha(x)\|_2)^2 \, , $$
one has $C_2(R) \le \frac{1}{2\, c_{\min}}$, hence $\mathcal{E}_2 \subseteq \mathcal{E}_3$, and therefore $\mathbb{P}(\mathcal{E}_3) \ge \mathbb{P}(\mathcal{E}_2) \ge 1 - \delta_1$. 
On $\mathcal{E}_3$, the event $\mathcal{E}_4$ is also verified by the inclusion $\mathcal{E}_3 \subseteq \mathcal{E}_4$, which allows us to bound $D_n(x) \ge \frac{1}{2 c_{\min}}$ uniformly in $x \in B_R$.
Therefore, with probability $1 - \delta_1$, and with $n \ge n'_{min}(\delta_1, \Sigma, A, R, d)$, 
\begin{equation}
\sup_{x \in B_R}\|\frac{N_n}{D_n} - \frac{N}{D}\|_2 \le 2 \, c_{\min} (\sup_{x \in B_R}\|N_n - N\|_2 + \sup_{x \in B_R}\|N\|_2 \cdot C'_2(R) \, c_{\min} ). \label{eq:eq3}\end{equation}
where we used $\sup_{x \in B_R}(\|N\|_2 \cdot |D_n(x)-D(x)|) \le \sup_{x \in B_R}\|N\|_2 \cdot \sup_{x \in B_R}|D_n(x)-D(x)|$, and $\sup_{x \in B_R}|D_n(x)-D(x)| \le C'_2(R)$ on $\mathcal{E}_2$.

By Proposition \ref{prop:numerator}, if
$$C'_1(R) := c \cdot \frac{\sigma\sqrt{d} }{\delta \sqrt{n}}\cdot \bigl[ R\,( \sigma\sqrt{d}\, \|A\|_2 \, +\, \sup_{x\in B_R}\|\nabla\alpha(x)\|_2) + 1 \Bigr]$$
the event
$$
\mathcal{E}_1 =  \{\sup_{x \in B_R} \|N_n(x)-N(x)\|_2 \le C'_1(R)\} 
$$
is verified with probability $1 - \delta_2$. 

Then, the same derivation as before yields $\|N\|_2 = \|\mathbb E[Y \, e^{\langle Ax, Y\rangle - \alpha(x)}]\|_2 \le c \sigma \sqrt{d}.$
Combining events $\mathcal{E}_1$ and $\mathcal{E}_2$, and for $n \ge n'_{min}(\delta, \Sigma, A, R, d)$, we have with probability $1-\delta$ (taking $\delta_1 = \delta_2 =  \frac{\delta}{2})$, 
$$\sup_{x \in B_R}\|\frac{N_n}{D_n} - \frac{N}{D}\|_2 \le \|V\|_2 \cdot\frac{ C \,\sigma \sqrt{d}\, (\|A\|_2 R R_0)^{ C_d } \bigl[(R + (\|A\|_2 R R_0)^{ C_d })(\sigma \, \sqrt{d}\, \|A\|_2 + \sup_{x\in B_R}\|\nabla \alpha(x)\|_2) + 1 \bigr]}{\delta \, \sqrt{n}}.
$$
where we recall that $\sigma = \sqrt{\|\Sigma\|_2}$.
This concludes the proof of Theorem \ref{thm:uniform_convergence_improved}.
\end{proof}

\end{document}